\pdfoutput 1
\documentclass[10pt,journal,compsoc]{IEEEtran}
\usepackage{algorithm}
\usepackage[noend]{algpseudocode}
\usepackage{amsfonts}
\usepackage{amsmath}
\usepackage{amssymb}
\usepackage{amstext}
\usepackage{bm}
\usepackage{bbm}
\usepackage{booktabs}
\usepackage{chngcntr}
\usepackage{color}
\usepackage{colortbl}
\usepackage[inline,shortlabels]{enumitem}
\usepackage{float}
\usepackage[symbol]{footmisc}
\usepackage[T1]{fontenc}
\usepackage{graphicx}
\usepackage[utf8]{inputenc}
\usepackage{mathtools}
\usepackage{multirow}
\usepackage{pgfplots}
\usepackage{setspace}
\usepackage[caption=false,font=footnotesize,position=top]{subfig}   
\usepackage{textcomp}
\usepackage[para,online,flushleft]{threeparttable}
\usepackage{tikz}
\usetikzlibrary{plotmarks,external}
\usepackage{times}
\usepackage{xcolor}
\usepackage{xspace}

\pgfplotsset{compat=newest}
\usepgfplotslibrary{patchplots}

\usepackage[most]{tcolorbox}
\usepackage{ragged2e}
\newcommand{\secref}[1]{Sec.~{\ref{#1}}}
\newcommand{\figref}[1]{Fig.~{\ref{#1}}}
\newcommand{\tabref}[1]{Table~{\ref{#1}}}

\newcommand{\secsref}[1]{Secs.~{\ref{#1}}}
\newcommand{\figsref}[1]{Figs.~{\ref{#1}}}

\newcommand{\RANSAC}{\textsc{ransac}\xspace}

\newcommand{\LORANSAC}{\textsc{lo-ransac}\xspace}

\newcommand{\etal}{et al.\ }

\newcommand{\M}[1]{\mathtt{#1}}

\newcommand{\R}[1][]{\ensuremath{\mathbb{R}^{#1}}\xspace}
\newcommand{\RP}[1][]{\ensuremath{\mathbb{RP}^{#1}}\xspace}

\def\Gb{Gr{\"o}bner basis\xspace}

\def\Gbs{Gr{\"o}bner bases\xspace}

\mathchardef\mhyphen="2D 
\newcommand{\DES}{DES\xspace}

\newcommand{\EVP}{EVP\xspace}
\newcommand{\EVL}{EVL\xspace}

\makeatletter
\newenvironment{customlegend}[1][]
{%
    \begingroup
    \pgfplots@init@cleared@structures
    \pgfplotsset{#1}%
}
{
  \pgfplots@createlegend
    \endgroup
}

\def\addlegendimage{\pgfplots@addlegendimage}
\makeatother

\newlength\fwidth 

\DeclarePairedDelimiter{\diagfences}{(}{)}
\newcommand{\diag}{\operatorname{diag}\diagfences}

\newcommand{\inv}{{-1}}
\newcommand{\T}{{\!\top}}

\newcommand{\ma}[1]{\ensuremath{\mathtt{#1}}\xspace}
\newcommand{\ve}[1][x]{\ensuremath{\mathbf{#1}}\xspace}

\newcommand{\buildset}[3]{\ensuremath{ \{\,#1\,\}_{#2}^{#3} }\xspace}

\newcommand{\mA}[1][]{\ensuremath{\ma{A}_{#1}}\xspace}

\newcommand{\mH}{\ensuremath{\ma{H}}\xspace}
\newcommand{\mHu}{\ensuremath{\ma{H}_{\ve[u]}}\xspace}

\newcommand{\mP}{\ensuremath{\ma{P}}\xspace}

\newcommand{\invT}{-{\!\top}}

\newcommand{\vld}{\ensuremath{\tilde{\ve[l]}}\xspace}
\newcommand{\vl}{\ensuremath{\ve[l]}\xspace}
\newcommand{\linf}{\ensuremath{\ve[l]_{\infty}}\xspace}
\newcommand{\vlinf}{\ensuremath{\ve[l]_{\infty}}\xspace}
\newcommand{\vu}[1][]{\ensuremath{\ve[u]_{#1}}\xspace}
\newcommand{\vv}[1][]{\ensuremath{\ve[v]_{#1}}\xspace}

\newcommand{\vlhat}{\ensuremath{\ve[\hat{l}]}\xspace}

\newcommand{\ie}{{\em i.e.}\xspace}
\newcommand{\eg}{{\em e.g.}\xspace}
\newcommand{\Eg}{{\em E.g.}\xspace}

\newcommand{\sksym}[1]{\ensuremath{\left[{#1}\right]_{\times}}\xspace}

\newcount\colveccount
\newcommand*\colvec[1]{
        \global\colveccount#1
        \begin{pmatrix}
        \colvecnext
}
\def\colvecnext#1{
        #1
        \global\advance\colveccount-1
        \ifnum\colveccount>0
                \\
                \expandafter\colvecnext
        \else
                \end{pmatrix}
        \fi
}

\newtoks\rowvectoks
\newcommand{\rowvec}[2]{%
  \rowvectoks={#2,}\count255=#1\relax
  \advance\count255 by -1
  \rowvecnexta}
\newcommand{\rowvecnexta}{%
  \ifnum\count255>0
    \expandafter\rowvecnextb
  \else
    \setlength\arraycolsep{1pt}     
    \begin{pmatrix}\the\rowvectoks\end{pmatrix}
  \fi}
\newcommand\rowvecnextb[1]{%
  \ifnum\count255>1     
    \rowvectoks=\expandafter{\the\rowvectoks&#1,}%
  \else
    \rowvectoks=\expandafter{\the\rowvectoks&#1}%
  \fi
    \advance\count255 by -1
    \rowvecnexta}

\newcommand{\ic}[2]{\parbox{#1\linewidth}{\centering \includegraphics[width=0.99\linewidth]{#2}}}

\definecolor{Gray}{gray}{0.9}
\definecolor{LightGray}{gray}{0.95}
\newcolumntype{a}{>{\columncolor{Gray}}c}
\newcolumntype{b}{>{\columncolor{LightGray}}c}
\newcommand{\ra}[1]{\renewcommand{\arraystretch}{#1}}

\newcolumntype{L}[1]{>{\raggedright\let\newline\\\arraybackslash\hspace{0pt}}m{#1}}
\newcolumntype{C}[1]{>{\centering\let\newline\\\arraybackslash\hspace{0pt}}m{#1}}
\newcolumntype{g}[1]{>{\columncolor{Gray}\centering\let\newline\\\arraybackslash\hspace{0pt}}m{#1}}
\newcolumntype{R}[1]{>{\raggedleft\let\newline\\\arraybackslash\hspace{0pt}}m{#1}}

\pgfplotsset{
    legend image with text/.style={
        legend image code/.code={%
            \node[anchor=center] at (0.3cm,0cm) {#1};
        }
    },
}

\newcommand{\rmswarp}{\ensuremath{\Delta^{\mathrm{warp}}_{\mathrm{RMS}}}\xspace}
\newcommand{\rmsxfer}{\ensuremath{\Delta^{\mathrm{xfer}}_{\mathrm{RMS}}}\xspace}

\makeatletter
\def\munderbar#1{\underline{\sbox\tw@{$#1$}\dp\tw@\z@\box\tw@}}
\makeatother

\newcommand{\xp}[1][]{\ensuremath{x^{\prime}}\xspace}
\newcommand{\yp}[1][]{\ensuremath{y^{\prime}}\xspace}
\newcommand{\zp}[1][]{\ensuremath{z^{\prime}}\xspace}
\newcommand{\rp}[1][]{\ensuremath{r^{\prime}}\xspace}

\newcommand{\xr}[1][]{\ensuremath{\munderbar{x}_{#1}}\xspace}
\newcommand{\yr}[1][]{\ensuremath{\munderbar{y}_{#1}}\xspace}

\newcommand{\xd}[1][]{\ensuremath{\tilde{x}_{#1}}\xspace}
\newcommand{\yd}[1][]{\ensuremath{\tilde{y}_{#1}}\xspace}

\newcommand{\vU}[1][]{\ensuremath{\ve[U]_{#1}}\xspace}
\newcommand{\vV}[1][]{\ensuremath{\ve[V]_{#1}}\xspace}

\newcommand{\vX}[1][]{\ensuremath{\ve[X]_{#1}}\xspace}
\newcommand{\vXp}[1][]{\ensuremath{\ve[X]^{\prime}_{#1}}\xspace}

\newcommand{\vx}[1][]{\ensuremath{\ve[x]_{#1}}\xspace}
\newcommand{\vxp}[1][]{\ensuremath{\vx[#1]^{\prime}}\xspace}
\newcommand{\vxd}[1][]{\ensuremath{\ve[\tilde{x}]_{#1}}\xspace}
\newcommand{\vxdp}[1][]{\ensuremath{\ve[\tilde{x}]^{\prime}_{#1}}\xspace}
\newcommand{\vxr}[1][]{\ensuremath{\munderbar{\vx}_{#1}}\xspace}

\newcommand{\rgnd}[1][]{\ensuremath{\tilde{\mathcal{R}}_{#1}}\xspace}
\newcommand{\rgn}[1][]{\ensuremath{\mathcal{R}_{#1}}\xspace}
\newcommand{\rgnr}[1][]{\ensuremath{\munderbar{\mathcal{R}}_{#1}}\xspace}

\newcommand{\xcspond}[1][]{\ensuremath{\vx[#1] \leftrightarrow \vxp[#1]} \xspace}
\newcommand{\Xcspond}[1][]{\ensuremath{\vX[#1] \leftrightarrow \vXp[#1]} \xspace}
\newcommand{\xdcspond}[1][]{\ensuremath{\vxd[#1] \leftrightarrow \vxdp[#1]} \xspace}
\newcommand{\xijcspond}[1][]{\ensuremath{\vx[i] \leftrightarrow \vx[j]} \xspace}
\newcommand{\xpijcspond}[1][]{\ensuremath{\vxp[i] \leftrightarrow \vxp[j]} \xspace}

\newcommand{\pid}[1][]{\ensuremath{\tilde{\pi}}\xspace}
\newcommand{\pir}[1][]{\ensuremath{\munderbar{\pi}}\xspace}

\newcommand{\rgntwotwotwodes}{\ensuremath{\mH^{\text{\DES}}_{222}\vl\lambda}\xspace}

\newcommand{\rgntwoct}{\ensuremath{\mH_2\vl}\xspace}
\newcommand{\rgntwotwoct}{\ensuremath{\mH_{22}\vl}\xspace}
\newcommand{\rgntwordctevl}{\ensuremath{\mH_2\vl\lambda}\xspace}
\newcommand{\rgntwordctevlrand}{\ensuremath{\mH^{\text{RND}}_2\vl\lambda}\xspace}
\newcommand{\rgntwordctevp}{\ensuremath{\mH_2\vl\vu\lambda}\xspace}
\newcommand{\rgntwordsctevp}{\ensuremath{\mH_2\vl\vu s^{\vu}\lambda}\xspace}
\newcommand{\rgntwotwordctevp}{\ensuremath{\mH_{22}\vl\vu\vv \lambda}\xspace}
\newcommand{\rgntwotwordsctevp}{\ensuremath{\mH_{22}\vl\vu\vv s^{\vv}\lambda}\xspace}

\newcommand{\rgntwotwofitz}{\ensuremath{\mH_{22} \lambda}\xspace}
\newcommand{\rgntwotwokukelova}{\ensuremath{\mH_{22} \lambda_1 \lambda_2}\xspace}

\newcommand\tabcap[1]{\vspace{1pt}\justify\small\emph{#1}}

\ifCLASSOPTIONcompsoc
  \usepackage[nocompress]{cite}
\else
  \usepackage{cite}
\fi
\ifCLASSINFOpdf
\else
\fi

\begin{document}

\title{Minimal Solvers for Rectifying from Radially-Distorted Conjugate Translations}

\author{James~Pritts,~Zuzana~Kukelova,~Viktor~Larsson,Yaroslava~Lochman,~and~Ond{\v r}ej~Chum
  \IEEEcompsocitemizethanks{
    \IEEEcompsocthanksitem J. Pritts is with Czech Institute of Informatics, Robotics and Cybernetics (CIIRC) at Czech Technical University in Prague.\protect\\
    E-mail: prittjam@cvut.cz
    \IEEEcompsocthanksitem Z. Kukelova and O. Chum are with the Visual Recognition Group (VRG) in the Faculty of Electrical Engineering at Czech Technical University in Prague.
    \IEEEcompsocthanksitem V. Larsson is with the Computer Vision and Geometry Group
(CVG) in the Department of Computer Science at ETH Zurich.
    \IEEEcompsocthanksitem Y. Lochman is with The Machine Learning Lab at Ukrainian Catholic University in Lviv, Ukraine.}
    \thanks{Manuscript received April 19, 2005; revised August 26, 2015.}
  }

\markboth{}{}

\IEEEtitleabstractindextext{%
\justify
\begin{abstract}
This paper introduces minimal solvers that jointly solve for radial
lens undistortion and affine-rectification using local features
extracted from the image of coplanar translated and reflected scene
texture, which is common in man-made environments. The proposed
solvers accommodate different types of local features and sampling
strategies, and three of the proposed variants require just one
feature correspondence. State-of-the-art techniques from algebraic
geometry are used to simplify the formulation of the solvers. The
generated solvers are stable, small and fast. Synthetic and real-image
experiments show that the proposed solvers have superior robustness to
noise compared to the state of the art.  The solvers are integrated
with an automated system for rectifying imaged scene planes from
coplanar repeated texture. Accurate rectifications on challenging
imagery taken with narrow to wide field-of-view lenses demonstrate the
applicability of the proposed solvers.
\end{abstract}

\begin{IEEEkeywords}
rectification, radial distortion, minimal solvers, symmetry, repeated patterns, local features
\end{IEEEkeywords}

}

\maketitle

\IEEEdisplaynontitleabstractindextext
\IEEEpeerreviewmaketitle

\IEEEraisesectionheading{\section{Introduction}\label{sec:introduction}}
%
%
%
%
\IEEEPARstart{S}{cene}-plane rectification is used in many classic computer-vision
tasks, including symmetry detection and discovery of near-regular
textures \cite{Hays-ECCV06,Funk-ICCV17};
inpainting \cite{Lukac-ACMTG17}; single-view 3D
reconstruction \cite{Wu-CVPR11}; using repetitions to improve
multi-label segmentation \cite{Pritts-BMVC16,Zeng-CVPR08}, and
single-view auto-calibration using the Manhattan scene
assumption \cite{Wildenauer-CVPR12,Wildenauer-BMVC13,Antunes-CVPR17}. In
particular, the affine rectification of a scene plane transforms the
camera's principal plane so that it is parallel to the scene
plane. This restores the affine invariants of the imaged scene plane,
which include parallelism of lines and translational symmetries
\cite{Hartley-BOOK04,Pritts-CVPR14}. There is only an affine
transformation between the affine-rectified imaged scene plane and its
real-world counterpart. The removal of the effects of perspective
imaging is helpful to understanding the geometry of the scene plane.

\begin{figure}[htb]
  \centering 
  \captionsetup[subfigure]{labelformat=empty}
  \subfloat[GoPro Hero 4 Wide, 17.2mm \label{fig:GoPro_Hero4_wide_first}]
    { \includegraphics[width=0.495\columnwidth]{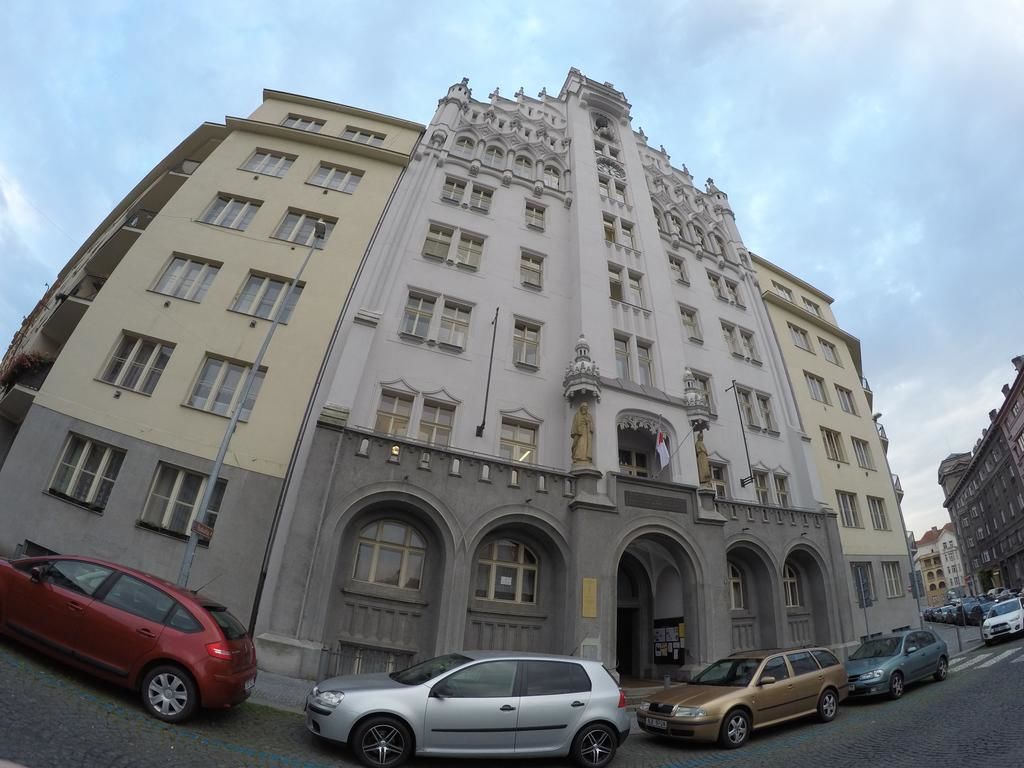} \hfill \includegraphics[width=0.495\columnwidth]{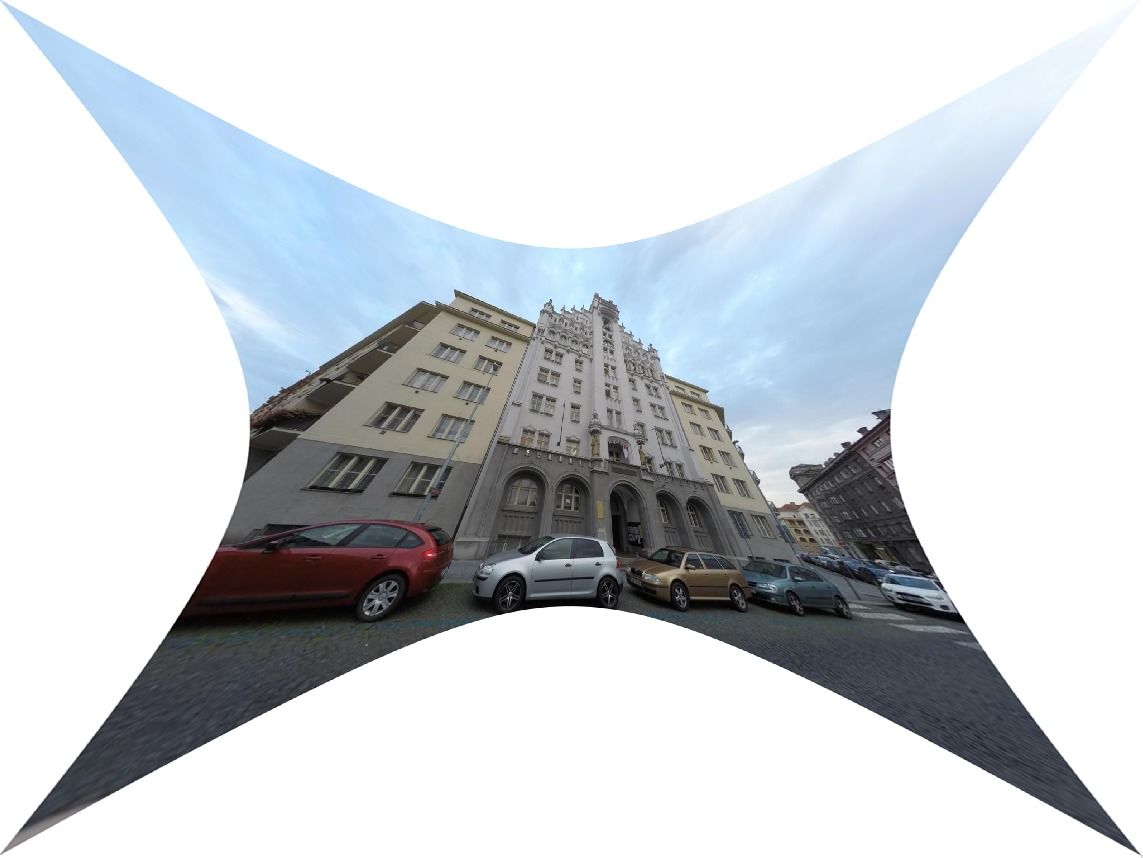}
      }
\vspace{0.1cm}
   \includegraphics[width=0.995\columnwidth]{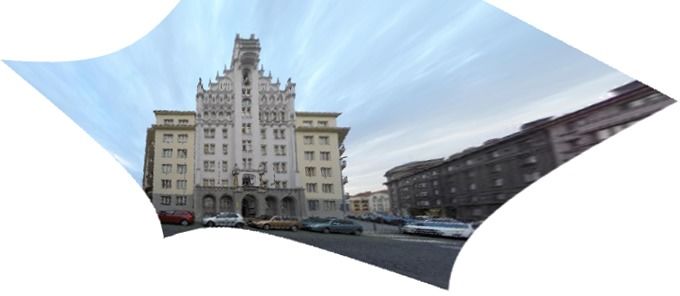}
   \caption{Input (top left) is a distorted view of a scene plane with
    translational symmetries and reflections, and the outputs (top
    right, bottom) are the radially undistorted image and the rectified scene plane. The
    method is fully automatic.}
    \vspace{-15pt}
    \label{fig:first}
\end{figure}

Wide-angle imagery that has significant lens distortion is common
since consumer photography is now dominated by mobile-phone and
GoPro-type cameras. High-accuracy rectification from wide-angle
imagery is not possible with only pinhole camera models
\cite{Kukelova-CVPR15,Wildenauer-BMVC13}. Lens distortion can be
estimated by performing a camera calibration apriori, but a fully
automated method is desirable.  Furthermore, in the case of Internet
imagery, the camera and its metadata are often unavailable for use
with off-line calibration techniques. 

\begin{figure*}[t!]
  \centering \resizebox{0.9\textwidth}{!}{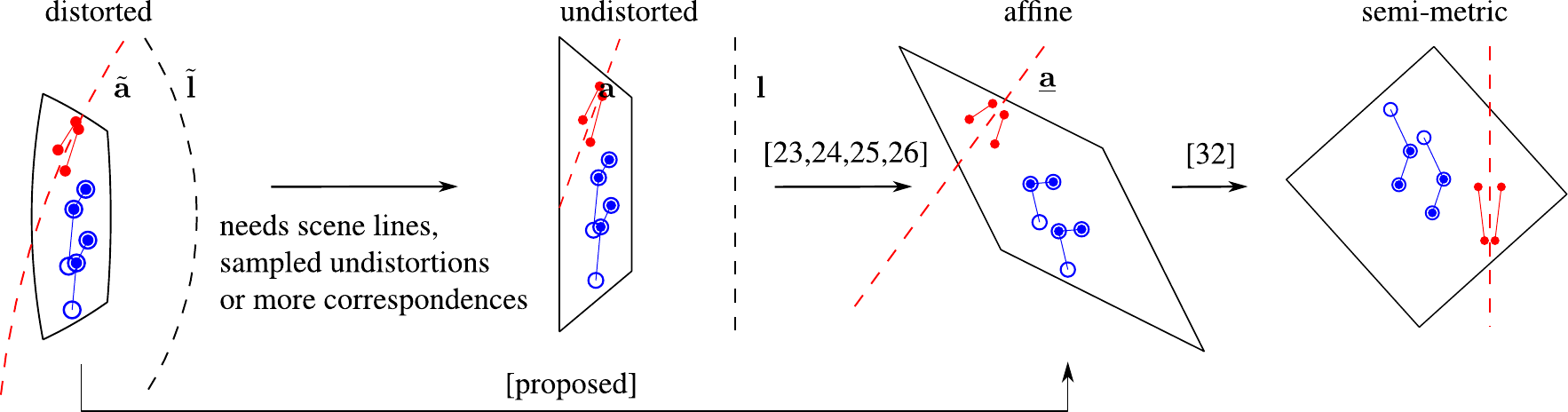} \caption{Direct
  Affine Rectification. The rectification hierarchy is traversed from
  left to right. The proposed solvers directly rectify from distorted
  local features. The state-of-the art requires sampled undistortions,
  scene lines \cite{Antunes-CVPR17,Bukhari-JMIV13,Wildenauer-BMVC13},
  or more region correspondences \cite{Pritts-ACCV18}. Color denotes
  how coplanar repeats relate: blue for translation and red for
  reflection. Marker type denotes the possible correspondence
  configurations: circles for three translated point correspondences
  and filled circles for two pairs of two translated point
  correspondences on a scene plane. Reflections (in red) can be
  detected as two point correspondences translating over different
  distances in the same direction. The distorted and undistorted
  images of the scene plane's vanishing line are denoted \vld and \vl,
  and the distorted and undistorted reflection axis is similarly
  denoted \ve[\tilde{a}] and \ve[a], where \munderbar{\ve[a]} is its
  rectification. Point correspondences (circles) are extracted from
  region correspondences (solid polylines), which reduces the required
  input to one or two region correspondences.}
  \label{fig:ct_composite_rect}
  \vspace{-8pt}
\end{figure*}

\begin{figure}[H]
    \centering
     \resizebox{0.45\textwidth}{!}{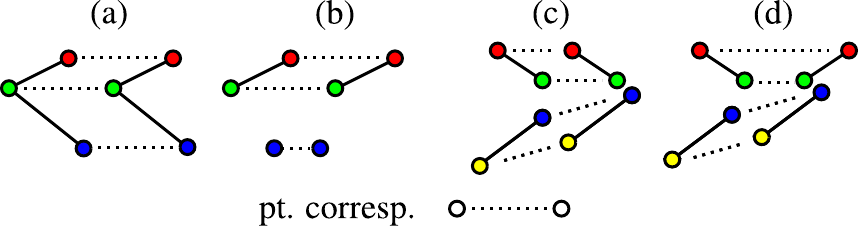} \caption[]{The
     proposed solver variants can use: \begin{enumerate*}[(a)] \item
     three points translated over the same distance in
     one-direction, \item three points translated in one direction
     with one point (blue) at an arbitrary distance, \item two point
     pairs translated over two distances in two directions, and \item
     two point pairs translated in two directions with one pair
     translating the same distance and one pair at different
     distances.\end{enumerate*}}
\label{fig:input_configurations}
\end{figure}

This paper proposes minimal solvers that jointly estimate lens
undistortion and affine rectification using point correspondences
extracted from the image of coplanar repeated texture that is related
by translations or reflections on the scene
plane. See \figref{fig:first} for an example rectification. The
proposed solver variants differ by the expected configuration and the
number of required point correspondences.  The solvers can rectify
from distorted local features that are translated on the scene plane
in one or two directions, where some of the point correspondences can
translate with arbitrary distances. There is also variant that admits
reflections. \figref{fig:input_configurations} shows an example of
each of the four configurations of point correspondences on the scene
plane that are handled by the proposed
solvers. \figref{fig:ct_composite_rect} shows these configurations in
each stage of the rectification hierarchy and that the solver variants
can use these configurations directly affine rectify the distorted
image of the scene plane. These configurations often occur in man-made
settings, where there are many symmetries. \Eg, windows have scene
structure that is consistent with all four correspondence
configurations.

The straightforward application of solver generators to this problem
results in unstable
solvers \cite{Kukelova-ECCV08,Larsson-CVPR17,Larsson-CVPR18}. Stable
solver generation requires the elimination of some unknowns from the
formulations of each variant. The generated solvers differ
significantly in time-to-solution and noise sensitivity with respect
to the choice of eliminated unknowns.  These different formulations
are derived and analyzed. In particular, an exceptionally fast and
robust solver is generated compared to the solvers introduced
in \cite{Pritts-CVPR18} for feature configuration (a)
of \figref{fig:input_configurations}. All solvers are optimized in C++
and are available at \url{https://github.com/prittjam/repeats}.

Automatic scene-plane rectification is an ill-posed single-view
geometry estimation task that is further complicated if there is lens
distortion. Robust sampling schemes like \RANSAC are required to
obtain high-accuracy rectifications from noisy or bad local feature
correspondences \cite{Fischler-CACM81,Schaffalitzky-BMVC98,Chum-ACCV10,Pritts-BMVC16}. Reducing
the cardinality of the minimal sample set can greatly reduce the
number of trials needed by \RANSAC to find a good
solution \cite{Hartley-BOOK04}. Region correspondences can be used to
simultaneously establish multiple point correspondences, which reduces
the the minimal sample size of point-based
solvers \cite{Barath-CVPR17,Mikolajczyk-IJCV04,Perdoch-ICPR06,Raposo-CVPR16}. The
proposed solvers are designed such that this substitution is
straightforward. The solid polylines connecting points
in \figref{fig:input_configurations} denote the cases in which a
region correspondence can replace multiple point correspondences. The
proposed solvers require either one or two region correspondences,
have fast times-to-solution, and are robust to noisy features, which
makes them well suited for use in a \RANSAC-based estimator.

The \RANSAC-based metric-rectification framework
of \cite{Pritts-CVPR14} is extended to use the proposed minimal
solvers to synthesize the fronto-parallel images presented in this
paper. The extension includes a method to use extra available point
correspondence constructions from a region correspondence to improve
rectification accuracy. The correspondences are partitioned into
complementary subsets of minimal samples and unused correspondences,
and cross validation is used to select the best rectification. In
addition, a method for reliably finding affine-rectified regions that
can be used for the estimation of the metric upgrade is proposed.

\begin{table*}[t!]
\centering
\caption{Scene Assumptions}
\vspace{-5pt}
\ra{1}
\resizebox{\textwidth}{!}{
\begin{tabular}{@{} R{20ex}C{28ex}C{25ex}C{28ex}g{28ex} @{} }
\toprule
& Wildenauer \etal \cite{Wildenauer-BMVC13} &
Antunes \etal \cite{Antunes-CVPR17} &
Pritts \etal \cite{Pritts-ACCV18,Pritts-IJCV20} & Proposed \\
\midrule
Feature Type & fitted circles & fitted circles & covariant regions & points, covariant regions \\
Number & set of 2 and 3 lines & set of 3 and 4 lines & 3 region correspondences & 1 region correspondence  \\
Assumption & parallelism & parallelism & rigidly transformed & translated, reflected \\
\bottomrule
\end{tabular}
}\\
\tabcap{Solvers rectifying from distorted images by
\cite{Wildenauer-BMVC13,Antunes-CVPR17} require distinct sets of
parallel scene lines as
input. Pritts \etal \cite{Pritts-ACCV18,Pritts-IJCV20} requires three
region correspondences. The proposed solvers can rectify from only one
region correspondence and admit point correspondences.}
\label{tab:state_of_the_art}
\vspace{-7pt}
\end{table*}

\subsection{Previous Work}
The problem of rectification is closely coupled with the detection of
coplanar repeats in a classic chicken-and-egg scenario: rectification
is easy if the repeats are grouped, and repeats are more easily
grouped if the affine invariants of the rectified plane are available
\cite{Pritts-CVPR14}. Most local-feature based state-of-the-art methods tentatively group coplanar repeats by
their texture, which are verified by testing affine or metric
invariants with a hypothesized
rectification \cite{Ahmad-IJCV18,Aiger-EG12,Chum-ACCV10,Criminisi-BMVC00,
Lukac-ACMTG17,Ohta-IJCAI81,Schaffalitzky-BMVC98,Zhang-IJCV12}. These
methods assume a pinhole camera model, which makes them of limited use
on lens-distorted images. In particular, the method of
Schaffalitzky \etal
\cite{Schaffalitzky-BMVC98} is the most similar to the solvers proposed in
this paper since it also uses constraints induced by imaged
translational symmetries. A survey of minimal solvers that rectify
using imaged coplanar repeats is given
in \tabref{tab:solver_properties}.

The state-of-the-art has extended minimal rectifying solvers to
lens-distorted
images \cite{Antunes-CVPR17,Pritts-ACCV18,Pritts-IJCV20,Wildenauer-BMVC13}. The
common choice of parameterization for lens undistortion in ill-posed
settings is the division model since it has only one parameter and can
effectively model a wide range of radial lens
undistortions \cite{Fitzgibbon-CVPR01}. The proposed solvers also use
the division model. The capabilities of the proposed versus
state-of-the-art solvers for affine rectification are illustrated
in \figref{fig:ct_composite_rect}.

Pritts \etal \cite{Pritts-ACCV18,Pritts-IJCV20} introduced solvers
that rectify from the distorted image of rigidly-transformed coplanar
repeats.  However, these solvers are about 2000 times slower than the
fastest of the proposed solvers, require three region correspondences,
and cannot use individual point correspondences (they are strictly
region based). Wildenauer \etal
\cite{Wildenauer-BMVC13} and Antunes \etal \cite{Antunes-CVPR17} 
 are two contour-based methods that rectify from the distorted images
of parallel scene lines. These methods require two sets of imaged
parallel lines, which must have different mutual orientations. In
particular, the solver
\cite{Wildenauer-BMVC13} requires five lines total
and \cite{Antunes-CVPR17} requires seven. These are strong scene
content assumptions. In addition, unlike points or regions, contours
cannot be reliably corresponded based on appearance. These rectifying
solvers that use the division model of lens undistortion are compared
with the proposed solvers in \tabref{tab:state_of_the_art}.

The method of Li \etal \cite{Li-ICCVW05} is a two-view method that
jointly estimates undistortion and the fundamental matrix. Both
Li \etal and the proposed solvers construct a matrix from the unknown
lens undistortion parameters and use the fact that it must be singular
to impose constraints on the unknowns.

The method of Pritts \etal \cite{Pritts-CVPR14} uses a non-linear
refinement step to generalize a rectification estimated under the
pinhole camera assumption to include lens undistortion. However, even
with relaxed thresholds, a robust estimator like \RANSAC
\cite{Fischler-CACM81} discards measurements around the boundary of
the image since this region is the most affected by radial distortion
and cannot be accurately modeled with a pinhole camera. Neglecting
lens distortion during the labeling of good and bad measurements, as
done during the verification step of \RANSAC, can give fits that are
biased \cite{Kukelova-CVPR15}, which degrades rectification accuracy.

\section{Preliminaries}
Without loss of generality, coplanar scene points are assumed to be on
the scene plane $z=0$. This permits the camera matrix $\mP$ to be
modeled as the homography that changes the basis from the scene-plane
coordinate system to the camera's image-plane coordinate system
\begin{equation}
  \underbrace{
    \begin{bmatrix}
      \ve[p]_1 & \ve[p]_2 & \ve[p]_3 & \ve[p]_4
  \end{bmatrix}}_{\mP^{3 \times 4}} \colvec{4}{X}{Y}{0}{1} = 
  \underbrace{
    \begin{bmatrix}
      \ve[p]_1 & \ve[p]_2 & \ve[p]_4
    \end{bmatrix}
  }_{\mP} \underbrace{\colvec{3}{X}{Y}{1}}_{\vX},
  \label{eq:camera_as_homography}
\end{equation}
where $\ve[p]_j=\rowvec{3}{p_{1j}}{p_{2j}}{p_{3j}}^{\T}$ encode the
intrinsics and extrinsics of the camera matrix $\mP^{3 \times 4}$. The
scene and image planes are denoted $\Pi$ and $\pi$,
respectively. Scene points are denoted $\vX=\rowvec{3}{X}{Y}{1}^{\T}$
and imaged points are denoted $\vx=\rowvec{3}{x}{y}{1}^{\T}$, where
$x,y$ are the image coordinates.

\begin{table}[ht]
\centering
\caption{Common Denotations}
\vspace{-5pt}
\resizebox{\columnwidth}{!}{
\ra{1}
\begin{tabular}{@{} R{13ex}L{53ex} @{} }
  \toprule
  Term & Description \\
  \midrule
  \mP & $3 \times 3$ camera matrix viewing $z=0$ (see \eqref{eq:camera_as_homography}). \\
  \vX & homogeneous scene point in \RP[2] \\
  \vx,\,\vxd & homogeneous pinhole and distorted image point \\
  \vxr & affine-rectified point (see \eqref{eq:udrect}) \\
  \xcspond & \vx,\vxp are in correspondence with some transformation \\
  \vU,\,\vV & translations in the scene plane \\
  \vu,\,\vv & vanishing points of the trans. \vU,\vV as imaged by \mP\\
  $\ve[m]_i$ & join of undistorted point correspondence \xcspond[i] \\
  $\ve[m]_{ij},\ve[m]^{\prime}_{ij}$ & joins of \xijcspond and \xpijcspond, respectively \\
  \sksym{\cdot} & skew-symmetric operator for computing cross products \\
  \ma{T} & homogeneous translation matrix \\
  \vl,\,\vld & image of vanishing line and distorted vanishing line \\
  \vlinf & the line at infinity \\
  \mH & affine-rectifying homography \\
  \mHu & conjugate translation in the imaged trans. direction \vu \\
  $\lambda$ & division model parameter for undistortion (see \secref{sec:radial_lens_distortion}) \\
  $\Pi,\pi$ & the scene plane and image plane (in \RP[2]) \\
  \rgnd,\,\rgn,\,\rgnr  & distorted, undistorted, and affine-rectified regions \\
  \bottomrule
\end{tabular}
}
\label{tab:common_denotations}
\end{table}

\begin{figure*}[ht]
  \centering 
  \subfloat[GoPro Hero 4 Medium, 21.9mm \label{fig:GoPro_Hero4_medium_fov}] {
    \includegraphics[width=0.135\textwidth]{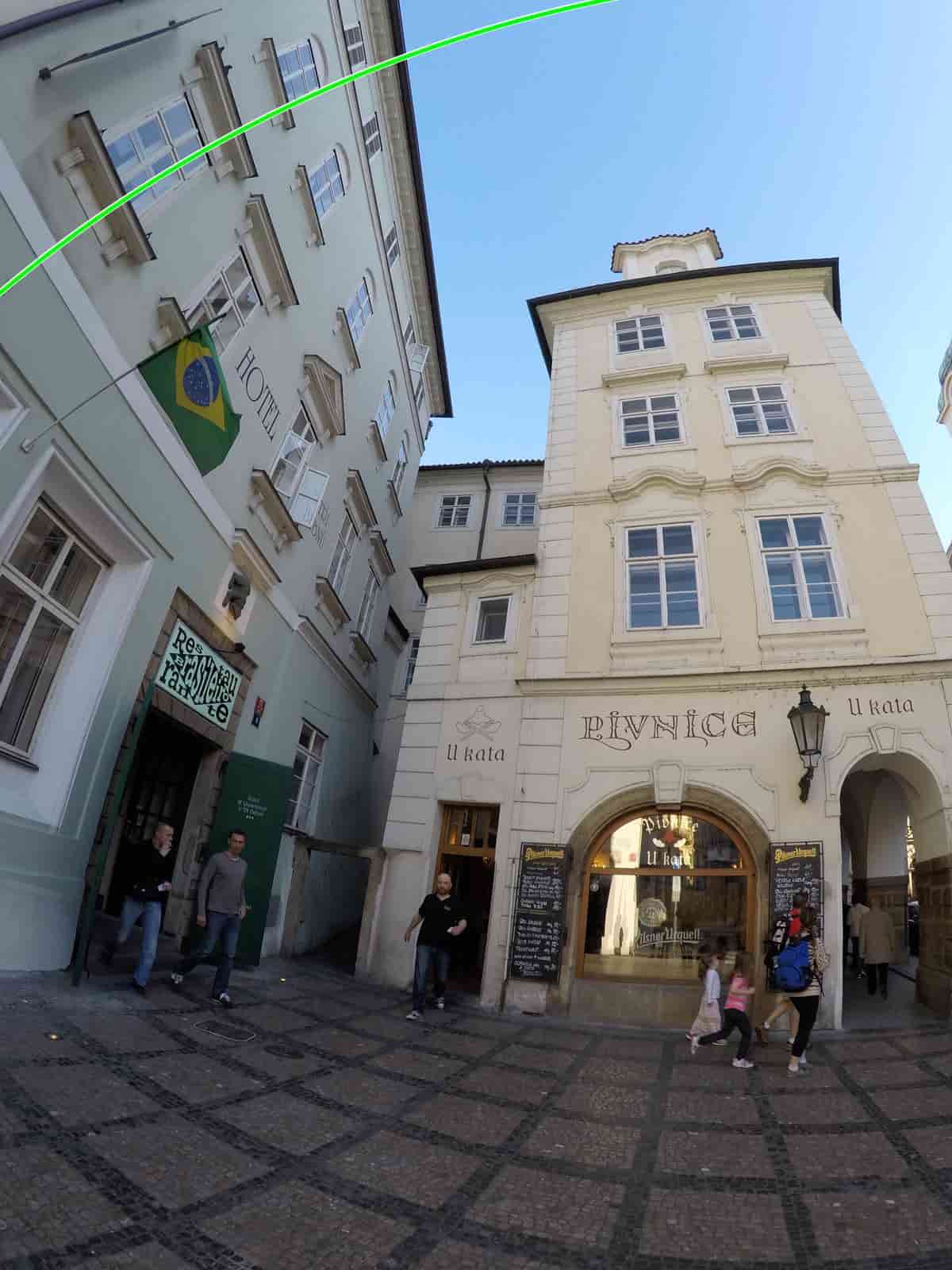}
    \hfill
    \includegraphics[width=0.24\textwidth]{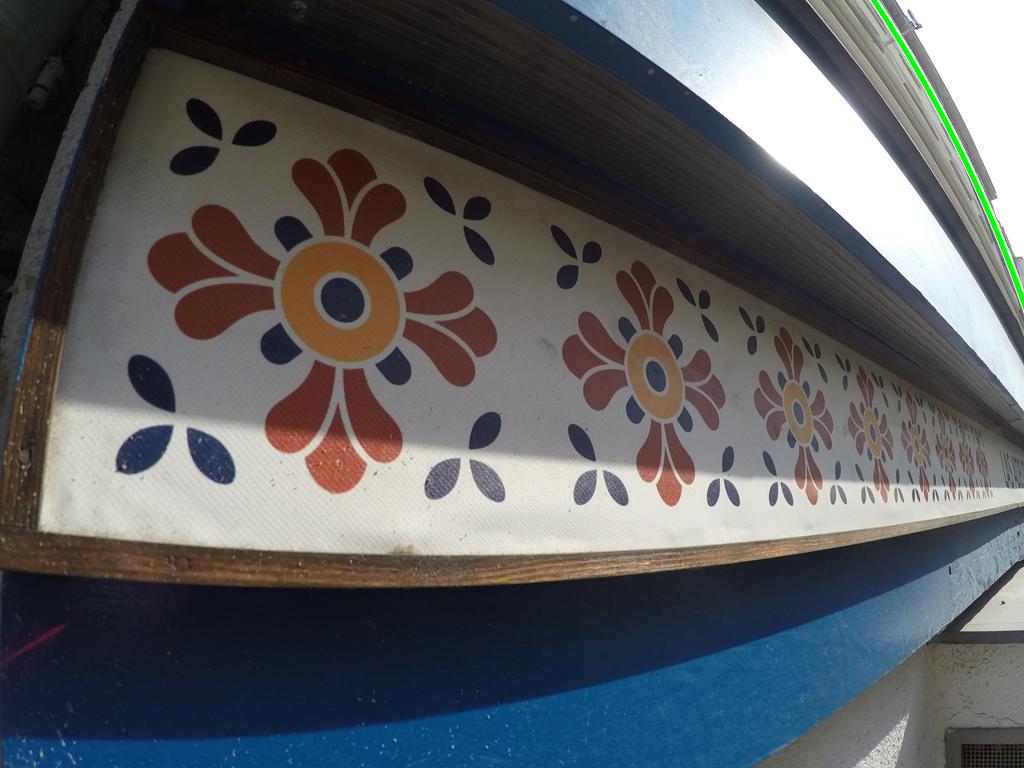}
  }
  \subfloat[GoPro Hero 4 Wide, 17.2mm \label{fig:GoPro_Hero4_wide_fov}] {
    \includegraphics[width=0.24\textwidth]{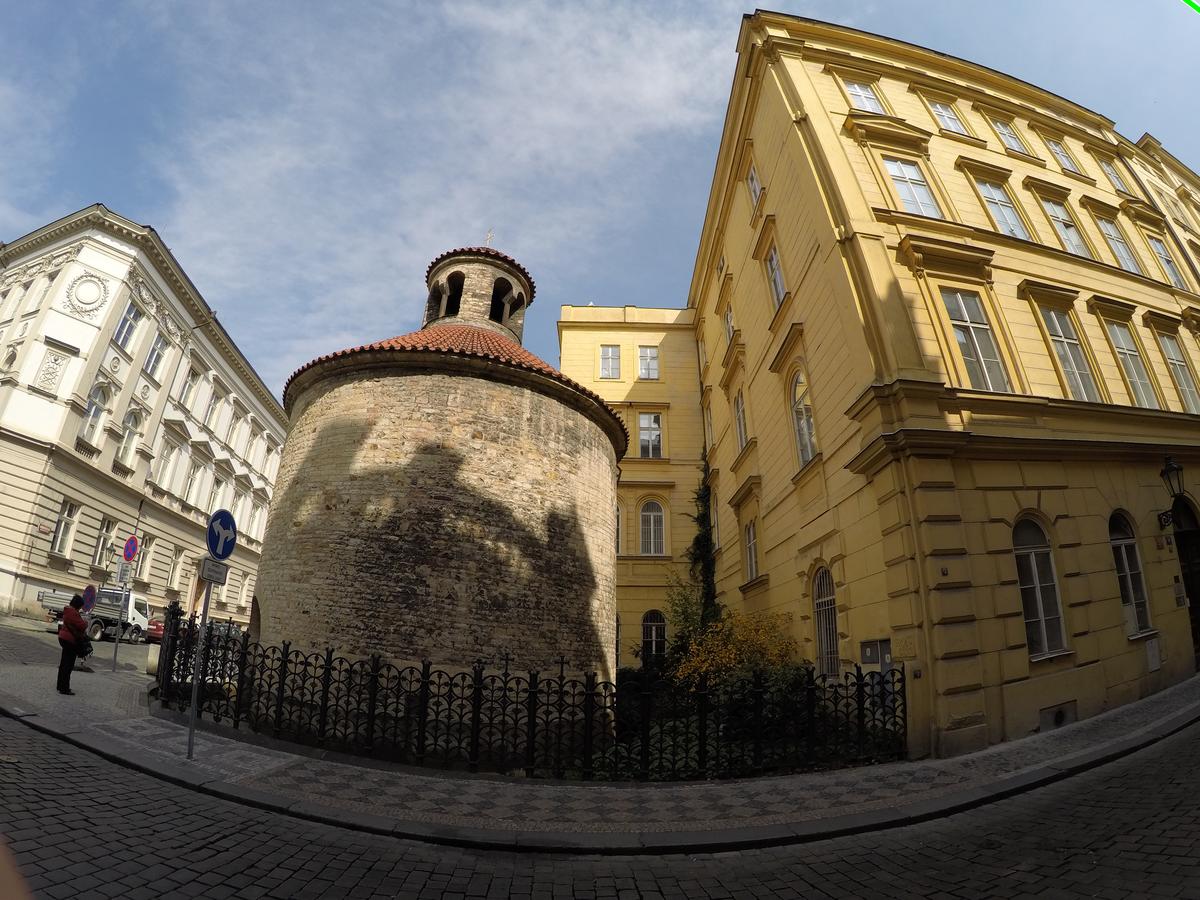}
    \hfill
    \includegraphics[width=0.24\textwidth]{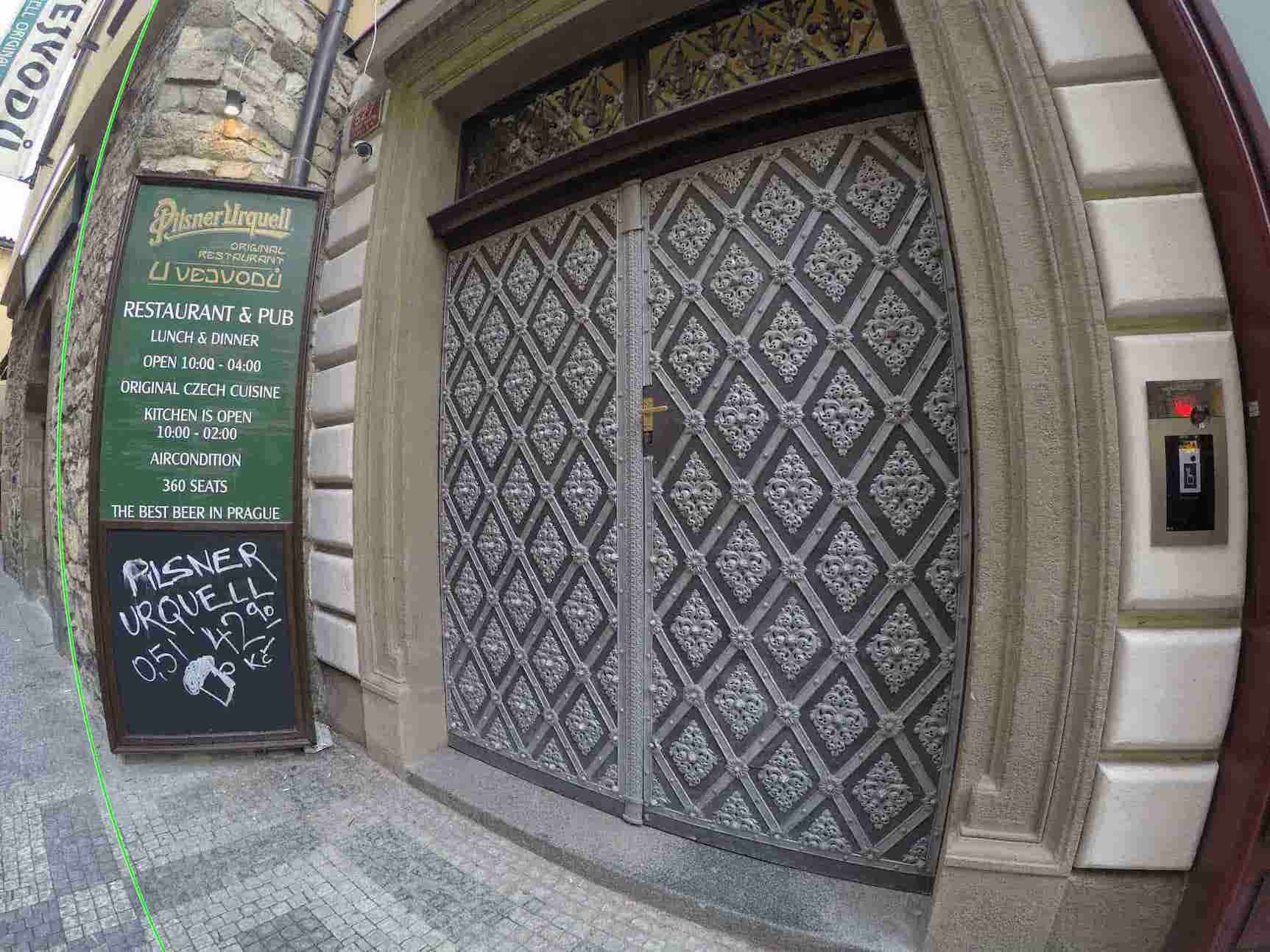}
  }
  \subfloat[Samyang, 7.5mm \label{fig:samyang75mm_fov}]{
    \includegraphics[width=0.128\textwidth]{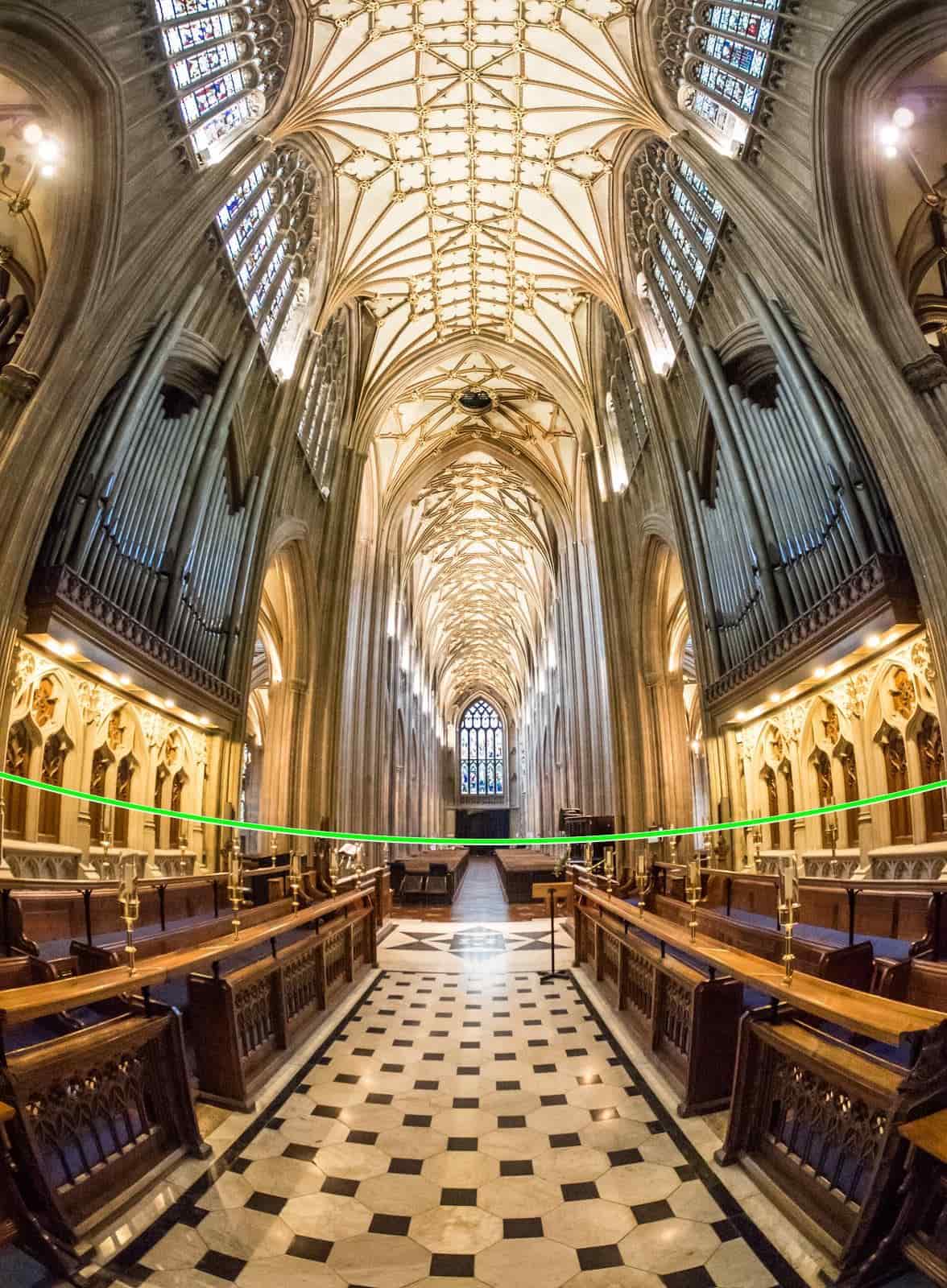}
  }
  \\[-1.5ex]
  \subfloat{
    \includegraphics[width=0.135\textwidth]{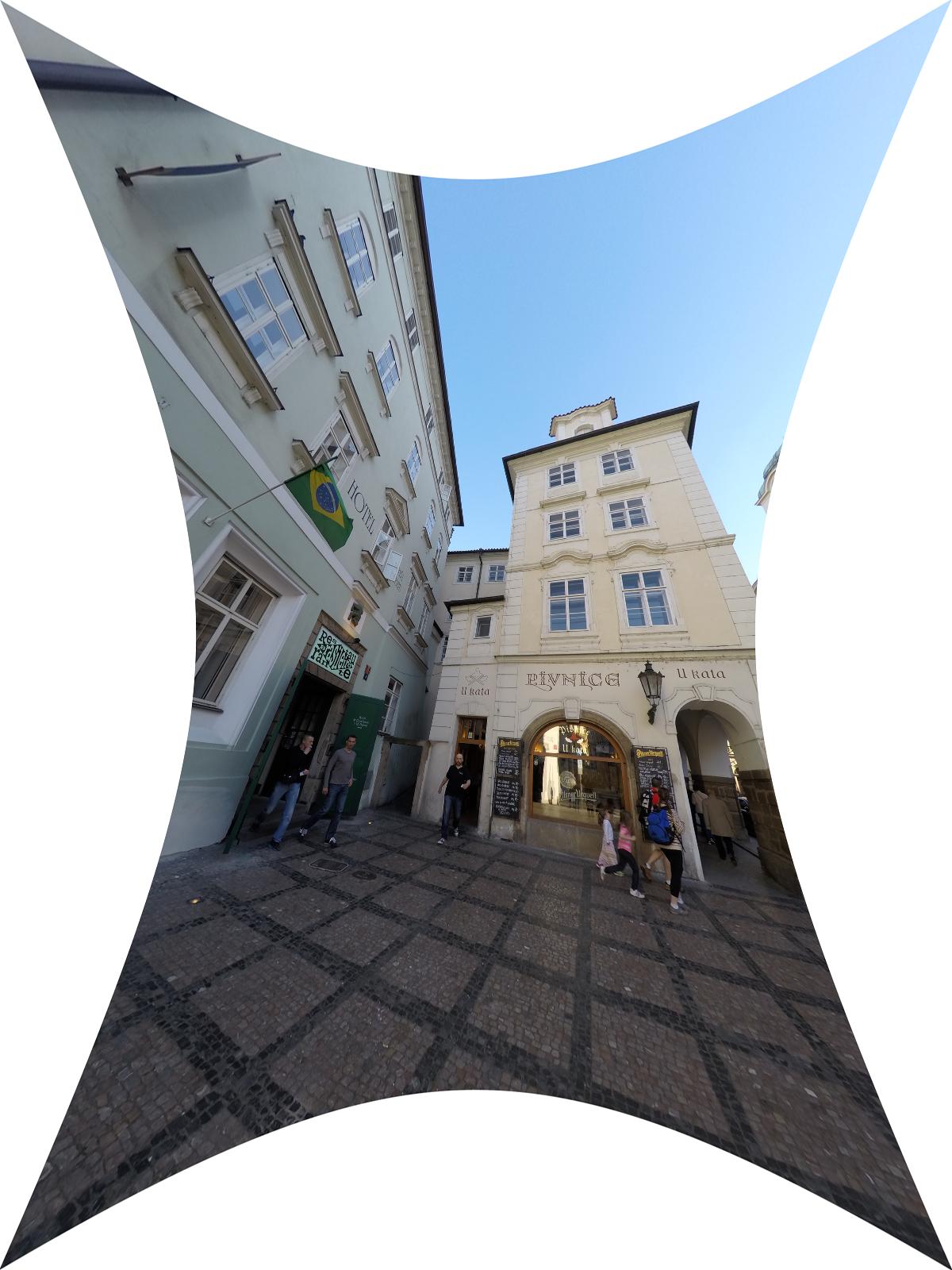}
  }
  \subfloat{
    \includegraphics[width=0.24\textwidth]{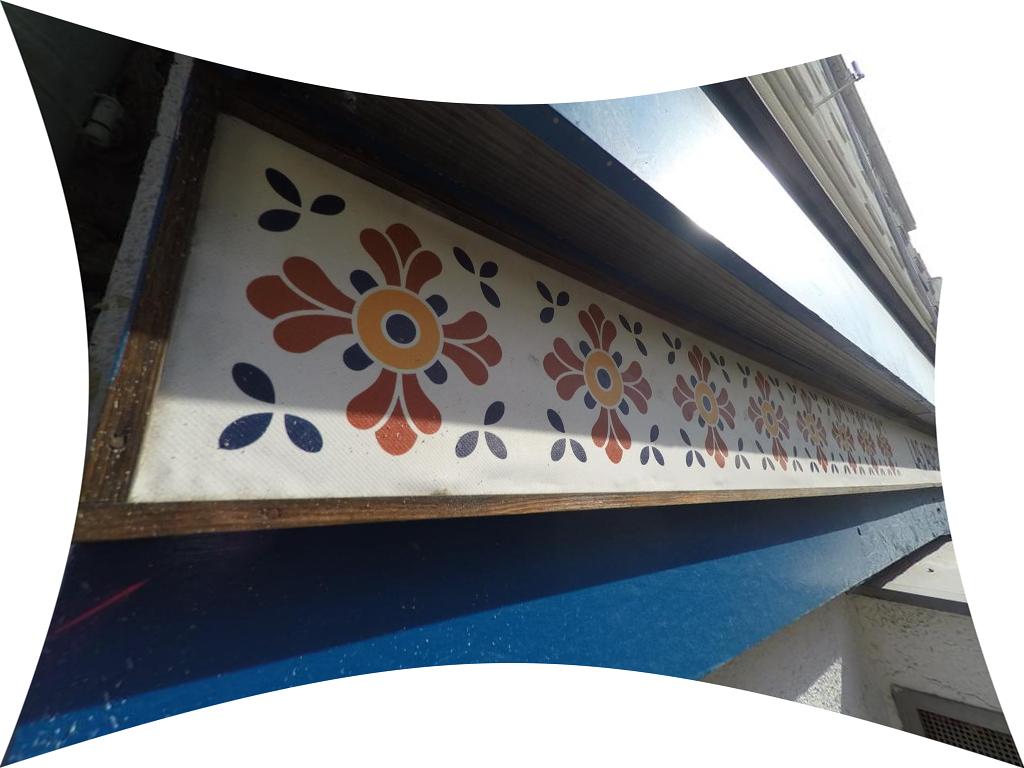}
  }
  \subfloat{
    \includegraphics[width=0.24\textwidth]{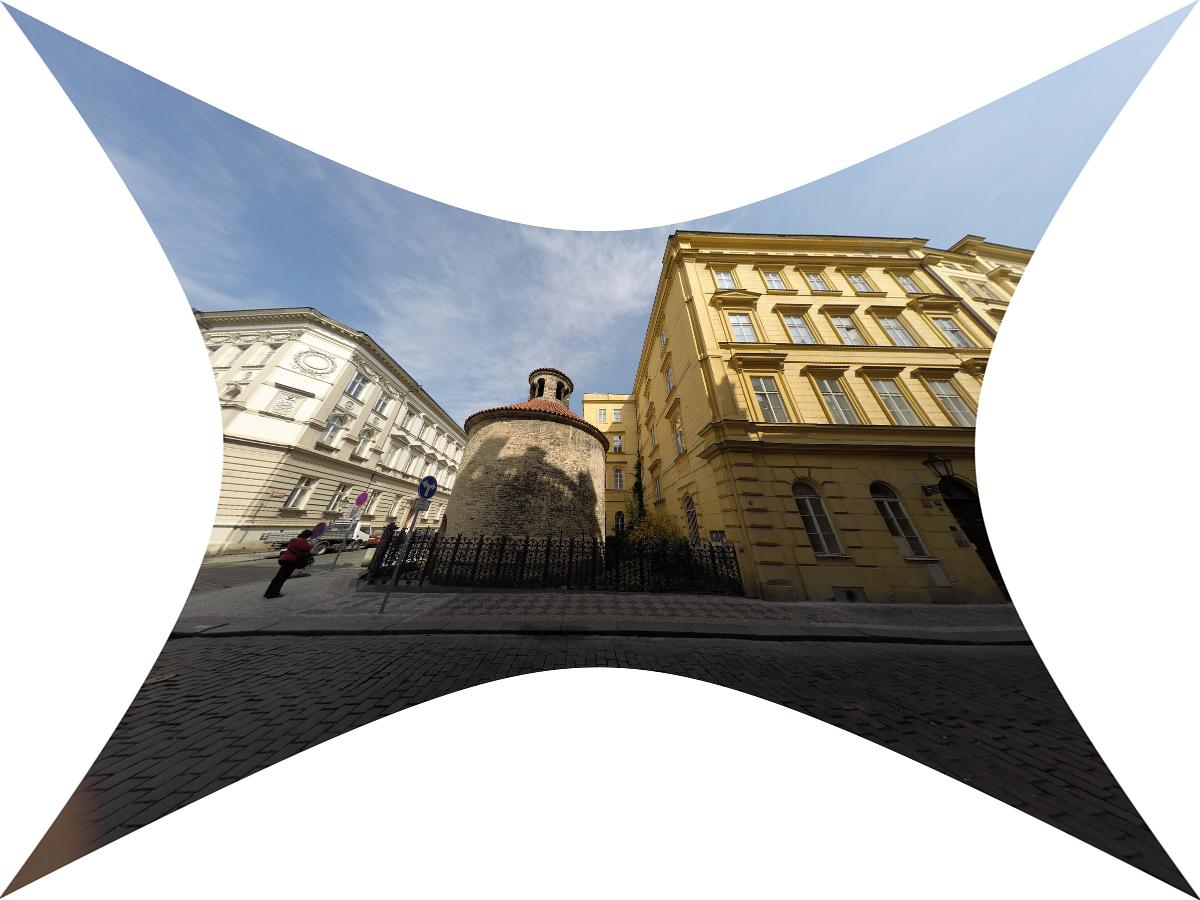}
  } 
  \subfloat{
    \includegraphics[width=0.24\textwidth]{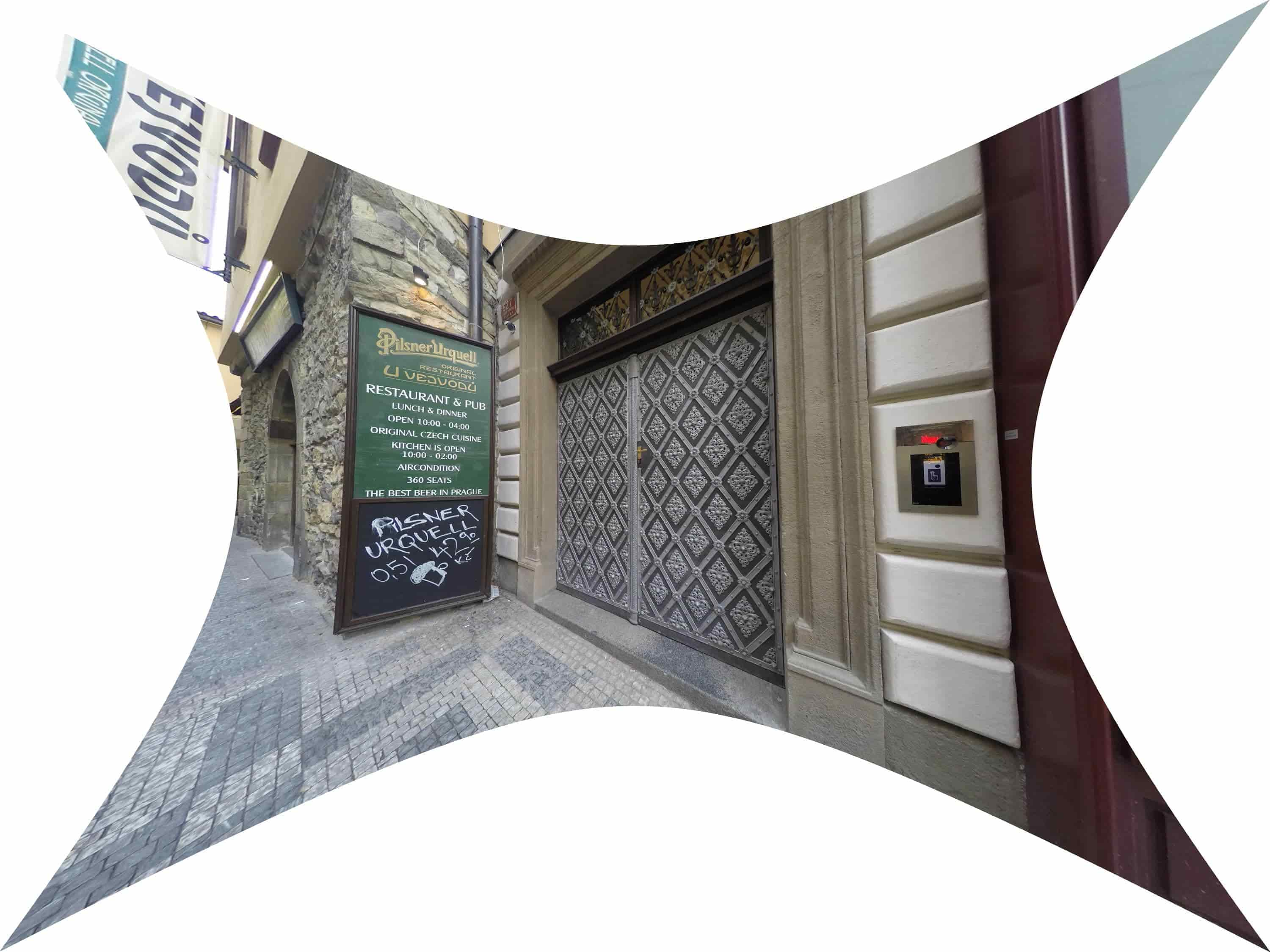}
  }
  \subfloat{
    \includegraphics[width=0.128\textwidth]{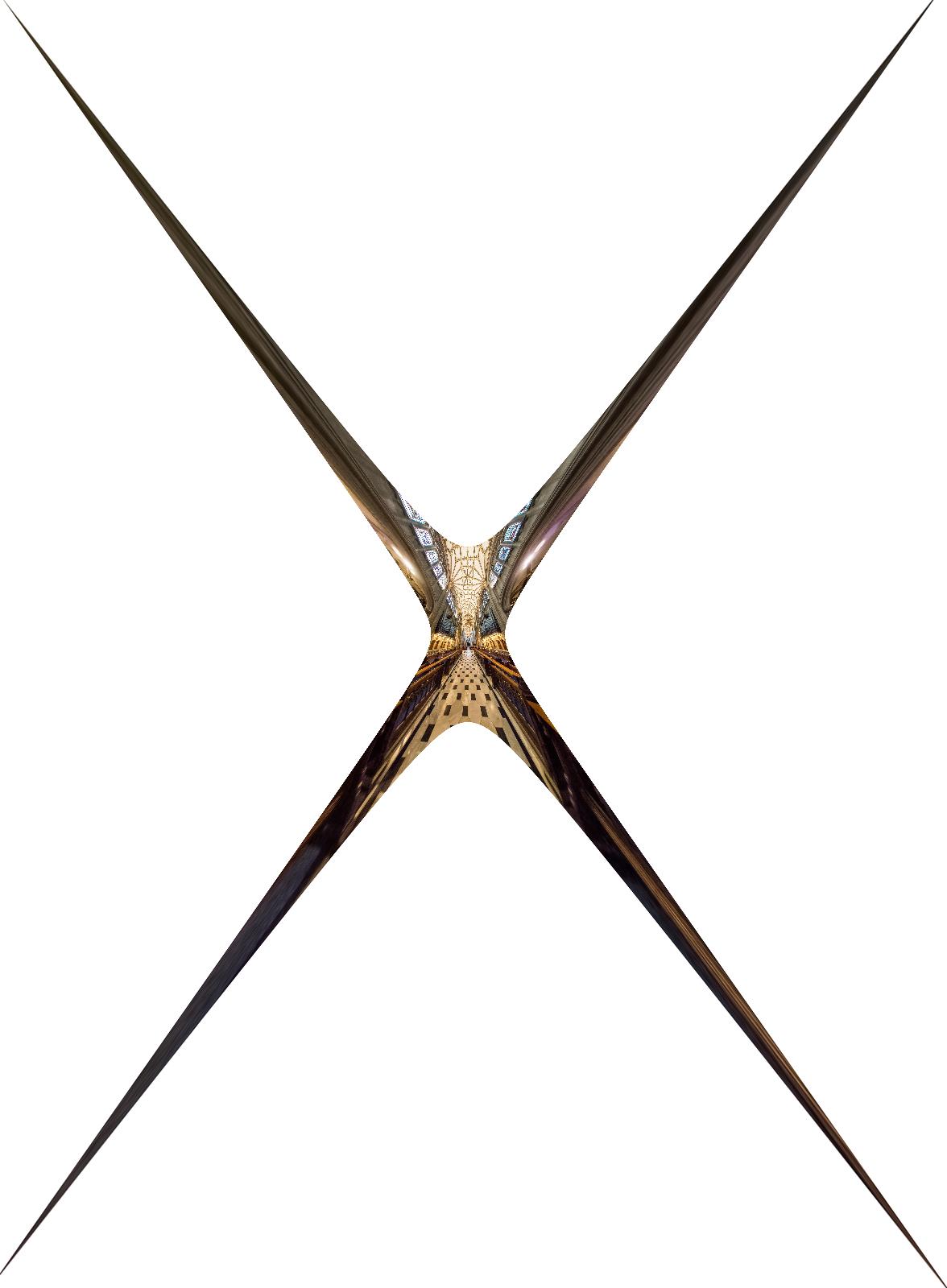}
  }
  \\[-1.5ex]
  \subfloat{
    \includegraphics[width=0.135\textwidth,trim={15cm 6cm 10cm 8cm},clip]{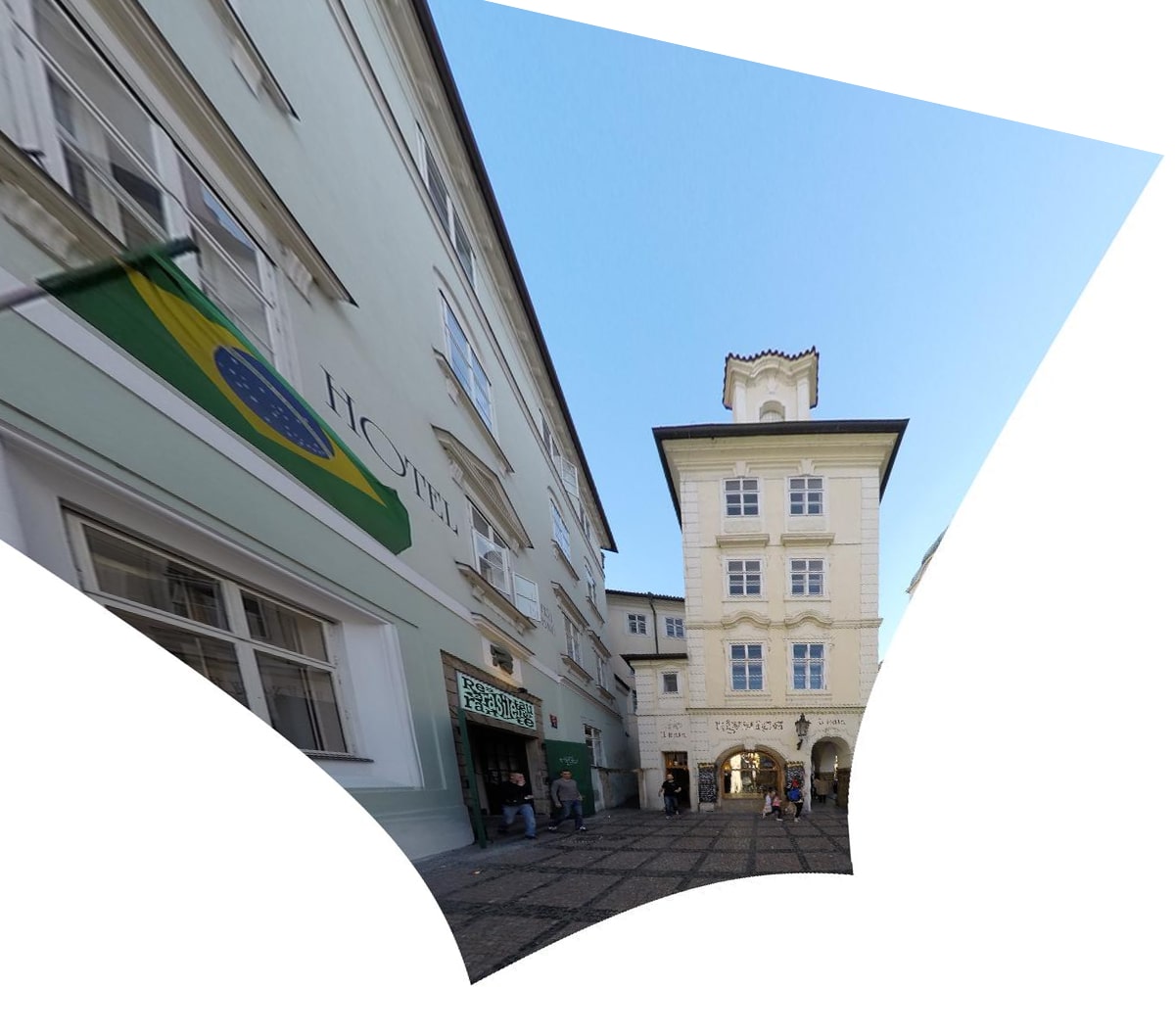}
  }
  \subfloat{
    \includegraphics[width=0.24\textwidth,trim={0cm 9cm 10cm 19cm},clip]{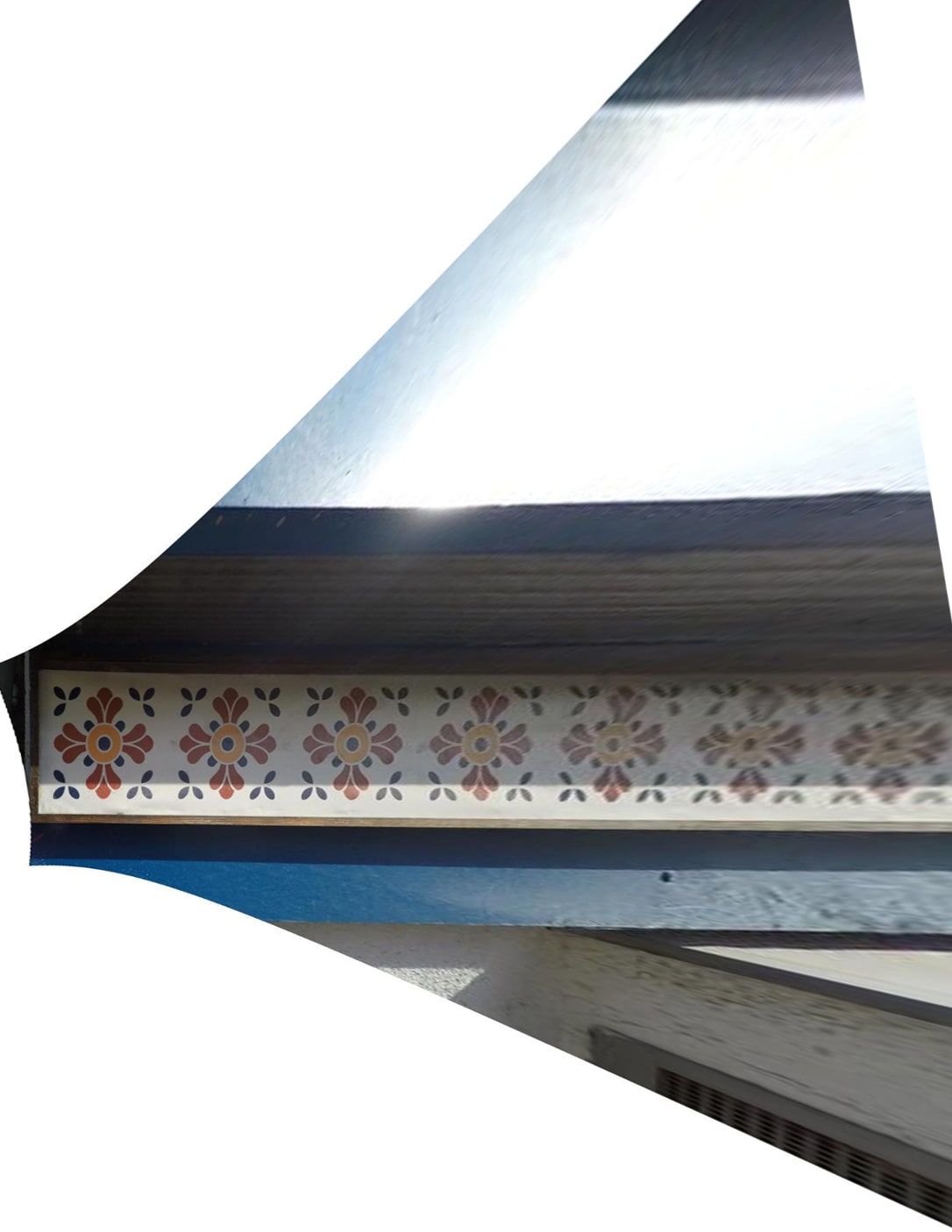}
  }
  \subfloat{
    \includegraphics[width=0.24\textwidth,trim={5cm 0cm 5cm 0cm},clip]{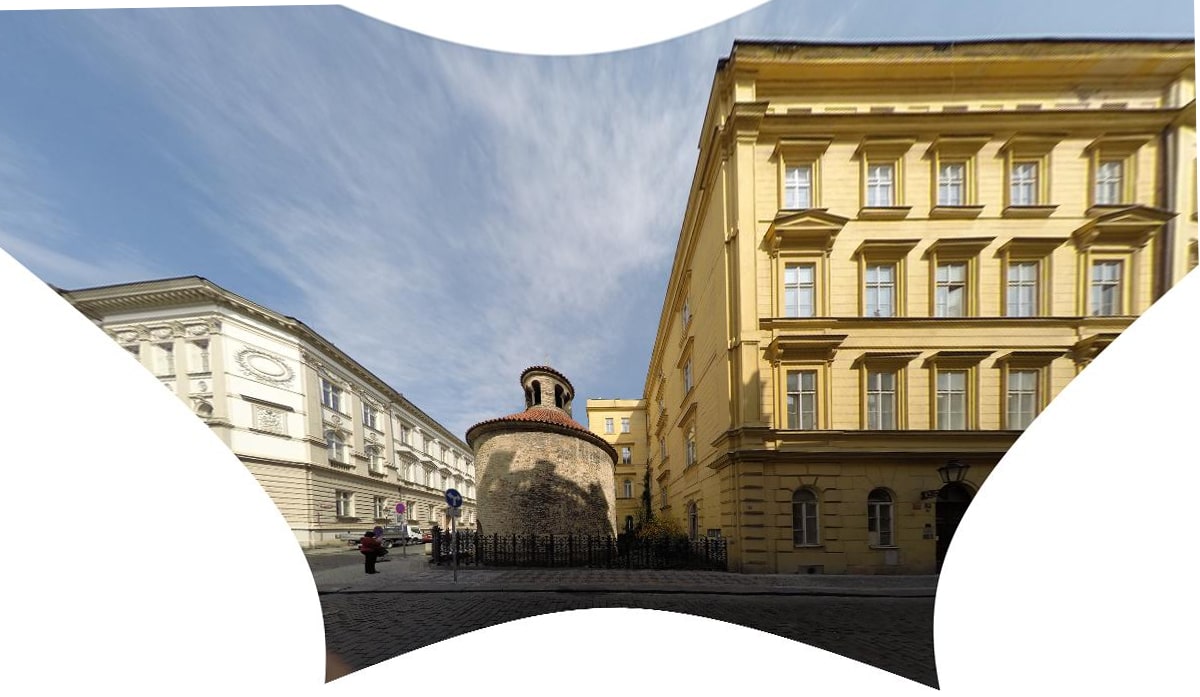}
  } 
  \subfloat{
    \includegraphics[width=0.24\textwidth,trim={1.7cm 29cm 0cm 27cm},clip]{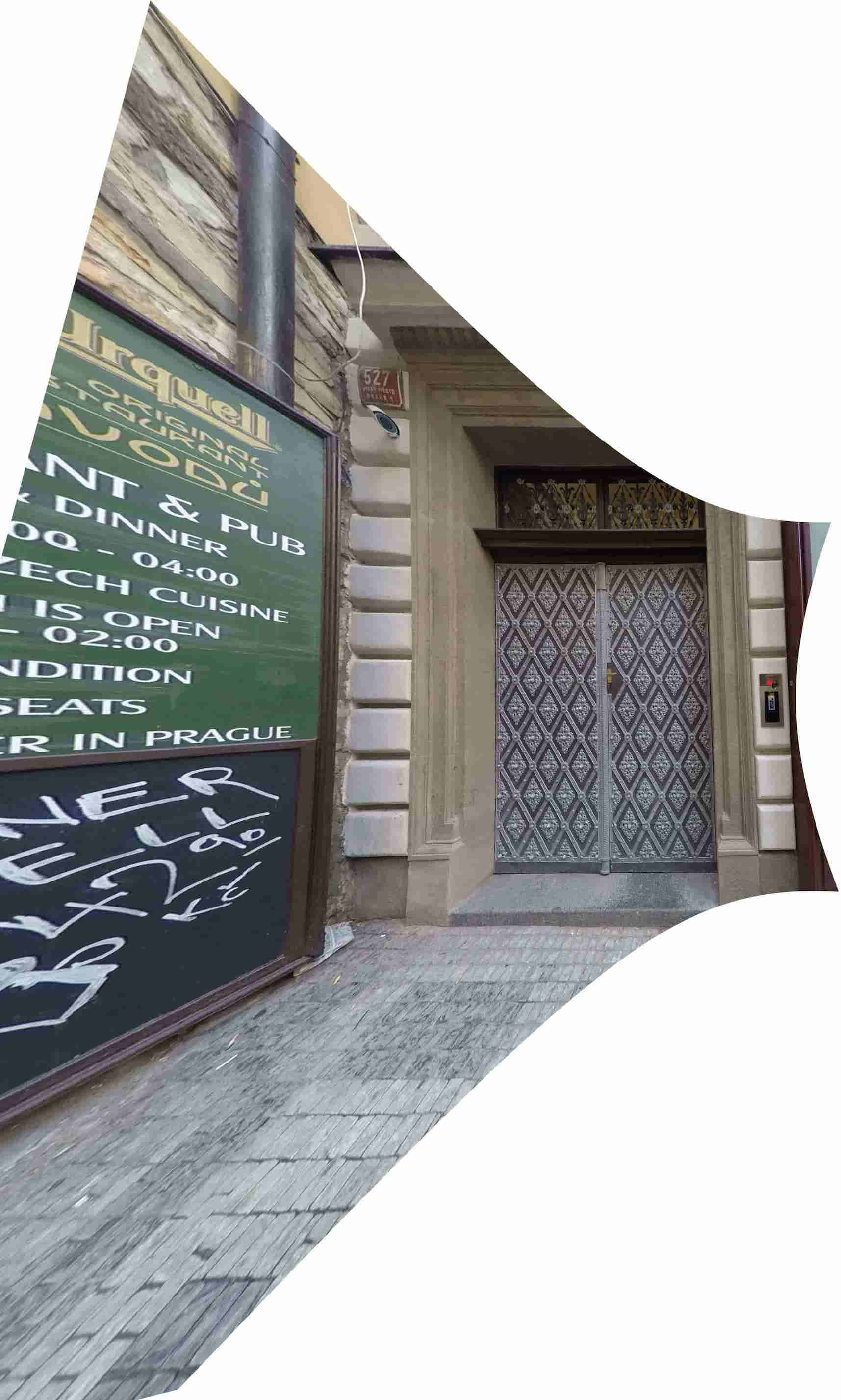}
  }
  \subfloat{
    \includegraphics[width=0.128\textwidth,trim={10cm 1cm 10cm 9cm},clip]{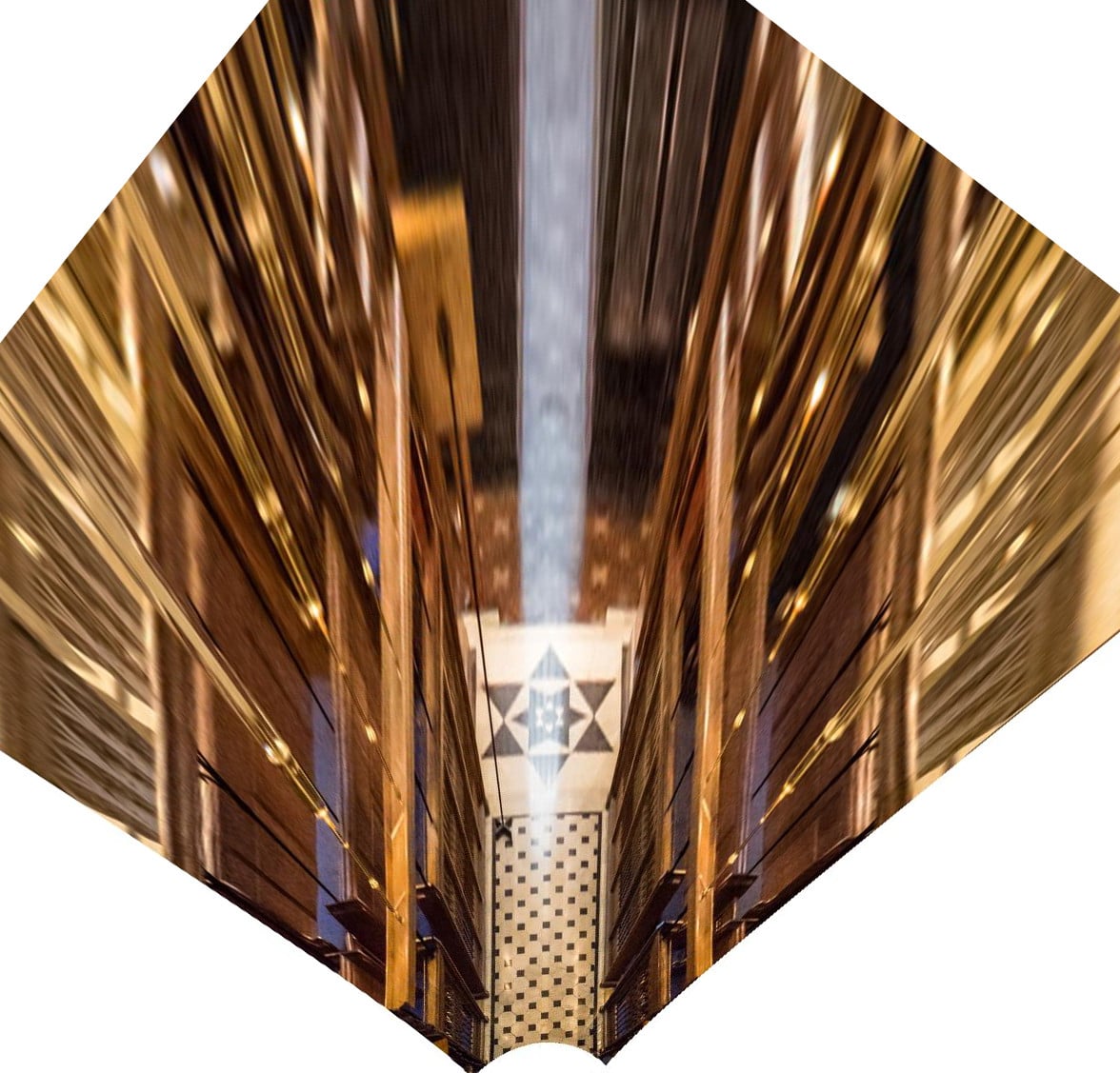}
  }
  \caption[fieldofview]{The proposed solvers give accurate
    undistortions and rectifications across all fields-of-view. The
    distorted image of the vanishing line is rendered in
    green. Left-to-right with increasing levels of
    distortion: \begin{enumerate*}[(a)] \item GoPro Hero 4 at the
      medium-FOV setting, \item GoPro Hero 4 at the wide-FOV
      setting, \item and a Samyang 7.5mm fisheye lens.\end{enumerate*}
    The outputs are the undistorted (middle row) and rectified images
    (bottom row). Note the stability of the undistortion estimates for
    the GoPro images. The rotunda image is rectified from features
    extracted mostly from the wrought iron fence below the
    rotunda. Focal lengths are 35mm equivalents.}
  \label{fig:field_of_view}
  \vspace{-8pt}
\end{figure*}

The image of a scene plane's vanishing line is denoted
$\vl=\rowvec{3}{l_1}{l_2}{l_3}^{\T}$ and the line at infinity is
$\vlinf=\rowvec{3}{0}{0}{1}^{\T}$. The phrase \emph{vanishing
  translation direction} is motivated by the fact that all imaged
scene point correspondences translating in the same direction meet at
a vanishing point.  A vanishing point is denoted by either \vu or \vv
and the vanishing translation directions of scene-plane translations
\vU or \vV as imaged by \mP, respectively. Matrices are in typewriter
font; \eg, a homography is \mH and a conjugate translation (also a
homography) with vanishing point \vu is denoted \mHu (see
\secref{sec:conjugate_translations}).

In general, a point correspondence \xcspond is two points \vx and \vxp
that are related by some geometric transformation. A covariant region
detection (see \secref{sec:acregions}) is a distorted function of some
region from the pinhole image and is denoted \rgnd. Likewise, a
distorted point extracted from a region detection is denoted
$\vxd=\rowvec{3}{\xd}{\yd}{1}^{\T}$. The
affine-rectified images of homogeneous points and regions are
denoted as $\vxr=\rowvec{3}{\xr}{\yr}{1}^{\T}$ and \rgnr,
respectively. \tabref{tab:common_denotations} summarizes this
notation.


\subsection{The Hidden Variable Trick}
\label{sec:hidden_variable_trick}
Each of the proposed solvers uses the hidden variable trick to
transform its polynomial constraint equations into a tractable
form. The hidden variable trick is a resultant technique in algebraic
geometry that is used to eliminate subsets of variables from
multivariate polynomial systems of equations
\cite{Cox-BOOK05}. Suppose that a multivariate polynomial system of
$m$ equations in $n$ unknowns is given. The hidden variable trick
works by assuming that a set $\xi_i$ of $k < n$ unknowns are
parameters that are used to construct a coefficient matrix
$\ma{M}(\xi_1,\ldots,\xi_k) \in \R[m \times l]$, such that the system
can be rewritten as
\begin{equation}
  \label{eq:hidden_variable_trick}
  \ma{M}(\xi_1,\ldots,\xi_k)\ve[y] = \ve[0],
\end{equation}
where $\ve[y] \in \R[l]$ is a vector of $l$ monomials in the remaining
$n-k$ unknowns (\ie, monomials of unknowns not appearing in $\ma{M}$).
A nontrivial solution to \eqref{eq:hidden_variable_trick} exists only
if \ma{M} is rank-deficient. The problem has been simplified since the
$n-k$ unknowns in \ve[y] are eliminated from solving $\det
\ma{M}=0$. In the case where $k > 1$, the $l \times l$ minors of
\ma{M} can be used to generate the necessary number of polynomial
constraint equations to solve for the unknowns
\buildset{\xi_1,\ldots,\xi_k}{}{}.  Once $\xi_i$ are recovered, the
original system of equations \eqref{eq:hidden_variable_trick} can be
solved for \ve[y] by back substitution.

\subsection{Solving Systems of Polynomial Equations}
The polynomial systems of equations encoding the rectifying
constraints for the Eliminated Vanishing Point (\EVP) solvers are
solved using an algebraic method based on \Gbs.  Automated solver
generators using the \Gb method~\cite{Kukelova-ECCV08,Larsson-CVPR17}
have been used to generate solvers for several camera geometry
estimation problems (see e.g.~\cite{Kukelova-ECCV08,Kukelova-CVPR15,Larsson-CVPR17,Larsson-ICCV17,Pritts-CVPR18,Pritts-ACCV18,Pritts-IJCV20}). However,
the straightforward application of automated solver generators to the
proposed constraints resulted in unstable solvers. In \cite{Larsson-ICCV17}, Larsson \etal
 proposed a method for creating polynomial solvers
for problems with unwanted solutions using ideal saturation. We use the hidden variable
trick together with ideal saturation to eliminate unknowns from the
polynomial system of equations arising in the formulations of the
EVP solvers. This results in significantly
more numerically stable solvers compared to the solvers generated directly from the
original constraint equations. For more details about \Gbs and how they are used in polynomial solvers we refer the interested reader to \cite{Larsson-ICCV17, Larsson-CVPR17, Kukelova-ECCV08, Cox-BOOK05}.


\subsection{Solver Naming Convention}
\label{sec:naming_convention}
We apply the solver naming convention of Pritts \etal
\cite{Pritts-ACCV18,Pritts-IJCV20} to the proposed and
state-of-the-art solvers evaluated in this paper. The minimal
configuration of region correspondences is given as the subscript to
\mH (denoting a homography); e.g., a solver requiring 3
affine-covariant region correspondences is denoted $\mH_{222}$.
The unknowns that are recovered by the solver are suffixed to
$\mH_{\cdot}$, \eg, the proposed solver requiring one region
correspondence and returning the vanishing line \vl and division model
parameter $\lambda$ of lens distortion is denoted \rgntwordctevl.

\section{Problem Formulation}
\label{sec:formulation}
An affine-rectifying homography \mH transforms the image of the scene
plane's vanishing line $\vl = \rowvec{3}{l_1}{l_2}{l_3}^{\T}$ to
the line at infinity $\vlinf=\rowvec{3}{0}{0}{1}^{\T}$
\cite{Hartley-BOOK04}. Thus any homography $\mH$ satisfying the
constraint
\begin{equation}
  \label{eq:vline_constraint}
  \eta \vl = \ma{H}^{\T} \vlinf =
  \begin{bmatrix}  \ve[h]_1 & \ve[h]_2 & \ve[h]_3 \end{bmatrix}\colvec{3}{0}{0}{1}, \quad \text{$\eta \neq 0$,}
\end{equation}
and where \vl is an imaged scene plane's vanishing line, is an
affine-rectifying homography.  Constraint \eqref{eq:vline_constraint}
implies that $\ve[h]_3=\vl$, and that the image of the line at
infinity is independent of rows $\ve[h]^{\T}_1$ and $\ve[h]^{\T}_2$ of
\mH. Thus, assuming $l_3 \neq 0$ \cite{Hartley-BOOK04}, the
affine-rectification of image point \vx to the affine-rectified point
\vxr can be defined as
\begin{equation}
  \label{eq:recthg}
  \begin{split}
    \alpha \vxr = &\rowvec{3}{\alpha \xr}{\alpha \yr}{\alpha}^{\T}  = \mH(\vl) \vx  \\
    & \textup{s.t.}  \quad
    \mH(\vl) =
    \begin{bmatrix}
      1 & 0 & 0 \\
      0 & 1 & 0 \\
      & \vl^{\T} &
    \end{bmatrix}
    \quad \text{and} \quad \alpha \neq 0.
  \end{split}
\end{equation}

\subsection{Radial Lens Undistortion}
\label{sec:radial_lens_distortion}
Affine rectification as given in \eqref{eq:recthg} is valid only if
$\vx$ is imaged by a pinhole camera. Cameras always have some lens
distortion, and the distortion can be significant for wide-angle
lenses. For a lens distorted point, denoted $\vxd$, an undistortion
function $f$ is needed to transform $\vxd$ to the pinhole point
$\vx$. We use the one-parameter division model to parameterize the
radial lens undistortion function,
\begin{equation}
\gamma\vx =
f(\vxd,\lambda)=\rowvec{3}{\xd}{\yd}{1+\lambda(\xd^2+\yd^2)}^{\T}
\label{eq:division_model} 
\end{equation}
where $\vxd=\rowvec{3}{\xd}{\yd}{1}^{\T}$ is a feature point with the
distortion center subtracted. 

The strengths of this model were shown by
Fitzgibbon~\cite{Fitzgibbon-CVPR01} for the joint estimation of
two-view geometry and non-linear lens distortion.  The division model
is especially suited for minimal solvers since it is able to express a
wide range of distortions (\eg, see second row of
\figref{fig:field_of_view}) with a single parameter (denoted
$\lambda$), as well as yielding simpler equations compared to other
distortion models.

For the remainder of the derivations, we assume that the image center
and distortion center are coincident and that \vxd is a
distortion-center subtracted point. While this may seem like a strong
assumption, Willson~\etal~\cite{Willson-OPTSOC94} and
Fitzgibbon~\cite{Fitzgibbon-CVPR01} showed that the precise
positioning of the distortion center does not strongly affect image
correction. Furthermore, we will see in the experiments in
\secref{sec:experiments} that the proposed method is robust to
deviations in the distortion center. Importantly, no constraints are
placed on the location of the principal point of the camera by these
assumptions, which is an influential calibration parameter
\cite{Willson-OPTSOC94}. However, the choice to fix the distortion
center at the image center does make it difficult to remove a modeling
degeneracy at the image center, which is discussed in detail in
\secref{sec:pencil_through_origin}.

Affine rectified points \vxr[i] can be expressed in terms of distorted
points \vxd[i] by substituting \eqref{eq:division_model} into
\eqref{eq:recthg}, which gives
\begin{equation}
  \label{eq:udrect}
  \begin{split}
    \alpha\vxr & = \rowvec{3}{\alpha \xr}{\alpha \yr}{\alpha}^{\T} = \mH(\vl) f(\vxd,\lambda) = \\ 
    & \rowvec{3}{\xd}{\yd}{l_1\xd+l_2\yd+l_3(1+\lambda(\xd^2+\yd^2))}^{\T}. 
  \end{split}
\end{equation}
Interestingly, the rectifying function $\mH(\vl) f(\vxd,\lambda)$ in
\eqref{eq:udrect} also acts radially about the distortion center, but
unlike the division model in \eqref{eq:division_model}, it is not
rotationally symmetric.

The distortion function of the lens as parameterized by the division
model is denoted $f^{d}(\cdot,\lambda)$. Under the division model, the
radially-distorted image of the vanishing line is a circle and is
denoted \vld
\cite{Bukhari-JMIV13,Fitzgibbon-CVPR01,Strand-BMVC05,Wang-JMIV09}. \figsref{fig:field_of_view}
and \ref{fig:evl_results} render the distorted vanishing line in the
source images, which affirm the accuracy of the rectifications by the
proposed solvers.

\subsection{Covariant Region Parameterization}
\label{sec:region_parameterization}
Covariant region detections reduce the number of required
correspondences to as few as one for the proposed solvers, but corners
or combinations of corners and covariant regions can also be used as
input.
Since the proposed solvers are derived from constraints
induced by point correspondences, points are extracted from the region
correspondences as input to the proposed solvers. The geometry of an
affine-covariant region \rgn is given by a right-handed affine basis in the image coordinate system called a \emph{local affine frame} (LAF). The affine frame is
minimally parameterized by three points
$\{\,\ve[o],\,\vx,\,\ve[y]\,\}$. For similarity-covariant regions,
there is the additional constraint that $\vx-\ve[o] \perp
\ve[y]-\ve[o]$ (see \cite{Vedaldi-SOFTWARE08}). This construction is
also referred to as an \emph{oriented circle}, where \ve[o] is the origin of the circle and \vx defines
the circle's orientation and radius. Similarity-covariant regions are
minimally parameterized by two points. Examples of both frame
constructions are shown in \figsref{fig:ct_composite_rect},
\ref{fig:input_configurations}, and \ref{fig:evl_geometry}, and an
example of affine frames constructed from the combined methods of
\cite{Matas-BMVC02,Matas-ICPR02,Obdrzalek-BMVC02} are shown in
\figref{fig:local_features}.

\section{Conjugate Translations}
\label{sec:conjugate_translations}
Assume that the scene plane $\Pi$ and a camera's image plane $\pi$ are
related point-wise by the camera $\mP$ so that $\alpha \vxp=\mP\vXp$, where
$\alpha$ is a non-zero scalar, $\vXp \in \Pi$ and $\vxp \in
\pi$. Furthermore, let $\vX$ and $\vXp$ be two points on the scene
plane $\Pi$ such that $\vU=\vXp-\vX=\rowvec{3}{u_x}{u_y}{0}^{\T}$. By
encoding $\vU$ in the homogeneous translation matrix $\ma{T}(\vU)$, the
points $\vX$ and $\vXp$ as imaged by camera $\mP$ can be expressed as
\begin{equation}
  \begin{split}
    \label{eq:conjugate_translation}
    \alpha\vxp=&\mP\vXp=\mP\ma{T}(\vU)\vX=\mP\ma{T}(\vU)\mP^{\inv}\vx=\ma{H_{\ve[u]}}\vx \\
    & \text{s.t.} \quad \ma{T}(\vU) = \begin{pmatrix} 1 & 0 & u_x \\ 0 & 1 & u_y \\ 0 & 0 & 1 \end{pmatrix}, \\
  \end{split}
\end{equation}
where the homography $\ma{H}_{\ve[u]}=\mP\ma{T}(\vU)\mP^{\inv}$ is
called a conjugate translation because of the form of its matrix
decomposition, and points \vx and \vxp are in correspondence (denoted
\xcspond) with respect to the conjugate translation \mHu
\cite{Hartley-BOOK04,Schaffalitzky-BMVC98}.

\begin{figure*}[t!]
  \centering
  \resizebox{0.9\textwidth}{!}{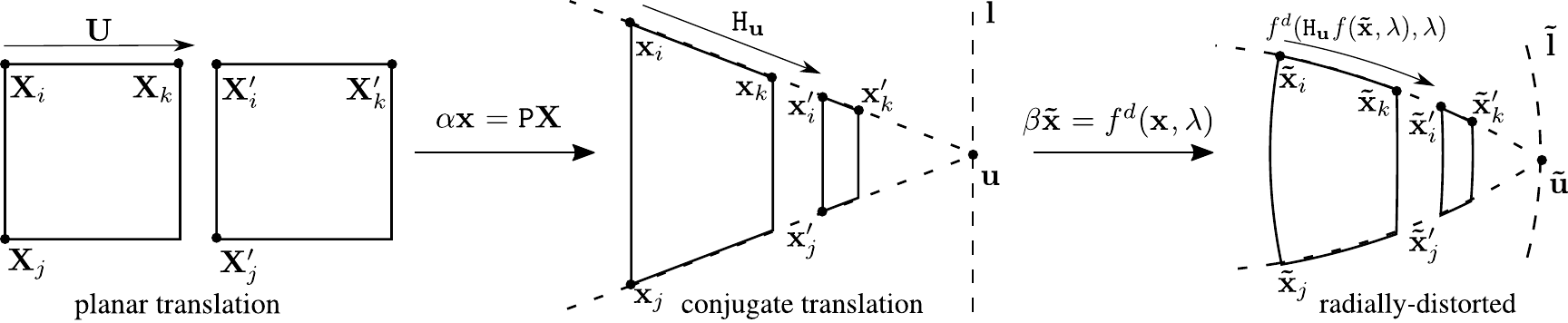}
  \caption{The Geometry of a Radially-Distorted Conjugate
      Translations. A translation of coplanar scene points
    \buildset{ \vX[i], \vX[j], \vX[k] }{}{} by \vU induces a
    conjugate translation \mHu in the undistorted image as viewed by
    camera \mP. Joined conjugately-translated point correspondences
    \xcspond[i], \xcspond[j] and \xcspond[k] must meet at the
    vanishing point \vu. Vanishing line \vl is the set of all
    vanishing points of translation directions. The division model
    images lines as circles, thus the distorted vanishing point
    \ve[\tilde{u}] is given by the intersection of three circles,
    two of which are coincident with the radially-distorted
    conjugately-translated point correspondences
    \xdcspond[i],\xdcspond[j] and \xdcspond[k], and the third is
    given by the distorted vanishing line \vld. Radially-distorted
    conjugately-translated points are related by $f^{d}(\mHu
    f(\vxd,\lambda),\lambda)$, where $f^{d}(\cdot,\lambda)$ is the
    division-model distortion function.}
  \label{fig:conjugate_translations}
\end{figure*}

Decomposing $\ma{H}_{\ve[u]}$ into its projective components gives
\begin{equation}
  \label{eq:decomposition}
  \begin{split}
    \alpha \vxp &= \mHu\vx = \left[
      \mP\ma{I}_3\mP^{\inv}+\mP\colvec{3}{u_x}{u_y}{0}\left[\mP^{-\T}\colvec{3}{0}{0}{1}\right]^{\T}
      \right] \vx \\ &= [\ma{I}_3+s^{\vu}\ve[u]\ve[l]^{\T}] \vx
  \end{split}
\end{equation}
where $\ma{I}_3$ is the $3\times 3$ identity matrix, and, also
consulting \figref{fig:conjugate_translations} to relate the unknowns
to the geometry,
\begin{itemize} 
\itemsep0em
\item line \vl is the imaged scene plane's vanishing line,
\item point \vu is the vanishing point of the translation direction,
\item and scalar $s^{\vu}$ is the magnitude of translation in the
  direction $\vu$ for the point correspondence \xdcspond \cite{Schaffalitzky-BMVC98}.
\end{itemize}

\subsection{Meets of Joins}
\label{sec:meets_of_joins}
Let $\ve[m]_i$ be the join of the conjugately translated point
correspondence \xcspond[i] Then $\ve[m]_i$ can be expressed in terms
of the camera matrix \mP, joined scene point correspondences
\Xcspond[i], and scene translation direction \vU as
\begin{equation}
  \begin{split}
    \label{eq:join_of_image_points}
    & \alpha \ve[m]_i = \alpha \left(\vx[i] \times \vxp[i]\right) =
    (\mP \vX[i] \times \mP \vXp[i]) / |\mP| = \\ & \quad (\mP\vX[i]
    \times \mP(\vX[i]+\vU))/|\mP| = \mP^{\invT} (\vX[i]+\vU),
  \end{split}
\end{equation}
where $\alpha \neq 0$ and $|\mP| = \det \mP$. 

Using \eqref{eq:join_of_image_points} to express the meet of joins
$\ve[m]_i$ and $\ve[m]_j$ in terms of the camera \mP and joined scene
point correspondences \Xcspond[i] and \Xcspond[j] gives
\begin{equation}
  \begin{split}
    \label{eq:join_meet_meet_vp}
    \alpha_i &\ve[m]_i \times \alpha_j \ve[m]_j = \left(\mP^{\invT}
    (\vX[i]+\vU)\right) \times \left(\mP^{\invT}
    (\vX[j]+\vU)\right) = \\
    & \mP ((\vX[i]+\vU) \times (\vX[j]+\vU)) / |\mP| = \\
    & \mP \left(\vU^{\T}(\vX[i]\times\vX[j])\right)\vU / |\mP|= \beta \mP\vU  = \eta \vu, \\
  \end{split}
\end{equation}
where $\beta=\ve[U]^{\T}(\vX[i] \times \vX[j])/|\mP|$, $\eta$ is
non-zero and $\vU^{\T}(\vX[i]\times\vX[j])$ is non-zero for
non-degenerate point configurations (see
\figref{fig:conjugate_translations}). In general
\eqref{eq:join_meet_meet_vp} shows that the image of all joined scene
point correspondences translating in the same direction meet at the
vanishing point of their translation direction, \ie
$\eta\vu=\beta\ma{P}\vU$. Note that if correspondence \xcspond[k] from
\figref{fig:conjugate_translations} were used in lieu of \xcspond[j]
in \eqref{eq:join_meet_meet_vp}, then
$\vU^{\T}(\vX[i]\times\vX[k])=0$, which implies that $\eta=0$. This is
a degenerate configuration of the solvers and is discussed in detail
in \secref{sec:pami19_degeneracies}.

Since \vU is coincident with \linf by construction (see
\figref{fig:conjugate_translations}) and point-line incidence is
invariant under projection by \mP \cite{Hartley-BOOK04}, \vu and \vl
are also coincident,
\begin{equation}
  \label{eq:vp_vl_incidence}
  \vl^{\T}\vu = 0.
\end{equation}

The \EVL solver introduced in \ref{sec:evl_constraints} uses the
relation between conjugately-translated points and vanishing points
derived in \eqref{eq:join_of_image_points} and
\eqref{eq:join_meet_meet_vp} and the vanishing point-vanishing line
incidence equation of \eqref{eq:vp_vl_incidence} to place constraints
on \vl.

\subsection{Radially-Distorted Conjugate Translations}
\label{sec:rdct}
Conjugate translations as defined in \eqref{eq:decomposition} can be
written in terms of radially-distorted conjugately-translated point
correspondences undistorted by \eqref{eq:division_model} as
\begin{equation}
  \label{eq:distorted_conjugate_translation}
  \alpha f(\vxdp,\lambda) = \mHu f(\vxd,\lambda) =
         [\ma{I}_3+s^{\vu}\vu\ve[l]^{\T}] f(\vxd,\lambda),
\end{equation}
\xdcspond is a
radially-distorted point correspondence that is consistent with the
conjugate translation \mHu. We call \xdcspond a
\emph{radially-distorted conjugately-translated point correspondence}
going forward.

Each of the \EVP solvers introduced in \secref{sec:evp_constraints}
uses the relation defined in
\eqref{eq:distorted_conjugate_translation} and the vanishing
point-vanishing line incidence equation of \eqref{eq:vp_vl_incidence}
to place constraints on \vl and $\lambda$.

\section{Solvers}
This section introduces five different minimal solvers for different
geometric configurations of radially-distorted conjugate translations,
which are distinguished by the number of directions and magnitudes of
translations that the proposed solver variants admit.  The designs of
the solver variants are motivated by the types of covariant feature
detectors that can be used to extract point correspondences, which
give the constraints needed to jointly solve for the division model
parameter of lens undistortion, vanishing line and the vanishing point
of the translation direction(s).

Each of the proposed minimal solvers exploits the following properties
of radially-distorted conjugate translations:
\begin{enumerate*}[(i)] \item The affine-rectified image
of the meet of the joins of conjugately-translated point
correspondences is on the line at infinity, and \item a
radially-distorted conjugate translation is a transformation with
exactly four degrees of freedom.
\end{enumerate*}

The proposed solvers can also be differentiated by the choice to use
the hidden variable trick to either eliminate the unknown parameters
of the vanishing point of the imaged translation direction or the
imaged scene plane's vanishing line from the solver's polynomial
system of equations \cite{Cox-BOOK05}. The solver groups are
eponymously named after their eliminated
unknowns. 

\begin{table*}[t!]
  \centering
  \caption{Proposed Solvers (shaded in grey) vs. State of the Art}
  \vspace{-5pt}
  \ra{1}
   \resizebox{\textwidth}{!} {
    \begin{threeparttable}
      \begin{tabular}{@{} R{15ex}C{15ex}C{10ex}C{10ex}C{15ex}C{10ex}C{10ex}C{15ex}C{15ex} @{} }
        \toprule
        &  &  &  &  &  \multicolumn{2}{c}{\# Correspondences} &   &  \\
        & Reference & Rectifies & Undistorts & Motion & Regions &  Points & \# Solutions & Size \\
        \midrule
        \rgntwoct & \cite{Schaffalitzky-BMVC98} & \checkmark & & translation & $1$ & $2$ & $1$ & closed form  \\
        \rgntwotwoct & \cite{Chum-ACCV10} & \checkmark & & rigid & $2$ & N/A & $1$ & closed form  \\
        \rowcolor{Gray}
        \rgntwordctevl & & \checkmark & \checkmark & translation & $1$ & $3$ & $4$ & closed form \\
        \rowcolor{Gray}
        \rgntwordctevp & & \checkmark & \checkmark  & translation & $1$ & $3$ & $4$ & $14 \times 18$  \\
        \rowcolor{Gray}
        \rgntwordsctevp & & \checkmark & \checkmark  & translation & $1$ & $3$ & $2$ & $24 \times 26$  \\
        \rowcolor{Gray}
        \rgntwotwordctevp & & \checkmark & \checkmark & translation & $2$ & $4$ & $6$ & $54 \times 60$  \\
        \rowcolor{Gray}
        \rgntwotwordsctevp & & \checkmark & \checkmark & translation & $2$ & $4$ & $4$ & $76 \times 80$  \\
        \rgntwotwofitz & \cite{Fitzgibbon-CVPR01} &  & \checkmark & rigid\tnote{1} & $2$ & $5$ & $18$ & $18 \times 18$   \\ [1.5pt]
        \rgntwotwokukelova & \cite{Kukelova-CVPR15} &  & \checkmark & rigid\tnote{1} & $2$ & $5$ & $5$ & $16 \times 21$   \\ [1.5pt]     
        \rgntwotwotwodes & \cite{Pritts-ACCV18,Pritts-IJCV20} & \checkmark & \checkmark & rigid & $3$ & $9$ & $54$ & $133 \times 187$  \\ [1.5pt]
        \bottomrule
      \end{tabular}
      \begin{tablenotes}
      \item[1] The preimages of both region correspondences must be
        related by the same rigid transform in the scene plane.
      \end{tablenotes}
    \end{threeparttable}
   }
\tabcap{The proposed solvers require a few as 1 region
    correspondence instead of three and are significantly simpler than the
    undistorting and rectifying solver \rgntwotwotwodes of
    \cite{Pritts-ACCV18,Pritts-IJCV20}. The homography solvers of
    \cite{Fitzgibbon-CVPR01,Kukelova-CVPR15} do not directly recover
    the vanishing line and require two affine-covariant region
    correspondences or five points, all of which have the same relative
    orientation, which restricts sampling.}
  \label{tab:solver_properties}
\end{table*}

The group of Eliminated Vanishing Point (\EVP) solvers hide the lens
undistortion parameter and vanishing line parameters and have the
vanishing point eliminated. They provide flexible sampling in a
\RANSAC-based estimator: they can jointly recover undistortion and
rectification from radially-distorted conjugate-translations in one or
two directions, where some of the point correspondences can translate
with arbitrary distance (see \figref{fig:input_configurations}). In
addition, there is an \EVP variant that admits reflections.

The Eliminated Vanishing Line (\EVL) solver hides the lens
undistortion parameter and eliminates the vanishing line parameters
(the vanishing points are recovered by construction).  The \EVL
solver jointly recovers undistortion and rectification from
radially-distorted conjugately-translated point correspondences in one
direction. The elimination of the vanishing line results in a solver
that is exceptionally fast, stable and robust to feature noise.

\tabref{tab:solver_properties} summarizes the geometric assumptions,
inputs and complexity of the proposed solvers with respect to the
state of the art. From the undistorting solvers, the proposed solvers
require the fewest correspondences, and are much simpler than the
undistorting and rectifying solver of \cite{Pritts-ACCV18}.

The following sections show significant differences between the two
groups with respect to solver complexity, time to solution, stability
and noise sensitivity. The following sections describe how the solvers
are generated, and, in particular, detail how either the vanishing
point of the translation direction or the vanishing line is eliminated
to simplify the systems of polynomial equations that arise from
constraints induced by radially-distorted conjugately-translated local
features.


\subsection{The Eliminated Vanishing Point (EVP) Solvers}
\label{sec:evp_constraints}
The model for radially-distorted conjugate translations in
\eqref{eq:distorted_conjugate_translation} defines the unknown
geometric quantities:
\begin{enumerate*}[(i)]
\item division-model parameter $\lambda$, \item imaged scene-plane
  vanishing line $\ve[l]=\rowvec{3}{l_1}{l_2}{l_3}^{\T}$,
\item vanishing point of the translation direction
  $\vu=\rowvec{3}{u_1}{u_2}{u_3}^{\T}$,
\item scale of translation $s^{\vu}$ for correspondence \xdcspond,
\item and the homogeneous scale parameter $\alpha$.
\end{enumerate*}

The solution for the vanishing line \vl is constrained to the affine
subspace $l_3=1$ of the real-projective plane, which makes it unique.
This inhomogeneous choice of \vl is unable to represent the pencil of
lines that pass through the image origin; however, the degeneracy
remains even with a homogeneous representation of \vl. See
\secref{sec:pami19_degeneracies} for a more detailed discussion of the
degeneracies.

The vanishing direction \vu must meet the vanishing line \vl, which
defines a subspace of solutions for \vu. The magnitude of \vu is set
to the magnitude of conjugate translation $s^{\vu}_1$ of the first
correspondence \xdcspond[1], which defines a unique solution
\begin{equation}
\label{eq:orthogonality_constraint}
\vl^{\T}\vu=l_1u_1+l_2u_2+u_3=0 \quad \wedge \quad \|\vu\|=s^{\vu}_1.
\end{equation}
The relative scale of translation $\bar{s}^{\vu}_i$ for each
correspondence \xdcspond[i] with respect to the magnitude of $\|\vu\|$
is defined so that $\bar{s}^{\vu}_{i}={s^{\vu}_{i}}/{\|\vu\|}$. Note
that $\bar{s}^{\vu}_{1}=1$. The relationship between magnitude of
translation in the scene plane and the magnitude of conjugate
translation is derived in the \secref{sec:xfer_error} in
the supplemental materials.

Two \emph{one-direction solvers} are proposed, which require 3
radially-distorted conjugately-translated point correspondences.  A
radially-distorted conjugately-translated affine-covariant region
correspondence provides the necessary 3 point correspondences. Solver
\rgntwordctevp assumes that all point correspondences have the same
relative scales of translation, i.e.\ $\bar{s}^{\vu}_{1} =
\bar{s}^{\vu}_{2} = \bar{s}^{\vu}_{3} = 1$. Solver \rgntwordsctevp
relaxes the equal relative scale of translation assumption of the \rgntwordctevp
solver. In particular, solver \rgntwordsctevp assumes that two of the
point correspondences have the same magnitude of conjugate translation
(i.e.\ $\bar{s}^{\vu}_{1} = \bar{s}^{\vu}_{2} = 1$), and the third
point correspondence has an unknown relative scale of the translation
$\bar{s}^{\vu}_{3}$. The \rgntwordsctevp admits combinations of
similarity-covariant regions (defining 2 point correspondences) and
corner detections for flexible sampling of complementary features.

In addition, two \emph{two-direction solvers} are proposed that
require 4 coplanar point correspondences, 2 of which have the
vanishing point of translation direction \vu and the remaining 2 a
different vanishing point \vv. Two similarity-covariant region
correspondences consistent with two radially-distorted conjugate
translations provide 2 pairs of 2 point correspondences and thus
provide the necessary 4 point correspondences.

Solver \rgntwotwordctevp requires four points and assumes equal
relative scales of conjugate translation in both directions, namely
$\bar{s}^{\vu}_{1} = \bar{s}^{\vu}_{2} = 1$ with respect to
$\|\vu\|=s^{\vu}_1$ and $\bar{s}^{\vv}_{3} = \bar{s}^{\vv}_{4} = 1$
with respect to $\|\vv\|=s^{\vv}_3$.

Solver \rgntwotwordsctevp requires four point correspondences,
equivalently, two similarity-covariant region correspondences, and
relaxes the assumption of the \rgntwotwordctevp solver that both point
correspondences in the \vv direction have the same magnitudes of
conjugate translation. In particular, \rgntwotwordsctevp assumes that
the first two point correspondences translate in the direction $\vu$
with the same relative scale of translation, i.e., $\bar{s}^{\vu}_{1}
= \bar{s}^{\vu}_{2} = 1$.  The remaining two point correspondences
translate in the direction $\vv$ with arbitrary translation
magnitudes, i.e., the relative scales of translations of these two
correspondences with respect to $\|\vv\|=s^{\vv}_3$ are
$\bar{s}^{\vv}_{3} = 1$ and an unknown relative scale
$\bar{s}^{\vv}_{4}$. In the case that similarity-covariant regions are
extracted from the image and its reflection, reflected covariant
regions can be used for jointly solving for undistortion and
rectification (see \figref{fig:input_configurations}).

In all of the proposed solvers the scalar values $\alpha_i$ are
eliminated from \eqref{eq:distorted_conjugate_translation}. This is
done by multiplying \eqref{eq:distorted_conjugate_translation} by the
skew-symmetric matrix $[f(\vxdp,\lambda)]_{\times}$. The fact that the
join of a point $\vx$ with itself $[\vx]_{\times}\vx$ is $\ve[0]$
gives,
\begin{equation}
  \label{eq:scalar_elimination}
  \begin{split}
    &
    \begin{bmatrix}
      0 & -\tilde{w}^{\prime}_i & \tilde{y}^{\prime}_i \\
      \tilde{w}^{\prime}_i & 0 & -\tilde{x}^{\prime}_i \\
      -\tilde{y}^{\prime}_i & \tilde{x}^{\prime}_i & 0
    \end{bmatrix} \\ & 
    \begin{bmatrix}
      1+\bar{s}^{\vu}_{i}u_1l_1 & \bar{s}^{\vu}_{i}u_1l_2
      & \bar{s}^{\vu}_{i}u_1 \\
      \bar{s}^{\vu}_{i}u_2l_1   & 1+\bar{s}^{\vu}_{i}u_2l_2 & \bar{s}^{\vu}_{i}u_2  \\
      \bar{s}^{\vu}_{i}u_3l_1   & \bar{s}^{\vu}_{i}u_3l_2   & 1+\bar{s}^{\vu}_{i}u_3
    \end{bmatrix}  
    \colvec{3}{\tilde{x}_i}{\tilde{y}_i}{\tilde{w}_i}
    = \ve[0],
  \end{split}
\end{equation}
where $\tilde{w}_i=1+\lambda(\tilde{x}^2_i+\tilde{y}^2_i)$ and
$\tilde{w}^{\prime}_i=1+\lambda(\tilde{x}^{\prime
  2}_i+\tilde{y}^{\prime 2}_i)$. The matrix equation in
\eqref{eq:scalar_elimination} contains three polynomial equations from
which only two are linearly independent since the skew-symmetric
matrix $[f(\vxdp,\lambda)]_{\times}$ is rank two.

To solve the systems of polynomial equations resulting from the
presented problems, we use the \Gb method~\cite{Cox-BOOK05}.
In particular, we used the automatic generators proposed in
\cite{Kukelova-ECCV08,Larsson-CVPR17}; however, for our problems the
coefficients of the input equations are not fully independent.  This
means that using the default settings for the automatic generator
\cite{Kukelova-ECCV08,Larsson-CVPR17}, which initialize the
coefficients of equations by random values from $\mathbb{Z}_p$, does
not lead to correct solvers. Correct problems instances with values
from $\mathbb{Z}_p$ are needed to initialize the automatic generator
to obtain working Gr{\"o}bner basis solvers.

The straightforward application of the automatic
generator~\cite{Kukelova-ECCV08,Larsson-CVPR17} to the needed
constraints with correct coefficients from $\mathbb{Z}_p$ resulted in
large templates and unstable solvers, especially for the two-direction
problems.  The \Gb solvers generated for the original constraints have
template matrices with sizes $80 \times 84$, $74 \times 76$, $348
\times 354$, and $730 \times 734$ for the \rgntwordctevp,
\rgntwordsctevp, \rgntwotwordctevp and \rgntwotwordsctevp problems,
respectively. Therefore, we use the hidden-variable trick to eliminate
the vanishing translation directions together with ideal saturation to
eliminate parasitic solutions~\cite{Cox-BOOK05,Larsson-ICCV17}. The
reformulated constraints are simpler systems in only 3 or 4 unknowns,
and the solvers generated by the Gr{\"o}bner basis method are smaller
and more stable. The reduced elimination template sizes for the
simplified solvers are summarized in \tabref{tab:solver_properties},
and wall clock timings for the simplified solvers are reported in
\secref{sec:wall_clock}. Optimized C++ implementations for all the
proposed solvers are provided.

Next, we describe the solvers based on the hidden-variable trick in
more detail.

\subsubsection{One-Direction EVP Solvers}
\label{sec:evp_one_direction_solvers}
For the one-direction \rgntwordsctevp solver we have
$\bar{s}^{\vu}_{1} = \bar{s}^{\vu}_{2} = 1$.  Therefore the
constraints~\eqref{eq:scalar_elimination} result in two pairs of
linearly independent equations without the scale parameter
$\bar{s}^{\vu}_{i}$ for $i=1,2$, and two linearly independent
equations with an unknown relative scale $\bar{s}^{\vu}_{3}$ for the
third point correspondence, \ie, \ $i=3$.  Additionally, we have the
orthogonality constraint in~\eqref{eq:orthogonality_constraint}.  All
together we have seven equations in seven unknowns
($l_1,l_2,u_1,u_2,u_3,\bar{s}^{\vu}_{3},\lambda$).

\begin{figure*}[ht!]
  \centering
  \resizebox{0.9\textwidth}{!}{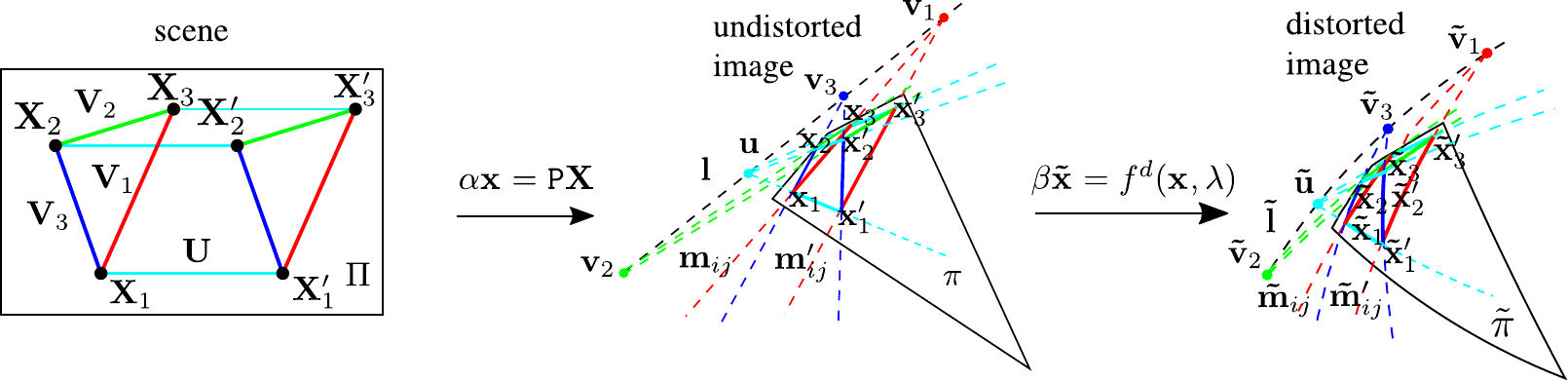}
  \caption{The Geometry of the EVL Constraints. The scene plane
    $\Pi$ contains the preimage of radially-distorted
    conjugately-translated affine-covariant regions, equivalently, 3
    translated points in the direction \vU. This configuration had 3
    additional translation directions \vV[1],\vV[2],\vV[3] that can be
    used to design a solver.  In the image plane $\pi$, the joins of
    each of the images of the 3 pairs of parallel lines (colored red,
    green and blue) meet at the imaged scene plane's vanishing line
    $\vl$. Each incidence of a vanishing point \vu,\vv[1],\vv[2] and
    \vv[3] with $\vl$ generates a scalar constraint equation. Two
    equations are needed to estimate $\vl$ and three are necessary to
    jointly estimate $\vl$ and $\lambda$. Note that \vu can be
    estimated from one of 3 meets of distinct joins of undistorted
    point correspondences, but only 1 such meet can be used as a
    constraint in the EVL formulation.}
  \label{fig:evl_geometry}
  \vspace{-5pt}
\end{figure*}

Note, that these equations are linear with respect to the vanishing
translation direction $\vu$.  Therefore, we can rewrite the seven
equations as
\begin{equation}
\M{M}(l_1,l_2,\bar{s}^{\vu}_{3},\lambda) \colvec{4}{u_1}{u_2}{u_3}{1}
= \ve[0],
\label{eq:H3_HV}
\end{equation}
where $\M{M}(l_1,l_2,\bar{s}^{\vu}_{3},\lambda)$ is a $7\times 4$
matrix whose elements are polynomials in
$(l_1,l_2,\bar{s}^{\vu}_{3},\lambda)$.

Since $\M{M}(l_1,l_2,\bar{s}^{\vu}_{3},\lambda)$ has a null vector, it
must be
rank deficient.  Therefore, all the $4 \times 4$ cofactors of
$\M{M}(l_1,l_2,\bar{s}^{\vu}_{3},\lambda)$ must equal zero.  This
results in $\binom{7}{4} = 35$ polynomial equations which only
involve four unknowns.

Unfortunately, the formulation~\eqref{eq:H3_HV} introduces a
one-dimensional family of false solutions. These are not present in
the original system and correspond to solutions where the first three
columns of $\M{M}$ become rank deficient. In this case there exist
null vectors to $\M{M}$ such that the last element of the vector is
zero, \ie, not of the same form as in \eqref{eq:H3_HV}.

These false solutions can be removed by saturating
\cite{Larsson-ICCV17} any of the $3\times 3$ cofactors from the first
three columns of $\M{M}$.
The matrix $\M{M}$ has the following form,
\begin{equation} \small
  \label{eq:Mstructure}
  \M{M}(l_1,l_2,\bar{s}^{\vu}_{3},\lambda) =
  \begin{bmatrix}
    m_{11} & m_{12} & 0 & m_{14} \\
    m_{21} & m_{22} & 0 & m_{24} \\
    m_{31} & 0 & m_{33} & m_{34} \\
    m_{41} & 0 & m_{43} & m_{44} \\
    m_{51} & m_{52} & 0 & m_{54} \\	
    m_{61} & 0 & m_{63} & m_{64} \\
    l_1 & l_2 & 1 & 0
  \end{bmatrix},
\end{equation}   
where $m_{ij}$ are polynomials in $l_1,l_2,\bar{s}^{\vu}_3$ and
$\lambda$. We choose to saturate the $3\times 3$ cofactor
corresponding to the first, second and last row since it reduces to
only the top-left $2\times 2$ cofactor,
\ie, $m_{11}m_{22}-m_{12}m_{21}$, which is only a quadratic polynomial
in the unknowns. The other $3\times 3$ determinants are more
complicated and leads to larger polynomial solvers. Using the
saturation technique from Larsson \etal \cite{Larsson-ICCV17}, we were
able to create a polynomial solver for this saturated ideal. The size
of the elimination template is $24 \times 26$. Note that without using
the hidden-variable trick the elimination template was $74\times
76$. The number of solutions is two.

For the \rgntwordctevp solver we can use the same hidden-variable
trick. In this case $\bar{s}^{\vu}_{1} = \bar{s}^{\vu}_{2} =
\bar{s}^{\vu}_{3} = 1$; therefore, the matrix $\M{M}$ in
\eqref{eq:H3_HV} contains only three unknowns $l_1,l_2$ and $\lambda$.
This problem is over-constrained, and one of the two constraints from a point correspondence goes unused. Thus, for this problem we can drop
one of the equations from~\eqref{eq:scalar_elimination}, e.g., for
$i=3$, and the matrix $\M{M}$ in~\eqref{eq:H3_HV} has size $6 \times
4$. In this case all $4 \times 4$ cofactors of
$\M{M}$ result in 15 equations in 3 unknowns.
Similar to the 3 point case, this introduces a one-dimensional family
of false solutions. The matrix $\M{M}$ has a similar structure as in
\eqref{eq:Mstructure} and again it is sufficient to saturate the
top-left $2\times 2$ cofactor. For this formulation we were able to
create a solver with template size $14 \times 18$ (compared with
$80\times 84$ without using hidden-variable trick). The number of
solutions is four.

\subsubsection{Two-Direction EVP Solvers}
\label{sec:two_direction}
In the case of the two-direction \rgntwotwordsctevp solver, the input
equations for two vanishing translation directions
$\vu=\rowvec{3}{u_1}{u_2}{u_3}^{\T}$ and
$\vv=\rowvec{3}{v_1}{v_2}{v_3}^{\T}$ can be separated into two sets
of equations, \ie, the equations containing $\vu$ and the equations
containing $\vv$. Note that in this case we have two equations of
the form~\eqref{eq:orthogonality_constraint}, \ie, the equation for
the direction $\vu$ and the equation for the direction $\vv$ and
we have an unknown relative scale $\bar{s}^{\vv}_{4}$.  Therefore, the
final system of 10 equations in 10 unknowns can be rewritten using two
matrix equations as
\begin{equation}
  \M{M}_1(l_1,l_2,\lambda) \colvec{4}{u_1}{u_2}{u_3}{1} = \ve[0], \;\;
  \M{M}_2(l_1,l_2,\bar{s}^{\vv}_{4},\lambda)
  \colvec{4}{v_1}{v_2}{v_3}{1} = \ve[0],
  \label{eq:H4_HV}
\end{equation}
where $\M{M}_1$ and $\M{M}_2$ are $5 \times 4$ matrices such that the
elements are polynomials in $(l_1,l_2,\lambda)$ and
$(l_1,l_2,\bar{s}^{\vv}_{4},\lambda)$, respectively.

Again all $4 \times 4$ cofactors of $\M{M}_1$ and $\M{M}_2$
must concurrently equal zero.  This results in $5+5 = 10$ polynomial
equations in four unknowns $(l_1,l_2,\bar{s}^{\vv}_{4},\lambda)$.  In
this case, only 39 additional false solutions arise from the
hidden-variable trick. The matrices $\M{M}_1$ and $\M{M}_2$ have a
similar structure as in \eqref{eq:Mstructure} and again it is
sufficient to saturate the top-left $2\times 2$ cofactors to
remove the extra solutions. By saturating these determinants we were
able to create a solver with template size $76\times 80$ (previously
$730\times 734$). The number of solutions is four. 

Finally, for the \rgntwotwordctevp two-direction solver,
$\bar{s}^{\vu}_{1} = \bar{s}^{\vu}_{2} = 1$ and $\bar{s}^{\vv}_{3} =
\bar{s}^{\vv}_{4} = 1$.
This problem is over-constrained, so we can drop one of the equations from
constraint~\eqref{eq:scalar_elimination}, e.g., for $i=4$.  Therefore,
the matrix $\M{M}_2$ from~\eqref{eq:H4_HV} has size $4\times 4$, and
it contains only three unknowns $(l_1,l_2,\lambda)$.  All $4 \times 4$
cofactors of $\M{M}_1$ and $\M{M}_2$ result in $5+1 = 6$ polynomial
equations in three unknowns $(l_1,l_2,\lambda)$.

For this case we get 18 additional false solutions. Investigations in
Macaulay2 \cite{M2} revealed that for this particular formulation, it
is sufficient to only saturate the top-left $2\times 2$ cofactor of
$\M{M}_1$ and the top-left element of $\M{M}_2$. Generating the
polynomial solver with saturation resulted in a template size of $54
\times 60$ (previously $348\times 354$). The number of solutions is
six.

\subsection{The Eliminated Vanishing Line (EVL) Solver}
\label{sec:evl_constraints}
Suppose $\buildset{ \xdcspond[i] }{i=1}{3}$ are point correspondences
extracted from a radially-distorted conjugately-translated
affine-covariant region correspondence as shown in
\figref{fig:evl_geometry}. Then their preimages $\buildset{
  \Xcspond[i] }{i=1}{3}$ on the scene plane $\Pi$ are in
correspondence with a translation, denote it \ve[U], which is color
coded cyan in \figref{fig:evl_geometry}. This point configuration has
three additional translation directions $\ve[V]_1$,$\ve[V]_2$ and
$\ve[V]_3$, (colored red, green and blue, respectively), where each of
the four imaged translation directions induces four radially-distorted
conjugate translations in the distorted image.

A vanishing point, \ie, \vu, \vv[1], \vv[2], \vv[3], can be recovered
from each meet of joins of pairs of conjugate-translations that share
the same translation direction in the scene plane, \eg,
\begin{equation}
 \gamma \vv[1] = (\vx[1] \times \vx[3]) \times (\vxp[1] \times
 \vxp[3]).
\end{equation}
There are six such pairs to choose from, one for each of \vv[1],\vv[2]
and \vv[3] and three for \vu, which is the vanishing point of the
translation direction for the undistorted point correspondences
$\buildset{ \xcspond[i] }{i=1}{3}$.

As proved in \secref{sec:meets_of_joins}, each meet of joins puts a
constraint on the vanishing line \vl. It will be shown that only three
of the six vanishing point constructions are necessary to solve for
the undistortion parameter $\lambda$ and vanishing line \vl. It will
also be shown that exactly one of any of the three meets of joins of
conjugate translations from $\buildset{ \xcspond[i] }{i=1}{3}$ can be
used to constrain \vl.

Without loss of generality, we use the joins of pairs of conjugate
translations meeting at $\ve[v]_1$,$\ve[v]_2$, and $\ve[v]_3$, which
are substituted into the vanishing point-vanishing line incident
constraint of \eqref{eq:vp_vl_incidence}
\begin{equation}
  \label{eq:vl_vp_coincidence}
   \vv[i]^{\T}\vl = \left((\vx[i] \times \vx[j]) \times (\vxp[i]
   \times \vxp[j])\right)^{\T}\vl = 0,
\end{equation}
where $i < j$ and $i,j \in \buildset{1 \ldots 3}{}{}$. The homogeneity
of \eqref{eq:vl_vp_coincidence} is used to eliminate any non-zero
scalars. Substituting radially-distorted points for undistorted points
in \eqref{eq:vl_vp_coincidence} using \eqref{eq:division_model} gives
\begin{equation}
  \label{eq:join_meet_meet_constraint}
  \left(f(\vxd[i],\lambda) \times f(\vxd[j],\lambda)\right) \times
  \left(f(\vxdp[i],\lambda) \times f(\vxdp[j],\lambda)\right)^{\T}\vl
  = 0.
\end{equation}
The skew-symmetric operator, denoted \sksym{\cdot}, is used to
transform \eqref{eq:join_meet_meet_constraint} into the homogeneous
matrix-vector equation
\begin{equation}
\label{eq:matrix_join_meet_meet_constraint}
\left(
\sksym{\sksym{f(\vxd[i],\lambda)}f(\vxd[j],\lambda)}\sksym{f(\vxdp[i],\lambda)}
f(\vxdp[j],\lambda)\right)^{\T} \vl = 0,
\end{equation}
where where $i < j$ and $i,j \in \buildset{1 \ldots
  3}{}{}$. Independent scalar constraint equations of the form
\eqref{eq:matrix_join_meet_meet_constraint} can be stacked to add the
necessary number of constraints for jointly estimating \vl and
$\lambda$.

\begin{figure*}[htb]
  \centering 
   \subfloat[Nikon D300, 16mm \label{fig:NikonD300_16mm_evl}] {
    \includegraphics[width=0.2\linewidth]{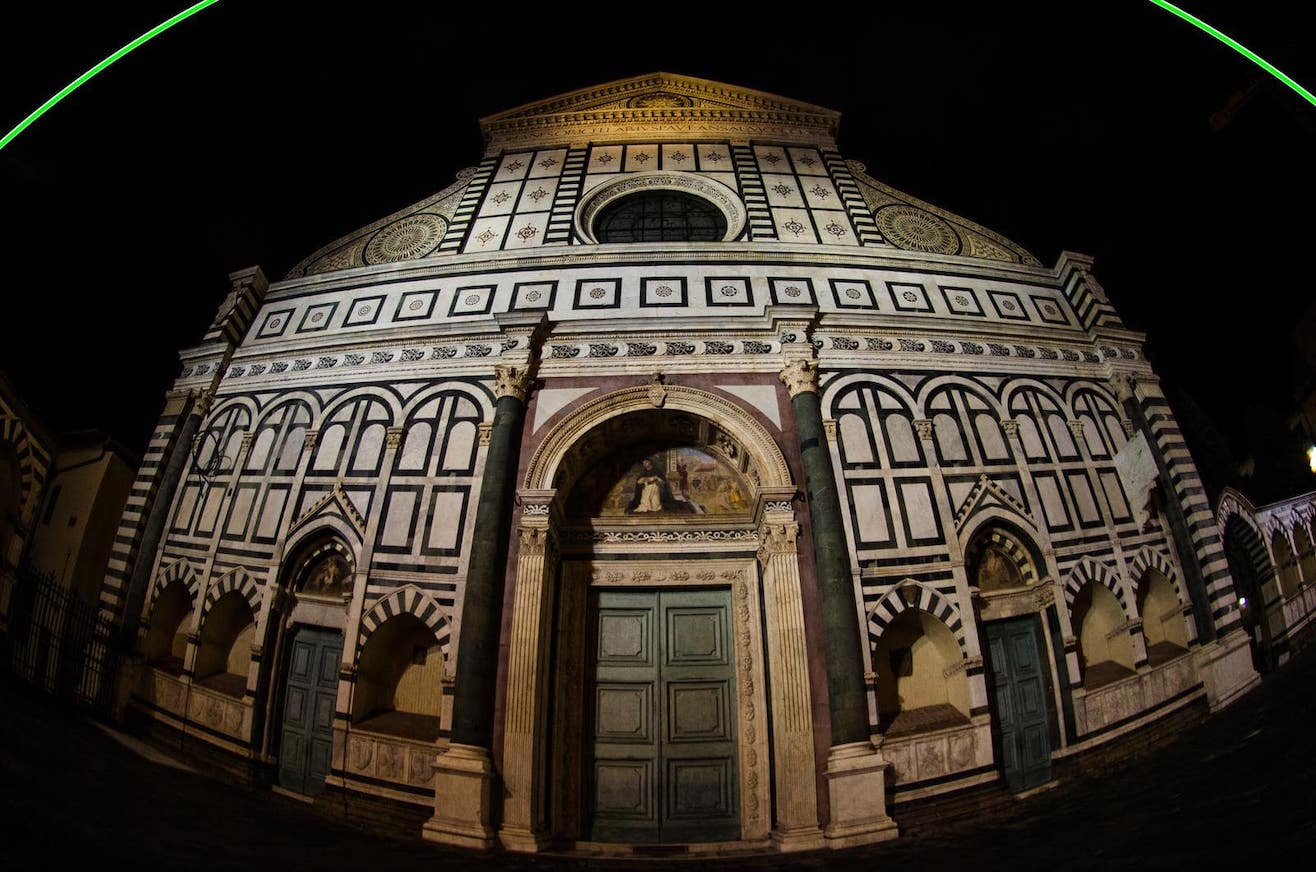}
  }
  \subfloat[Nikon D300, 15mm \label{fig:NikonD300_15mm_evl}] {
    \includegraphics[width=0.2\linewidth]{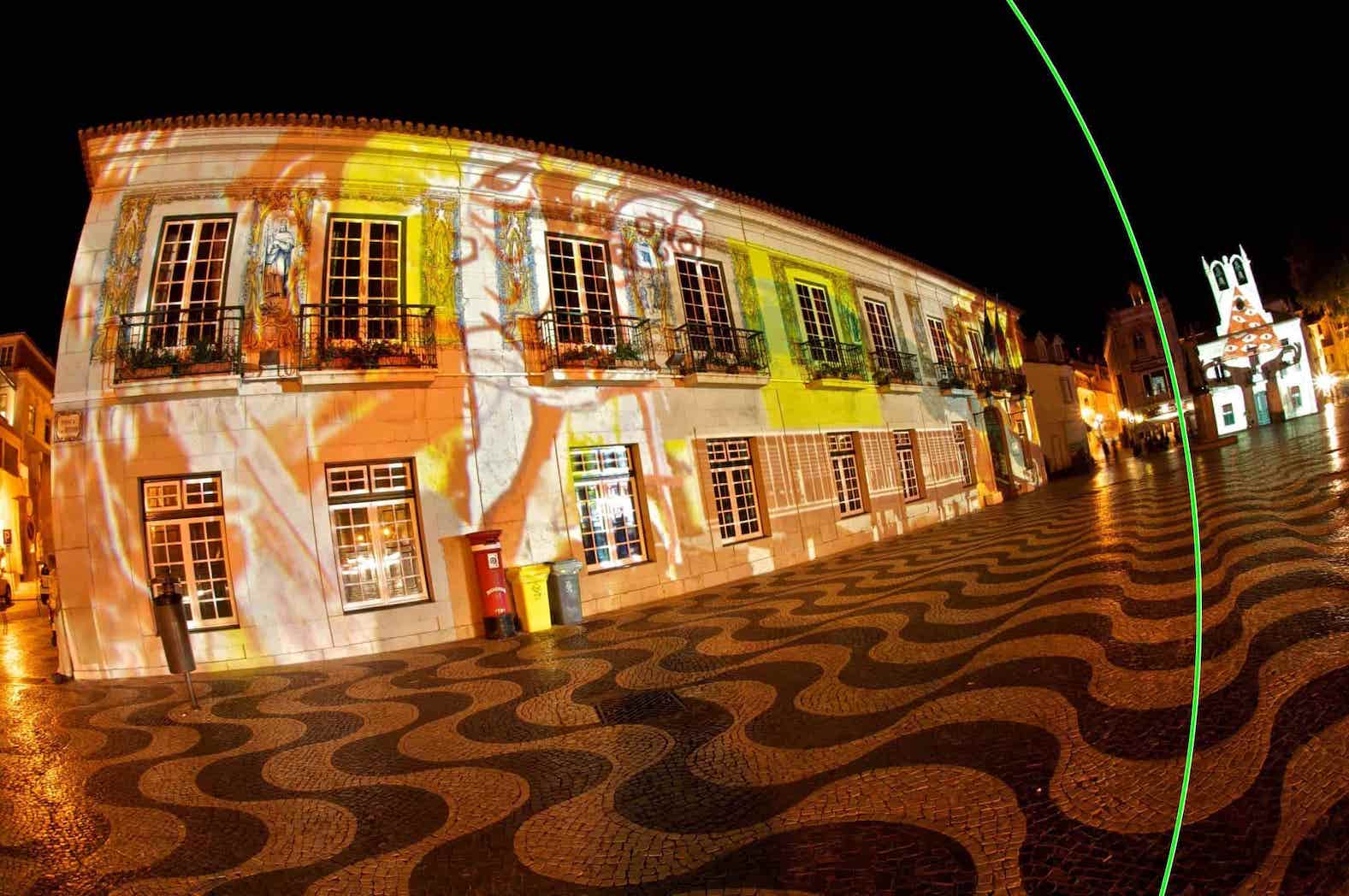}
  }
  \subfloat[unknown \label{fig:unknown_evl}]{
    \includegraphics[width=0.2\linewidth]{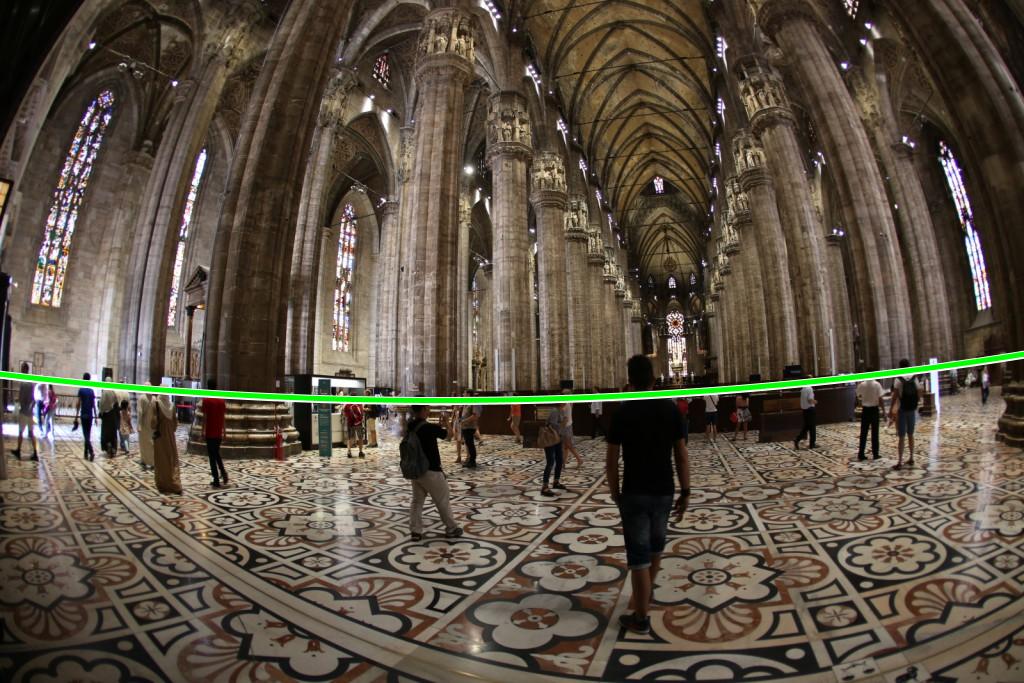}   
  }
  \subfloat[Olympus E M1, 15mm \label{fig:Olympus_E_M1_15mm_evl}]{
    \includegraphics[width=0.18\linewidth]{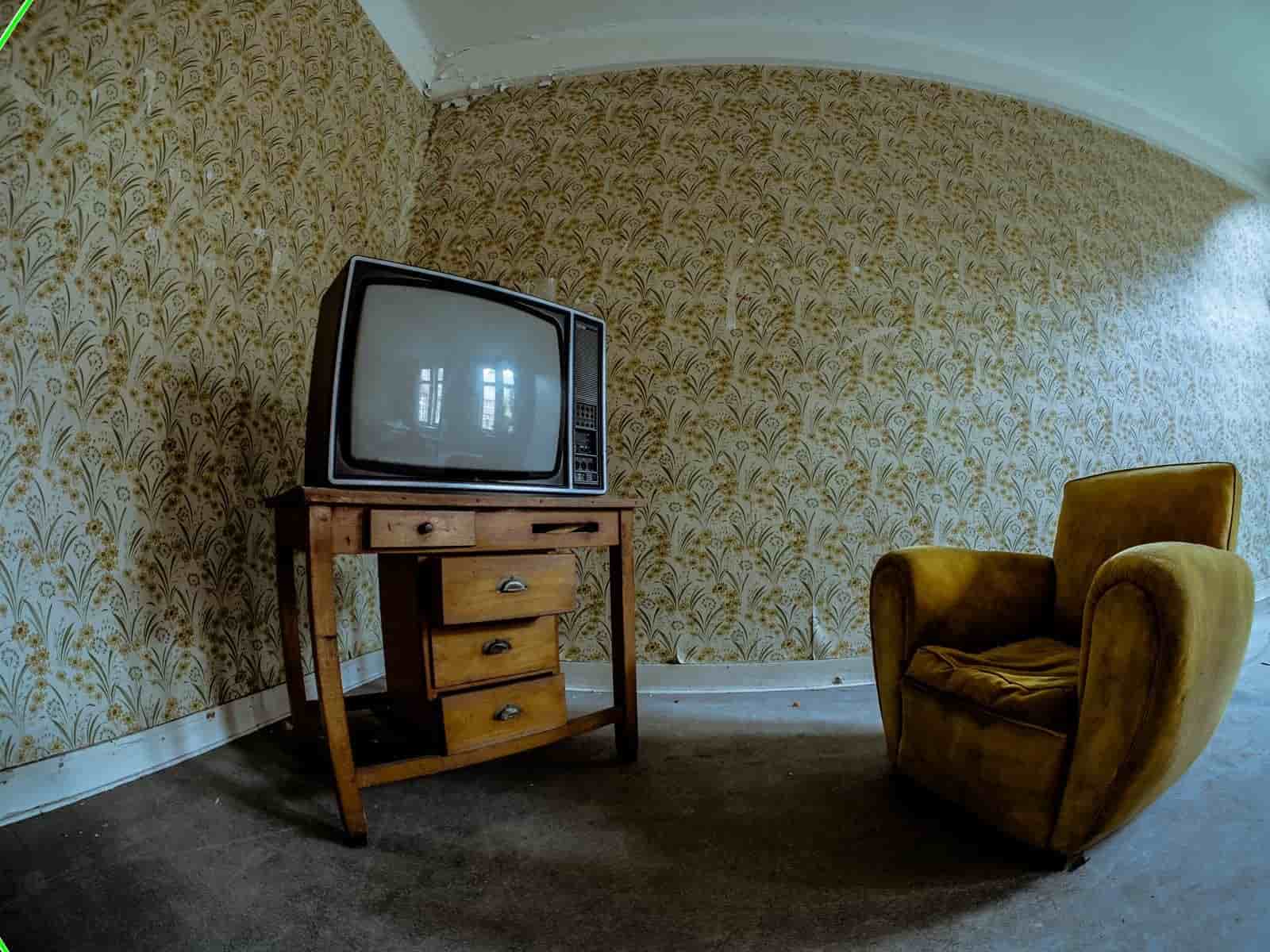}
  }
  \subfloat[Nikon D810, 14mm \label{fig:Nikon_D810_14mm_evl}]{
    \includegraphics[width=0.2\linewidth]{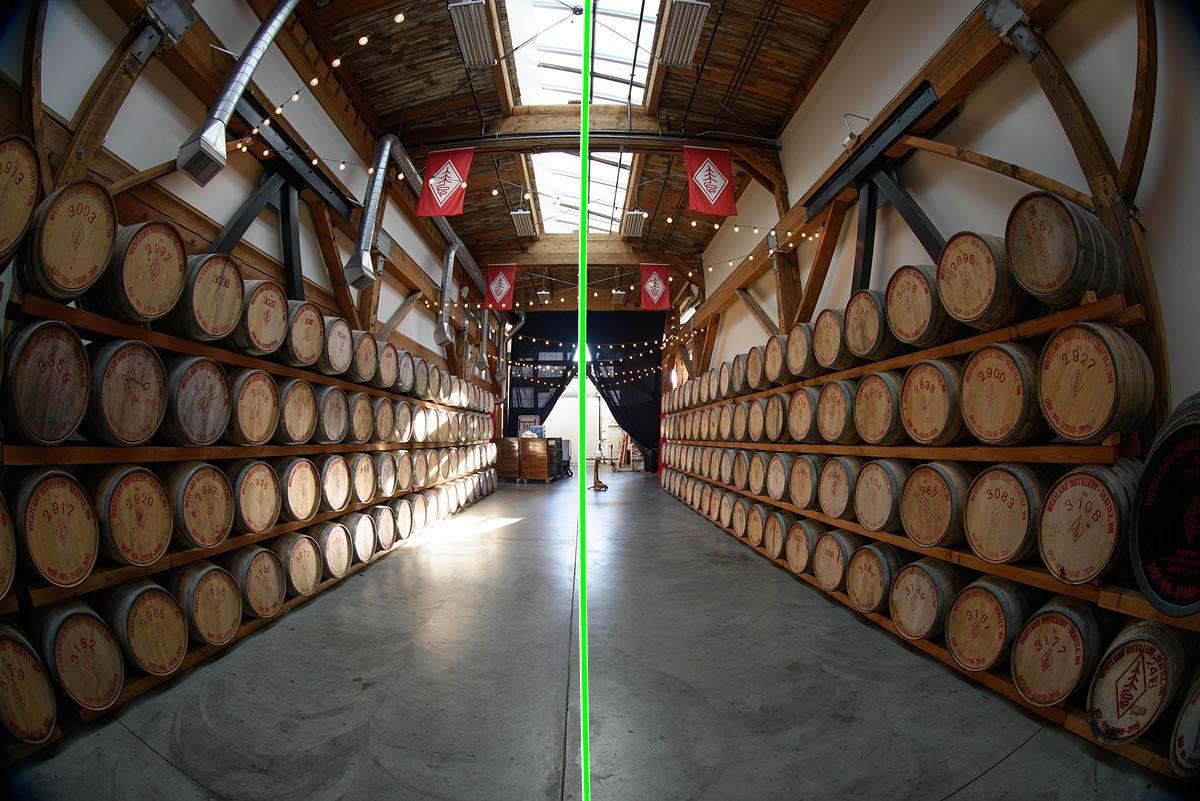}   
  }
  \\[-1.5ex]
  \subfloat {
    \includegraphics[width=0.2\linewidth]{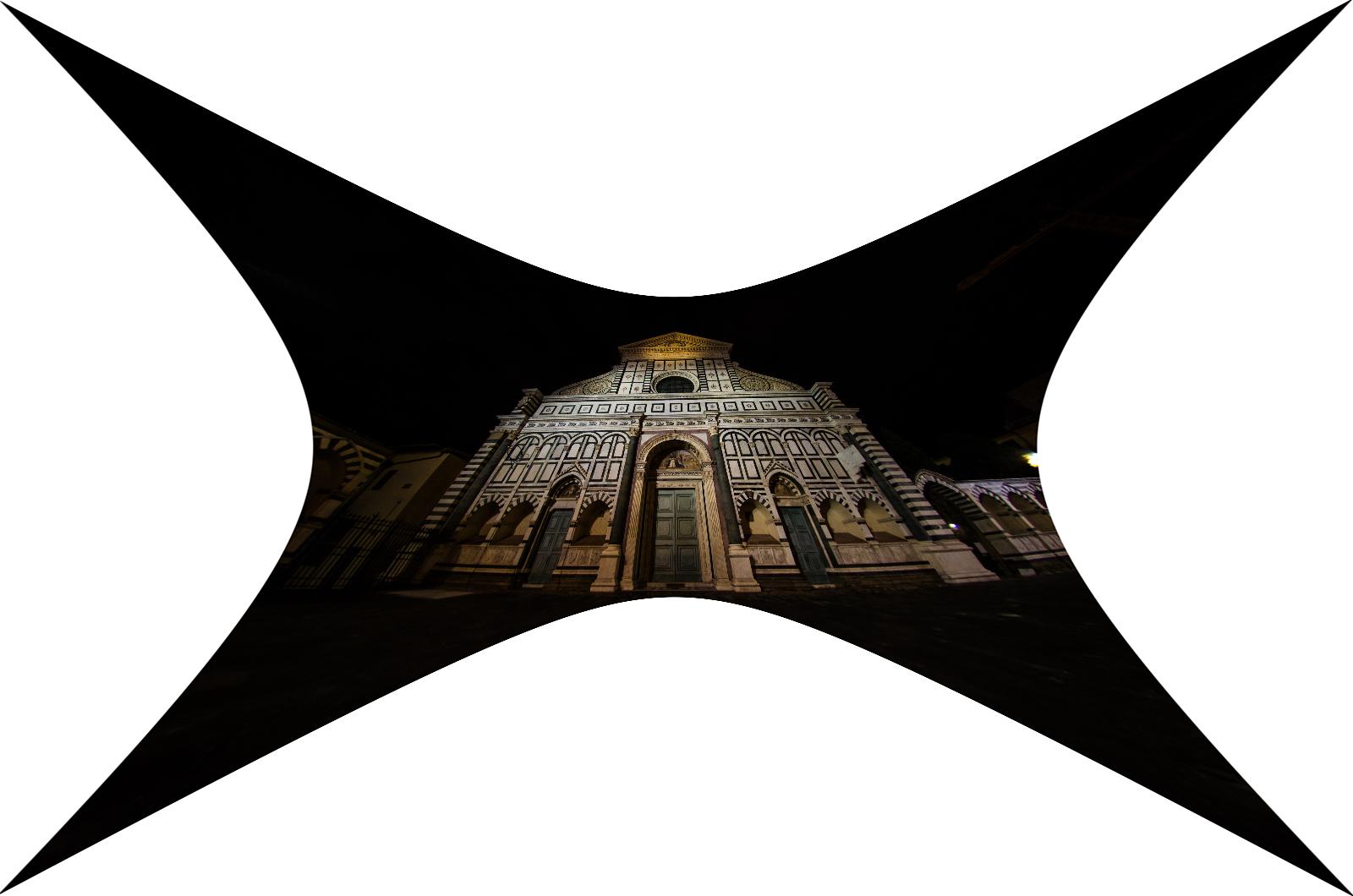}
  }
  \subfloat {
    \includegraphics[width=0.2\linewidth]{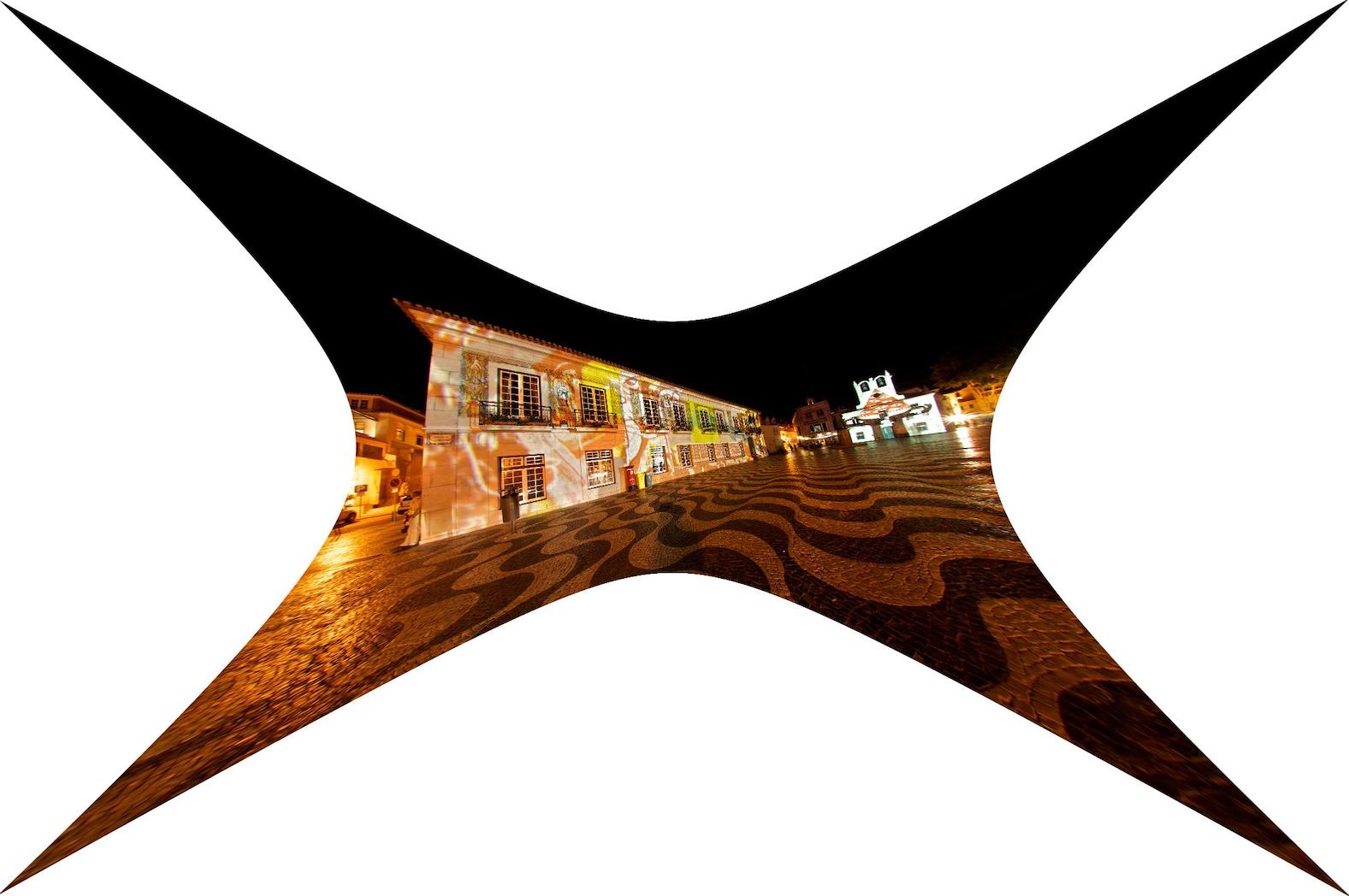}
  }
  \subfloat{
    \includegraphics[width=0.2\linewidth]{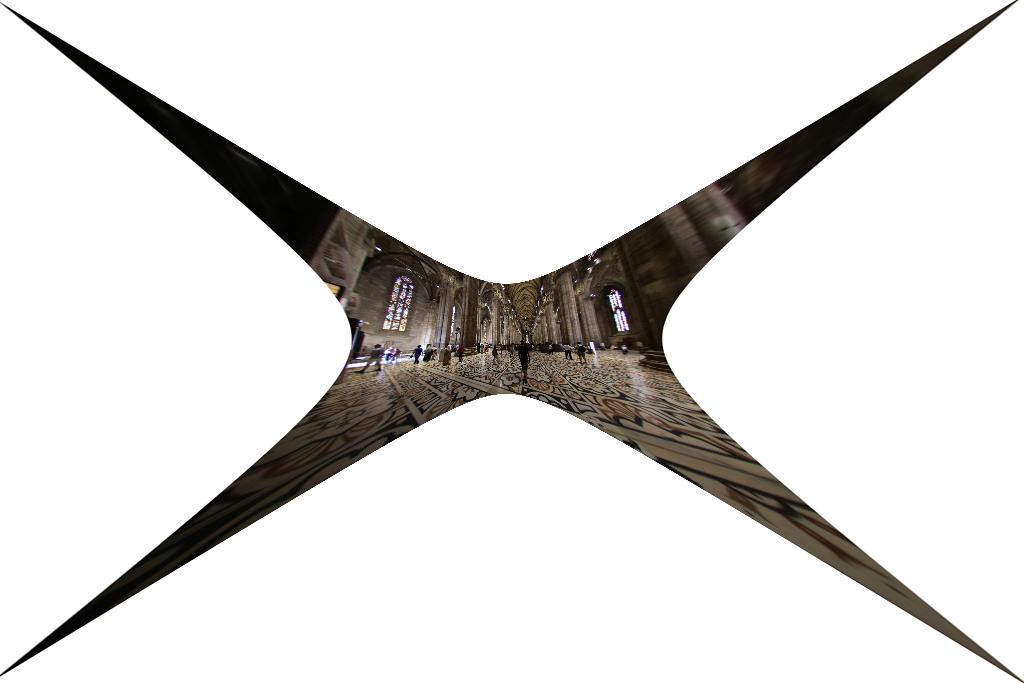}
  }
  \subfloat{
    \includegraphics[width=0.18\linewidth]{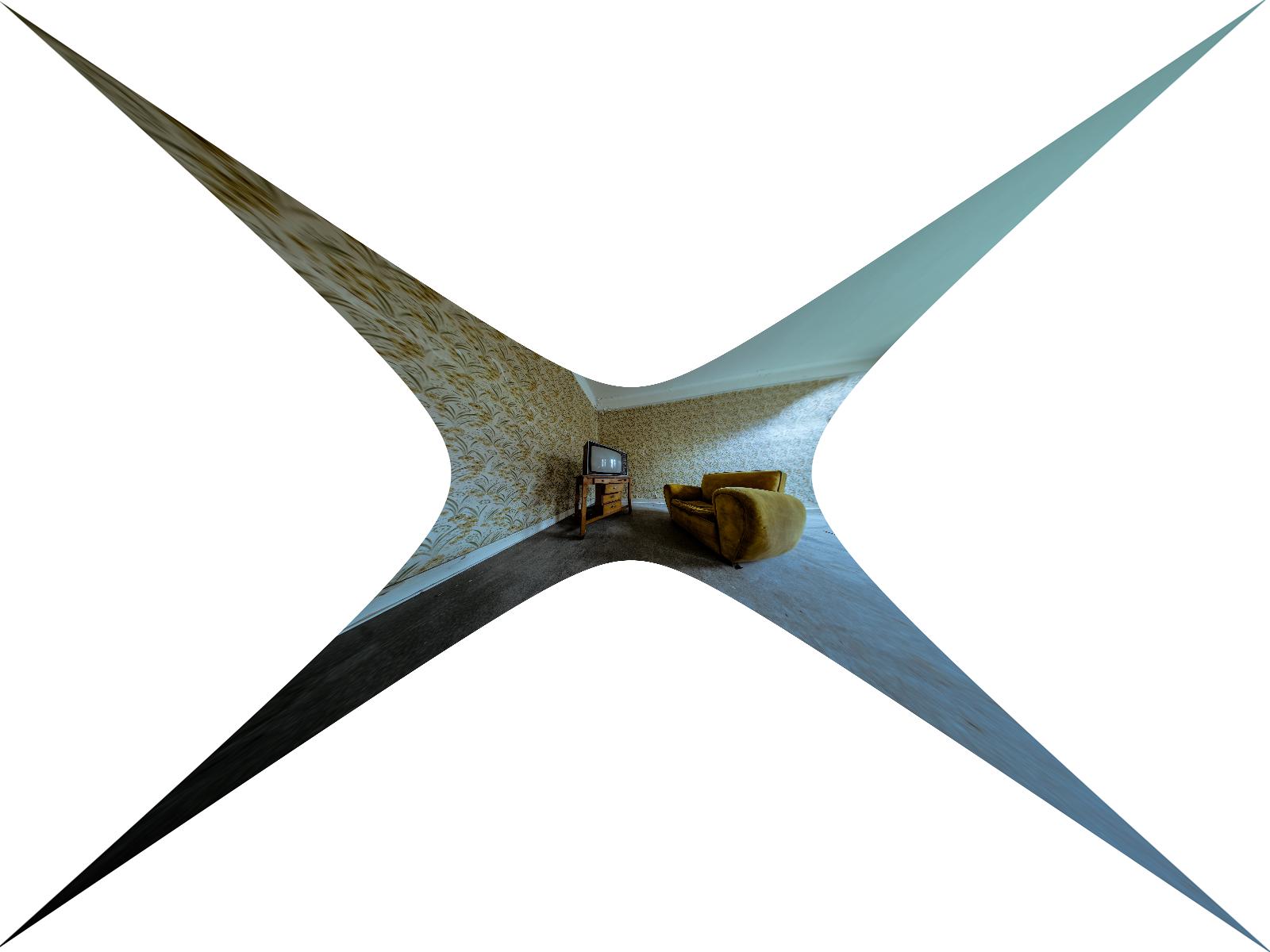}
  }
  \subfloat{
    \includegraphics[width=0.2\linewidth]{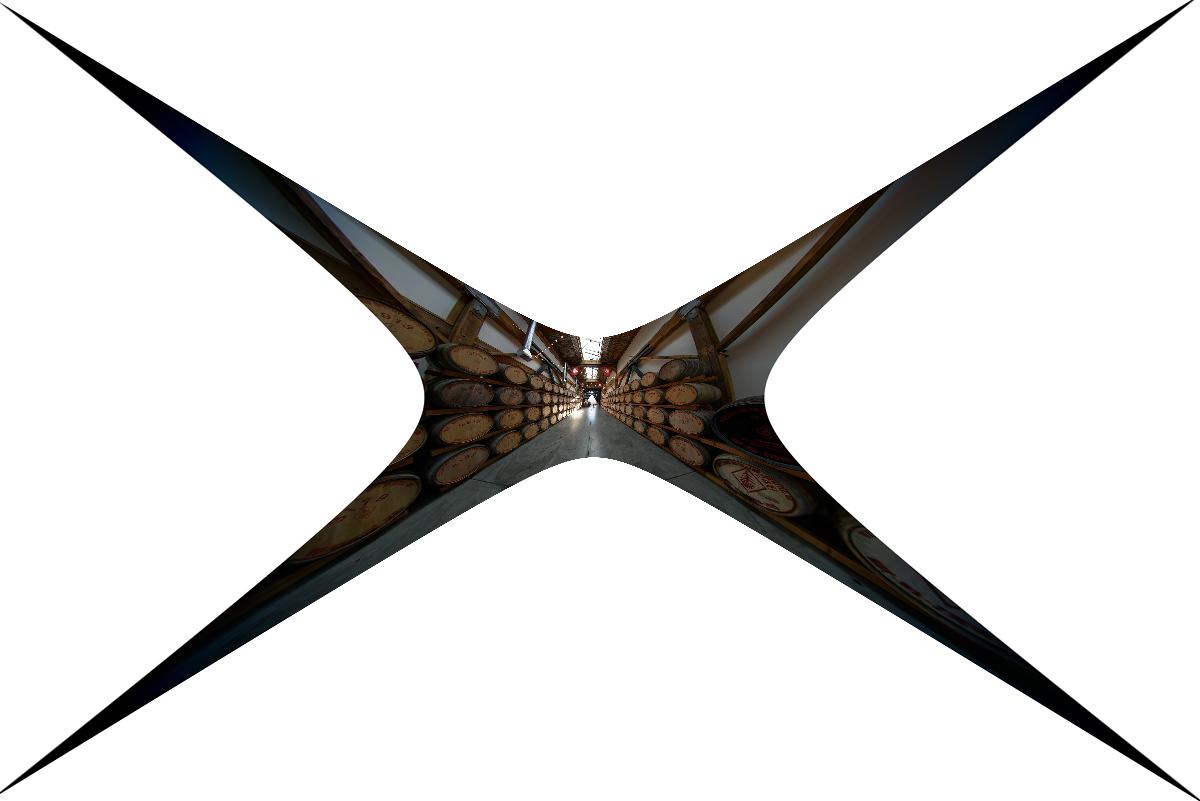}
  }
  \\[-1.5ex]
  \subfloat {
    \includegraphics[trim={13cm 4cm 13cm 12cm},clip,width=0.2\linewidth]{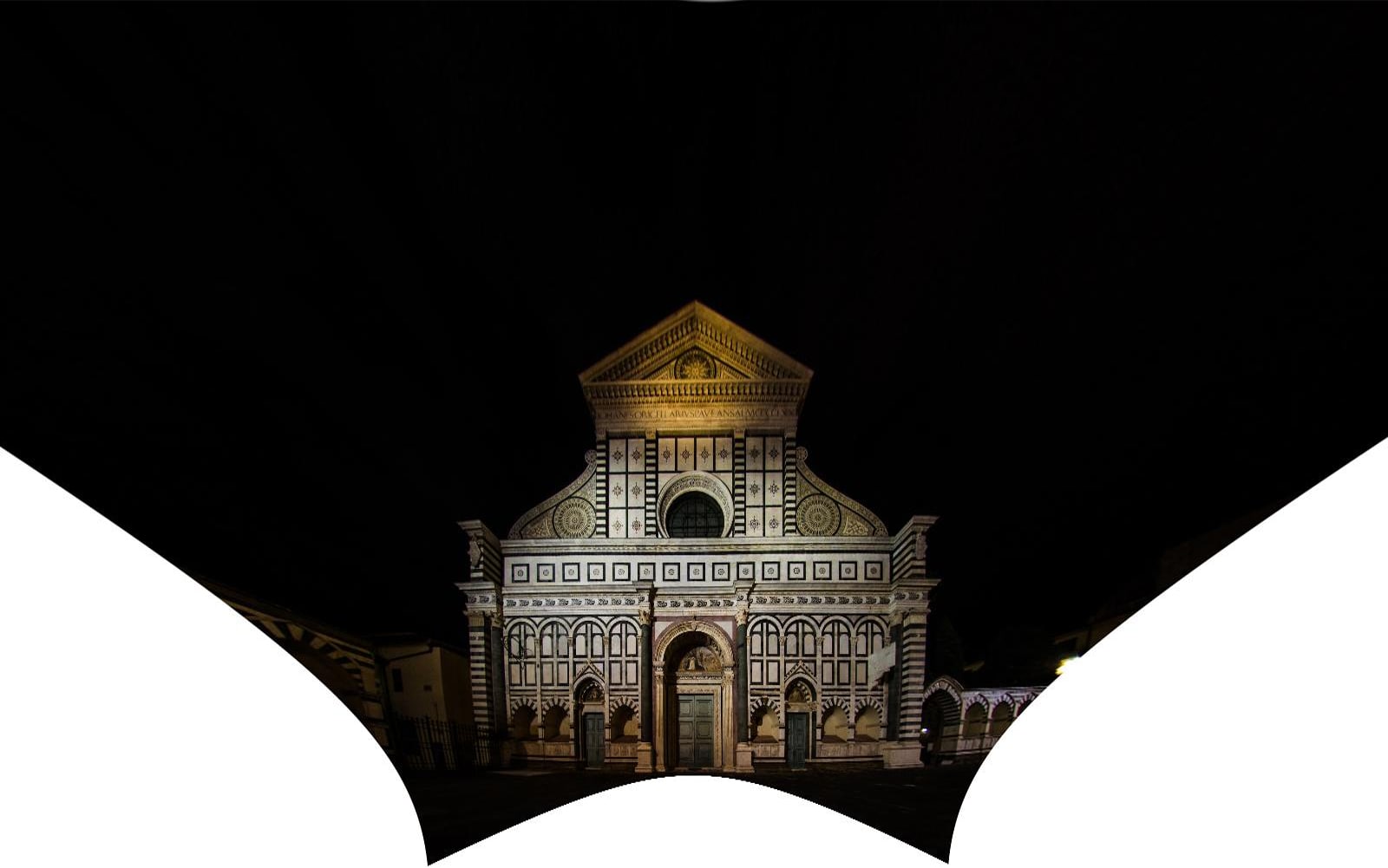}
  }
  \subfloat {
    \includegraphics[trim={5.5cm 22cm 23cm 15.5cm},clip,width=0.2\linewidth]{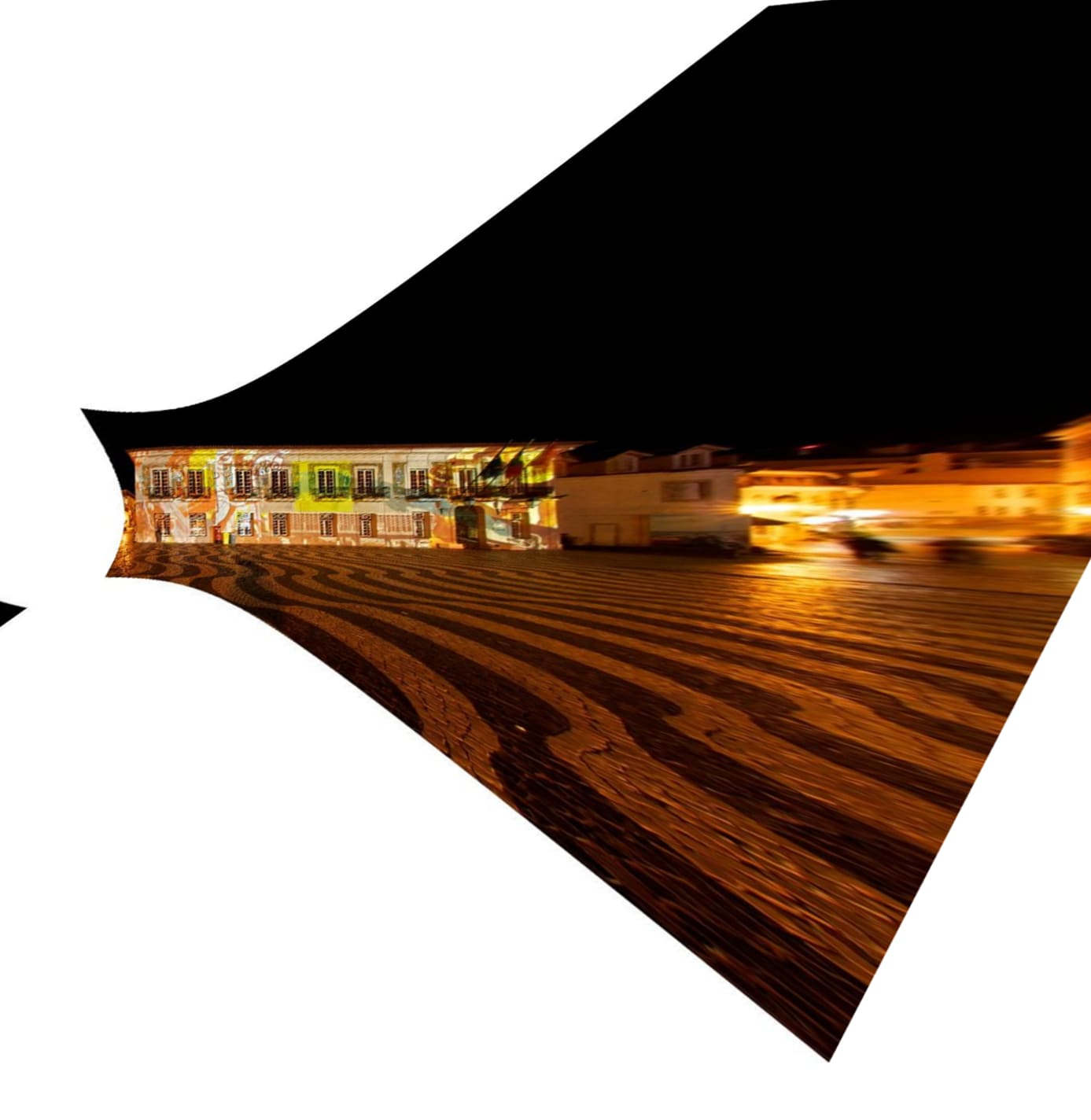}
  }
  \subfloat{
    \includegraphics[trim={40cm 0.6cm 95cm 16.5cm},clip,width=0.2\linewidth]{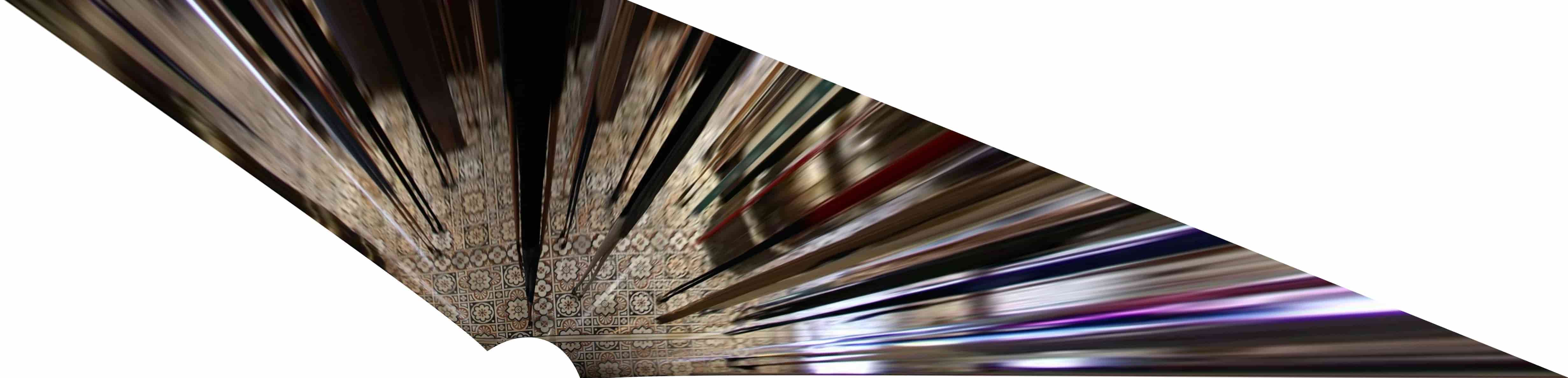}
  }
  \subfloat{
    \includegraphics[trim={3cm 6.5cm 4cm 9.5cm},clip,width=0.18\linewidth]{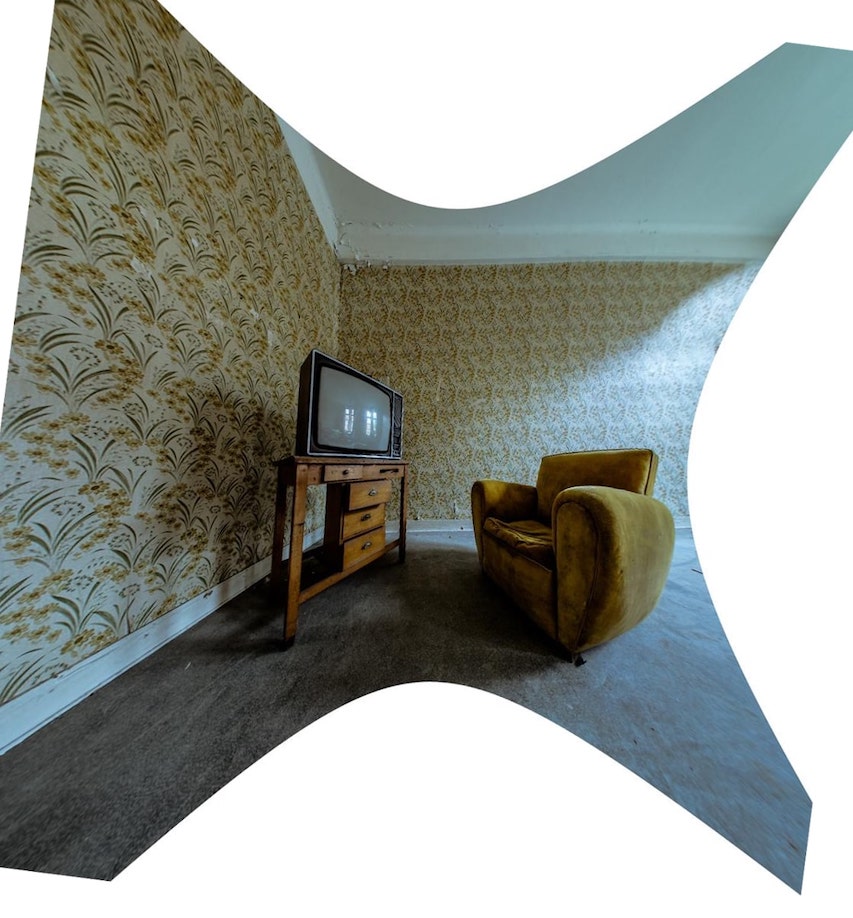}
  }
  \subfloat{
    \includegraphics[trim={15cm 23.2cm 8cm 25cm},clip,width=0.2\linewidth]{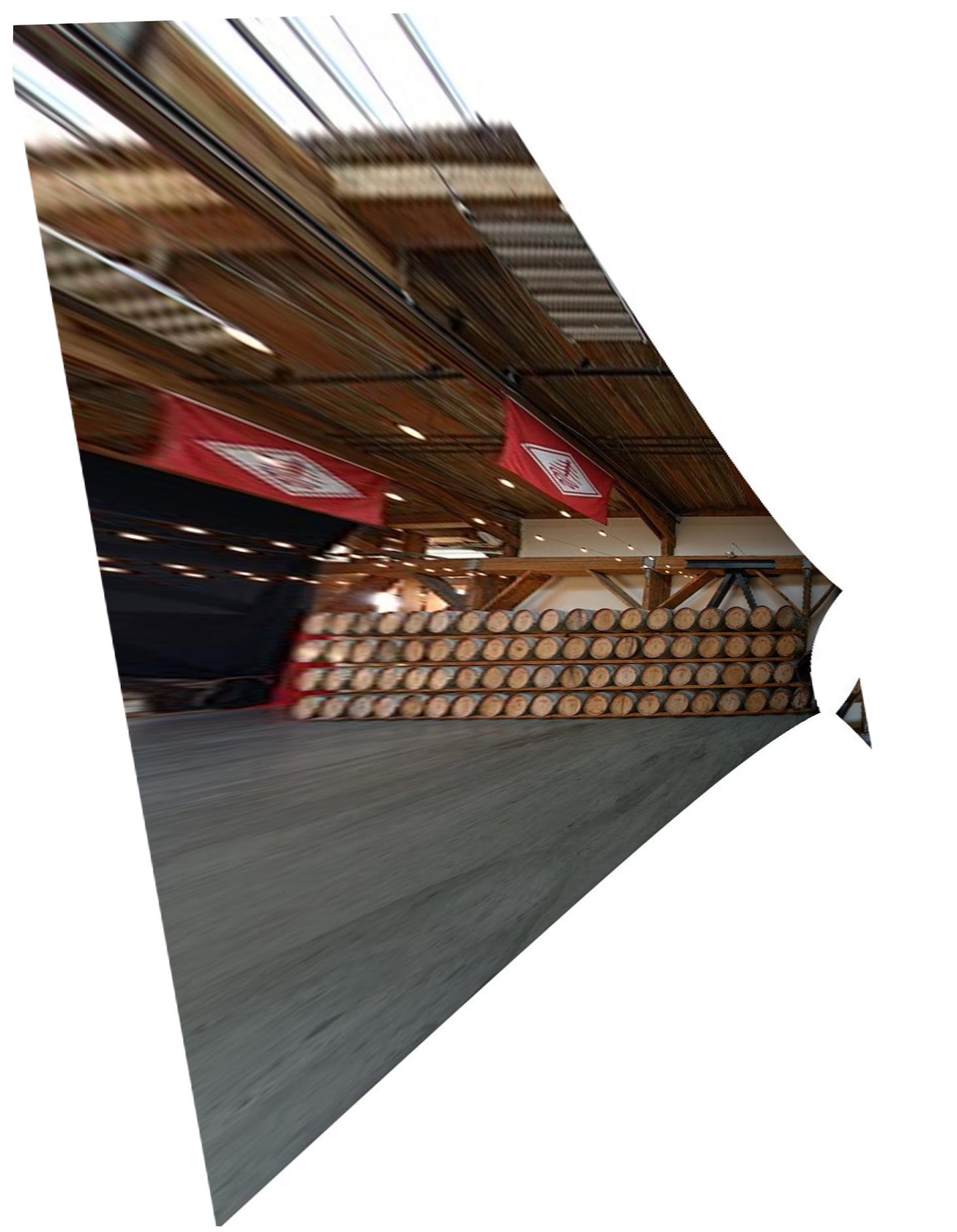}
  }
  \caption{\EVL Solver Results on Fisheye Images.  The
    distorted image of the vanishing line is rendered in green in the
    input images on the top row. Results were produced using the
    \rgntwordctevl with 1-correspondence sampling in a \RANSAC
    framework. The \rgntwordctevl solver runs in $0.5$
    {\textmu}s. Surprisingly, reasonable rectifications are possible
    using the 1-parameter division model for the extreme distortions of
    fisheye lenses. Focal lengths are reported as 35mm equivalent.}
\label{fig:evl_results}
\end{figure*}

\subsubsection{Creating the Solver}
\label{sec:evl_solver_creation}
Each vanishing point \vu,\vv[1],\vv[2] and \vv[3] generates one scalar
constraint on the vanishing line \vl. There are four unknowns in
constraint \eqref{eq:matrix_join_meet_meet_constraint}, namely $\vl =
\rowvec{3}{l_1}{l_2}{l_3}^{\T}$ and the division model parameter
$\lambda$. The vanishing line \vl is homogeneous, so it has only two
degrees of freedom. Thus $3$ scalar constraint equations of the form
\eqref{eq:matrix_join_meet_meet_constraint} generated by $3$ vanishing
points from the set $\buildset{ \vu,\vv[1],\vv[2],\vv[3] }{}{}$ are
needed, which, as shown in
\eqref{eq:matrix_join_meet_meet_constraint}, can be concisely encoded
in the matrix $\ma{M}(\lambda) \in \mathbb{R}^{3 \times 3}$ as
\begin{equation}
  \label{eq:evl_system_of_eqns}
  \ma{M}(\lambda)\colvec{3}{l_1}{l_2}{l_3} = \ve[0].
\end{equation}
Note that only 1 of the 3 meets of joins of conjugately-translated
point correspondences from $\buildset{ \xcspond[i] }{i=1}{3}$ can be
used since there is no constraint included that enforces
\begin{equation*}
  \label{eq:meets_must_be_equal}
  ((\vx[i] \times \vxp[i]) \times (\vx[j] \times \vxp[j])) \times ((\vx[i]
  \times \vxp[i]) \times (\vx[k] \times \vxp[k]))=\ve[0],
\end{equation*}
where $i,j,k \in \buildset{1 \ldots 3}{}{}$ and $i \neq j$. Therefore,
at least two of \vv[1],\vv[2], and \vv[3] must be used, and the two chosen
meets can be combined with exactly one of the meets the can be
constructed from $\buildset{ \xcspond[i] }{i=1}{3}$. Including the
case where each of \vv[1],\vv[2], and \vv[3] is used gives
$3\binom{3}{2}+1=10$ possible combinations of meets. Selecting the
optimal meets for the most accurate rectification is addressed in
\secref{sec:choosing_constraint_eqs}.

The division model parameter $\lambda$ is hidden in
\eqref{eq:evl_system_of_eqns} using the hidden-variable trick in the
entries of coefficient matrix $\ma{M}$, which are polynomials only in
$\lambda$ \cite{Cox-BOOK05}. Thus \vl has been eliminated, which
motivates the \EVL name.

Matrix $\ma{M}(\lambda)$ is rank deficient since it has a null vector,
which implies that $\det \ma{M}(\lambda)=0$. The determinant
constraint defines a univariate quartic with unknown $\lambda$, which
can be solved in closed form.  After $\lambda$ has been recovered, the
vanishing line \vl is obtained by solving for the null space of
$\ma{M}$. The \EVL solver is denoted \rgntwordctevl.

\subsubsection{Best Minimal Solution Selection}
\label{sec:choosing_constraint_eqs}
The \EVL geometry of \figref{fig:evl_geometry} has 10 meets that can
be used to generate scalar constraint equations in
\eqref{eq:matrix_join_meet_meet_constraint}.  However, only 3 meets
are needed to jointly estimate \vl and $\lambda$.  Since the time to
solution for the \rgntwordctevl is only $\mathbf{0.5}$
\textbf{{\textmu}s}, the solutions for all minimal subsets of meets
can be verified against the unused constraints, \eg, if the meets of
joins of the radially-distorted conjugately-translated correspondences
associated with \vv[1],\vv[2], and \vv[3] are used, then the
correspondences associated with \vu (cyan direction) can be used for
verification. The minimal subset of meets is chosen that minimizes the
sum of symmetric transfer errors
\begin{equation}
  \label{eq:symm_xfer_err}
  \sum_i d(\vxd[i],f^{d}(\mH^{\inv}f(\vxdp[i],\lambda),\lambda))^2+
  d(f^{d}(\mH f(\vxd[i],\lambda),\lambda),\vxdp[i])^2,
\end{equation}
where \xdcspond are radially-distorted conjugately-translated point
correspondences that are not included in a minimal configuration for
estimating rectification. We call this approach \emph{best minimal
  solution selection}.

Evaluating the quality of the minimal solution on
\eqref{eq:symm_xfer_err} has several
benefits: \begin{enumerate*}[(i)]\item Near degenerate correspondence
  configurations can be rejected, \item Correspondences with geometric
  properties that are more robust to noise will be preferred, \eg,
  regions that are further apart, \item and expensive \RANSAC
  consensus set construction can be preempted, if there is no minimal
  solution that has sufficiently small symmetric transfer error as
  defined in \eqref{eq:symm_xfer_err}.
\end{enumerate*}

Best minimal solution selection is evaluated in the sensitivity
studies in \secref{sec:experiments}. The solver incorporating best
minimal solution selection is denoted in the standard way,
\rgntwordctevl. For comparison we introduce a baseline solver, denoted
\rgntwordctevlrand, which randomly selects from the 10 possible
constraint configurations associated with the \EVL geometry (see
\figref{fig:evl_geometry}). As expected, the \rgntwordctevl performs
better than \rgntwordctevlrand on all sensitivity measures. 
 
\subsubsection{Optimal Estimate of the Vanishing Point}
\label{sec:unknown_vp_estimation}
Unlike the \EVP solvers in \secref{sec:evp_constraints}, which jointly
estimate the vanishing point \vu (shown in \figref{fig:evl_geometry})
using all constraints from the set of conjugate
translations \buildset{ \xcspond[i] }{i=1}{3} (see
\eqref{eq:scalar_elimination}), the \rgntwordctevl solver maximally
uses two joins from \buildset{ \xcspond[i] }{i=1}{3} and possibly none
if only the red, green and blue translation directions in
\figref{fig:evl_geometry} are selected as the best minimal solution.

The vanishing point \vu of the cyan translation direction can be
recovered after the vanishing line \vl and division model parameter
$\lambda$ are estimated (\eg, by \rgntwordctevl) by solving a
constrained least squares system that includes all constraints induced
by \buildset{ \xcspond[i] }{i=1}{3} (see
\figref{fig:evl_geometry}). The incidence of \vu with \vl is
explicitly enforced by including \eqref{eq:vp_vl_incidence} into the constraints. Define $\ve[h]_{\vu}^{1\T},\ve[h]_{\vu}^{2\T}$,
and $\ve[h]_{\vu}^{3\T}$ to be the rows of a conjugate translation,
\begin{equation}
  \label{eq:conj_trans_rows}
  \alpha \vxp = \mHu \vx =
  \begin{bmatrix}\ve[h]_{\vu}^{1} &
    \ve[h]_{\vu}^{2} & \ve[h]_{\vu}^{3}
  \end{bmatrix}^{\T} \vx
  = \begin{bmatrix} \ma{I}_3+\vu\vl^{\T} \end{bmatrix} \vx.
\end{equation}
The homogeneous scale in \eqref{eq:conj_trans_rows} can be eliminated
by substituting $\ve[h]_{\vu}^{3\T}\vx$ for $\alpha$, and the system
can be rearranged such that
\begin{equation}
  \label{eq:eliminated_alpha}
  \begin{split}
    & \vx^{\T}\ve[h]_{\vu}^{1} =(\xp\vx^{\T})\ve[h]_{\vu}^{3} \\
    & \vx^{\T}\ve[h]_{\vu}^{2} =(\yp\vx^{\T})\ve[h]_{\vu}^{3}.
  \end{split}
\end{equation}

Collecting the terms of vanishing point after expanding the dot
products in \eqref{eq:eliminated_alpha} for each pair of \buildset{ \xcspond[i] }{i=1}{3}
along with an incidence constraint $\vl^{\T}\vu = 0$ gives the constrained least squares
problem

\begin{equation*}
\begin{gathered}
\begin{aligned}
\underset{\vu}{\text{minimize}}~&
\left\lVert\ma{M}\vu
- \ve[y]\right\rVert^2\\
\text{subject to~} &~\vl^{\T}\vu = 0,
\end{aligned}\\
\text{where} \quad \ma{M} =
  \begin{bmatrix}
    & \vdots  & \\
    -\vl^{\T}\vx[i] & 0 & \xp[i](\vl^{T}\vx[i]) \\
    0 & \vl^{\T}\vx[i] & \yp[i](\vl^{\T}\vx[i]) \\
    & \vdots & \\
\end{bmatrix}, \quad
\ve[y]=
  \colvec{4}{\vdots}{x_i-\xp[i]}{y_i-\yp[i]}{\vdots}
\end{gathered}
\end{equation*}


Since the matrix $\begin{bmatrix}\ma{M}^{\T} & \vl \end{bmatrix}^{\T}$
has linearly independent columns, and $\vl^{\T}$ is trivially row
independent, \vu is recovered by solving
\begin{equation}
  \label{eq:cls_vp}
  \begin{bmatrix}
    \ma{M}^{\T}\ma{M} & \vl \\ \vl^{\T} & 0
  \end{bmatrix}
  \colvec{2}{\vu}{z}=\colvec{2}{\ma{M}^{\T}\ve[y]}{0},
\end{equation}
where $z$ is a nuisance variable \cite{Boyd-BOOK04}. Surprisingly, a
superior estimation of the vanishing point \vu is given by using
\eqref{eq:cls_vp} after rectifying with the \EVL \rgntwordctevl solver
than by jointly solving for the rectification, vanishing point, and
division model parameter as done with the \EVP group of solvers (see
the transfer error sensitivity study
\figref{fig:ransac_xfer_error_sensitivity}).

\section{Degeneracies}
\label{sec:pami19_degeneracies}
We identified three important degeneracies for the solvers: Two
geometric configurations of features such that there exists either a
subspace of rectifications or no valid solution, and a modeling
degeneracy introduced by the use of the expression \eqref{eq:recthg}
for the affine-rectifying homography, which requires \vl =
$\rowvec{3}{l_1}{l_2}{l_3}^{\T}$ such that $l_3 \neq 0$
\cite{Hartley-BOOK04}. The proposed solvers and the state-of-the-art
solvers of Pritts \etal in \cite{Pritts-ACCV18,Pritts-IJCV20} suffer
from this modeling degeneracy. It is shown that addressing this
degeneracy requires increasing the complexity of the solvers. There
are likely additional degeneracies between the \EVL and \EVP solver,
but an exhaustive analysis is a difficult theoretical problem.

\subsection{Degenerate Feature Configurations}
\label{sec:degenerate_features}
 Suppose that \begin{enumerate*}[(i)]
\item \mH is a rectifying homography other than the identity
  matrix, \item that the image has no radial distortion, \item and
  that all corresponding points from repeated affine-covariant regions
  fall on a single circle centered at the image
  center. \end{enumerate*} Applying the division model of lens
 undistortion uniformly scales the points about the image
 center. Given $\lambda \neq 0$, for a transformation by
 $f(\cdot,\lambda)$ defined in \eqref{eq:division_model} of the points
 lying on the circle there is a scaling matrix
 $\ma{S}(\lambda)=\diag{1/\lambda,1/\lambda,1}$ that maps the points
 back to their original positions. Thus there is a 1D family of
 rectifying homographies given by $\mH\ma{S}(\lambda)$ for the
 corresponding set of undistorted images given by $f(\cdot,\lambda)$.

Secondly, suppose that the conjugately-translated point
correspondences \xcspond[i] and \xcspond[k] are collinear as shown in
\figref{fig:conjugate_translations}. Let $\ve[m]_i=\vx[i] \times
\vxp[i]$ and $\ve[m]_k=\vx[k] \times \vxp[k]$. Then $\ve[m]_i \times
\ve[m]_k = \ve[0]$, which is not a point in the real-projective plane
\RP[2], and cannot be used to place a constraint on
\vl. Unfortunately, this point configuration is common, \eg, consider
a row of windows on a facade. It is possible that the feature
extraction pipeline will establish collinear correspondences. However,
affine frames constructed from covariant region detections are
typically not in this degenerate configuration since the origin is
defined by blob's center of mass or peak response in scale space and
one of the extents is constructed as a right angle to the first linear
basis vector (see \figref{fig:local_features}). Regardless, the
degeneracy can be avoided by using different meets.
 

\subsection{The Pencil of Vanishing Lines Through the Distortion Center}
\label{sec:pencil_through_origin}
If the vanishing line passes through the image origin, \ie
$\vl=\rowvec{3}{l_1}{l_2}{0}^{\T}$, then the radial term in the
homogeneous coordinate of \eqref{eq:udrect} is canceled. In this case,
it is not possible to recover the division model parameter $\lambda$
from the systems of equations \eqref{eq:Mstructure}, \eqref{eq:H4_HV}
or \eqref{eq:evl_system_of_eqns} solved by any of the proposed
solvers. However, the degeneracy does not arise from the problem
formulation. An affine transform can be applied to the undistorted
image such that the vanishing line \vl in the affine-transformed space
has $l_3 \neq 0$.

The division model requires the image origin to be the distortion
center \cite{Fitzgibbon-CVPR01}. The derivations in this paper assume
that image center, distortion center and image origin are
coincident. The proposed and state-of-the-art solvers of Pritts \etal
\cite{Pritts-ACCV18,Pritts-IJCV20} formulate joint undistortion and
rectification in terms of \eqref{eq:udrect}, which leaves the
distortion center stationary.

Directional cameras see only points in front of the camera
\cite{Hartley-IJCV98}, so the vanishing line cannot intersect the
convex hull of measurements. Therefore, changing basis in the
undistorted space such that any point in the convex hull of the
undistorted feature points (\ie, affine covariant region detections)
is the image origin guarantees that vanishing line will not pass
through the origin. Furthermore, if a point is in the convex hull of
measurements in the distorted space, then it is also in the convex
hull of undistorted measurements. However, the change of basis (\ie, a
translation) is a function of the undistorted point, and thus a
function of the unknown division model parameter $\lambda$, so
applying the coordinate transform increases the complexity of the
solvers. Empirically we did not find this degeneracy to be a
problem. \Eg, \figsref{fig:samyang75mm_fov}, \ref{fig:unknown_evl},
and \ref{fig:Nikon_D810_14mm_evl} show good undistortions of images
and rectifications of imaged scene planes that have vanishing lines
passing close to the center of distortion, which suggest that in these
near-degenerate cases the division-model parameter is sufficiently
observable. Thus we choose to preserve the simplicity of the solvers
(see \tabref{tab:wall_clock}). A new origin in the undistorted space
can be defined by a distorted measurement in the convex hull of
measurements, which will reduce the chance of encountering the
degeneracy, but not eliminate it.

\section{Robust Estimation}
\label{sec:ransac}
The solvers are used in a \LORANSAC-based robust-estimation framework
\cite{Chum-ACCV04,Pritts-CVPR14}. Affine rectifications and
undistortions are jointly hypothesized by one of the proposed
solvers. A metric upgrade is attempted and models with maximal
consensus sets are locally optimized by an extension of the method
introduced in \cite{Pritts-CVPR14}.  The metric-rectifications are
presented in the results.

\begin{figure}[ht]
  \centering
  \subfloat[MSER detection] {
    \includegraphics[trim={0cm 2.2cm 0cm 2.2cm},clip,width=0.3\columnwidth,]{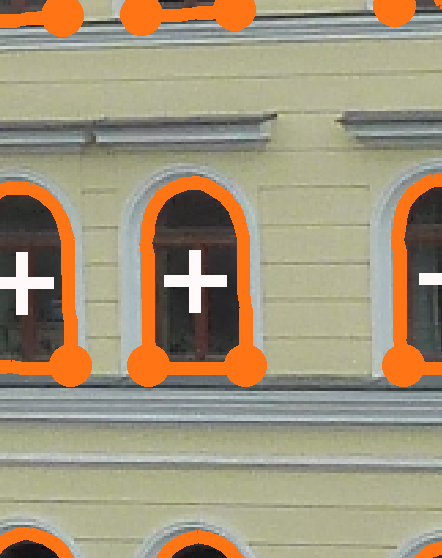}
  }
  \subfloat[Normalized Frame] {
    \parbox{.15\columnwidth} {
      \centering
      \ic{0.99}{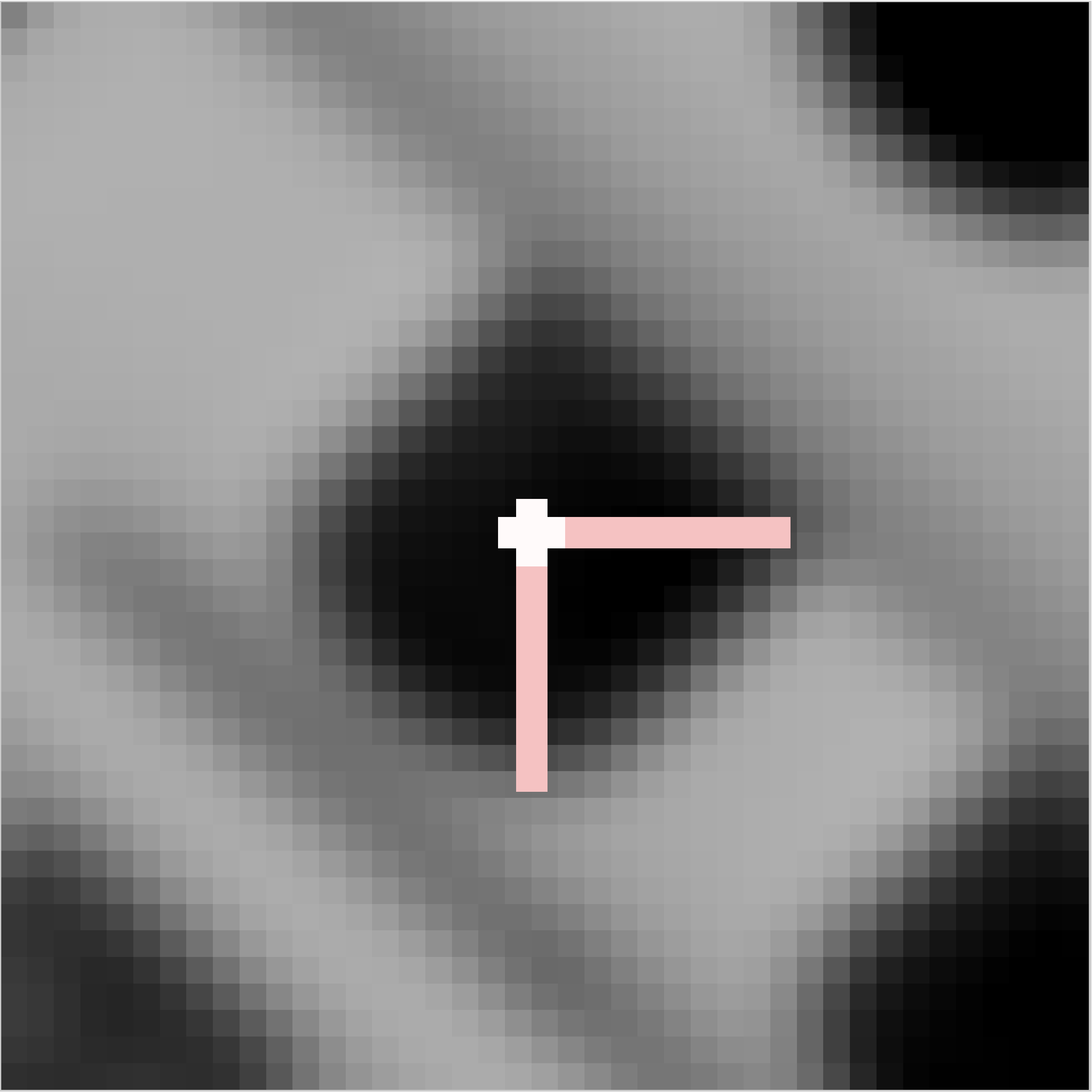}\\ 
      \ic{0.99}{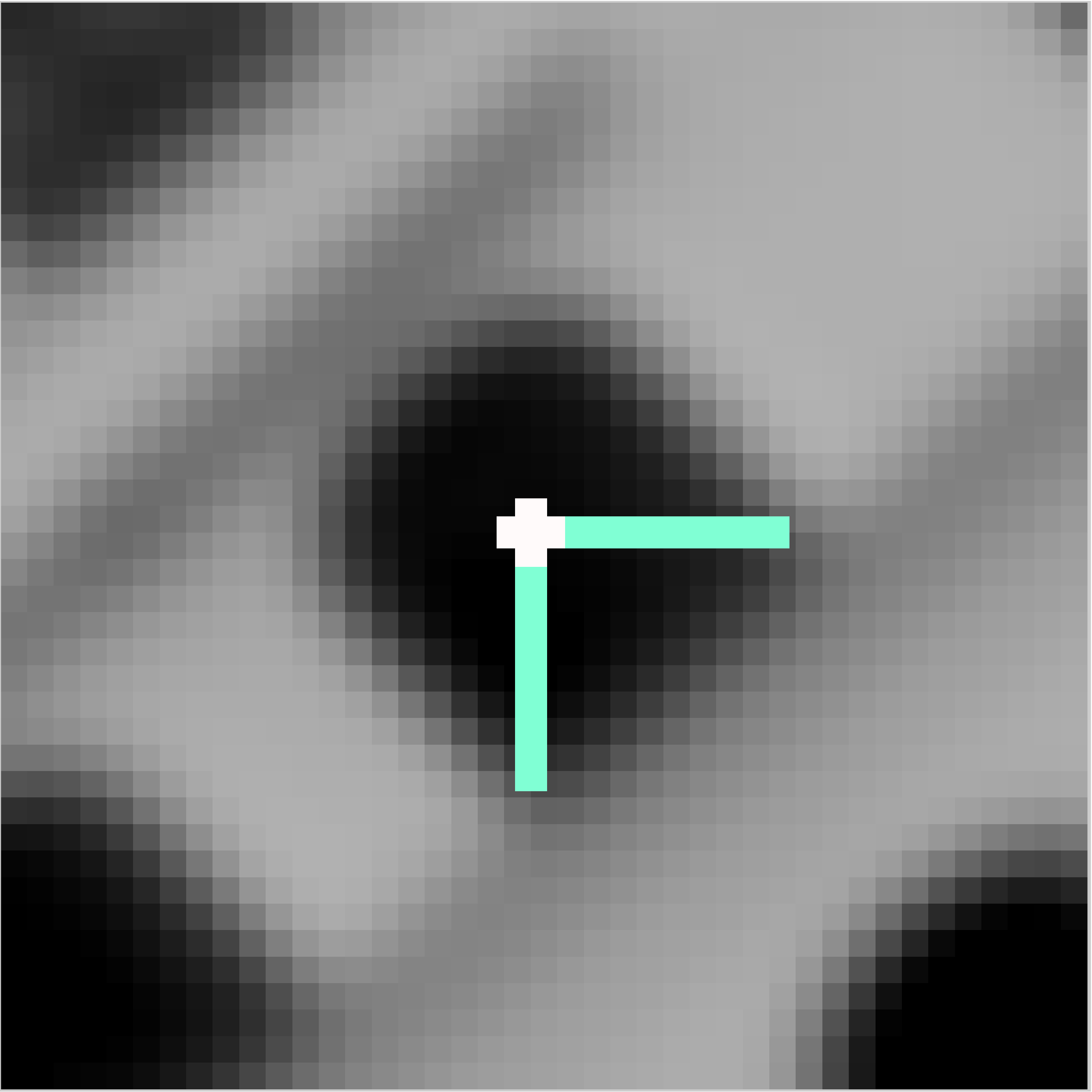}\\
    }
    
    \parbox{.15\columnwidth} {
      \centering
      \ic{0.99}{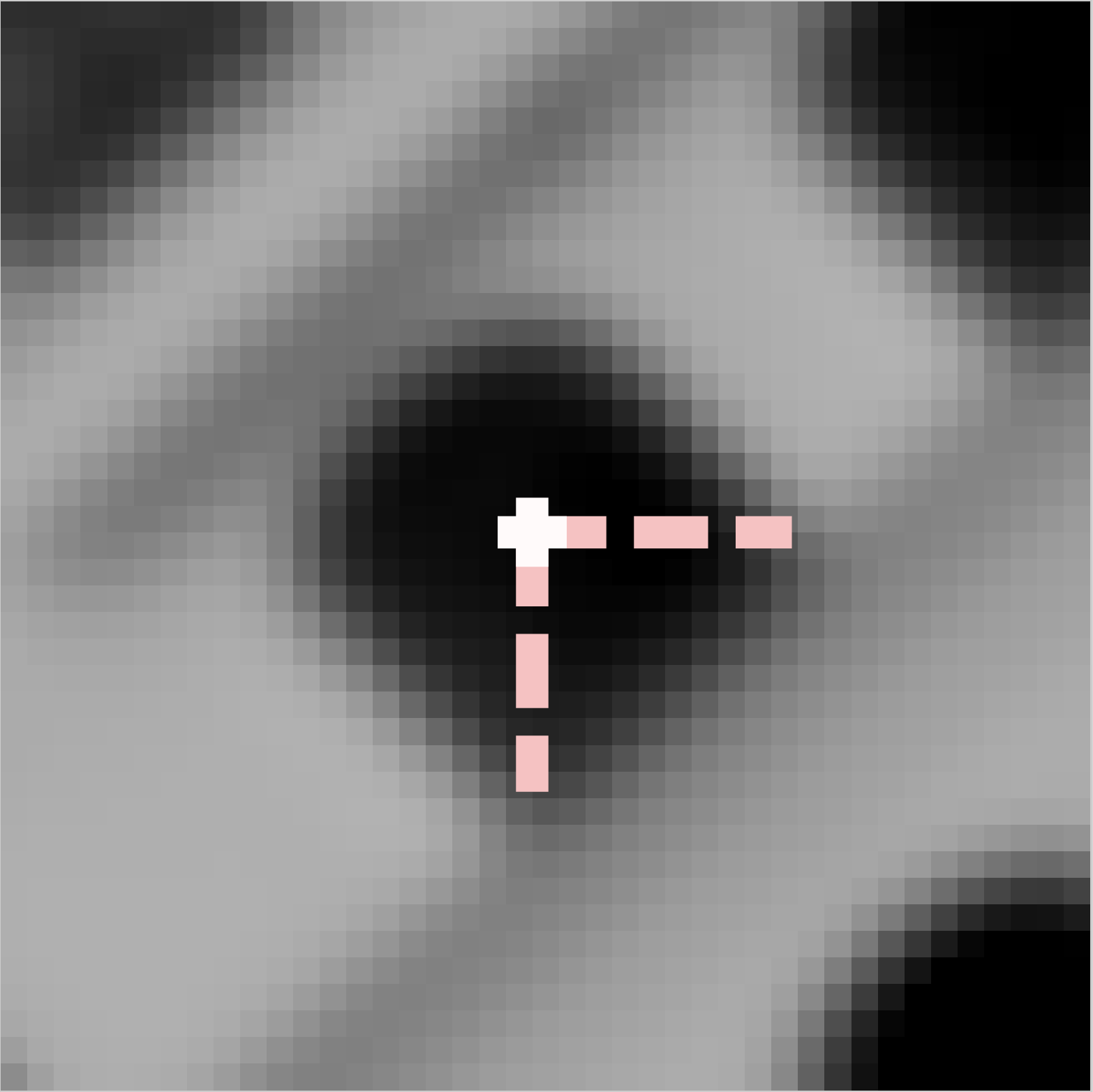}\\
      \ic{0.99}{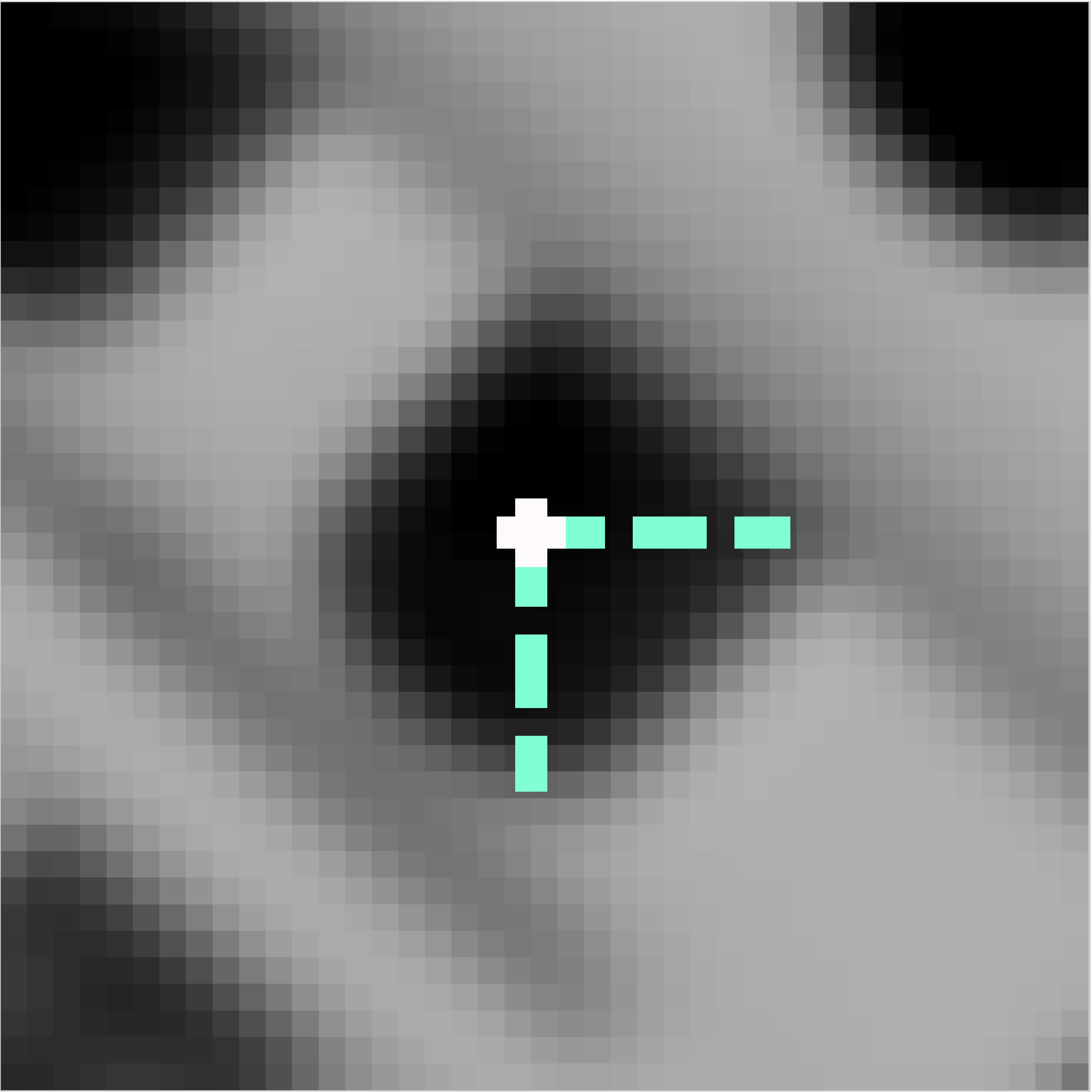}\\   
    }
  }
  \subfloat[LAF representation] {
    \includegraphics[trim={0cm 2.2cm 0cm 2.3cm},clip,width=0.3\columnwidth]{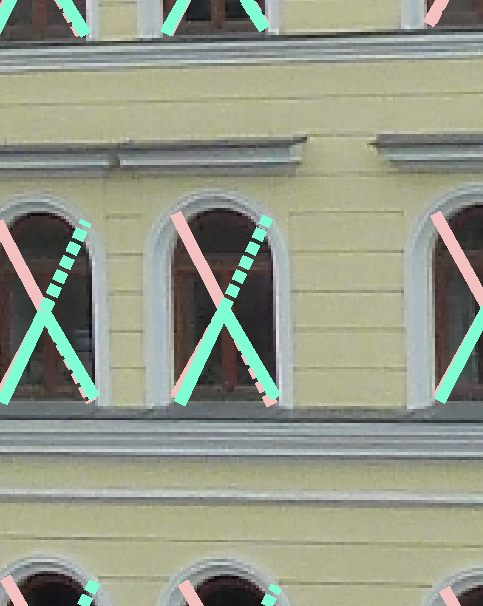}
  }
  \caption[test2]{Repeat Detection,
    Description, and Representation. \begin{enumerate*}[(a)] \item Center of gravity
      (white cross) and curvature extrema (orange circles) of a
      detected MSER (orange contour \cite{Matas-BMVC02}).  Patches are
      normalized to a square and oriented to define an affine frame as
      in \cite{Matas-ICPR02}, \item Bases are reflected for detecting
      axial symmetries.  The RootSIFT transform embeds the local
      texture \cite{Arandjelovic-CVPR12,Lowe-IJCV04}. \item Affine
      frames are mapped to the image.
    \end{enumerate*}
  }
\label{fig:local_features}
\vspace{-5pt}
\end{figure}

\subsection{Local Features and Descriptors}
\label{sec:acregions}
Covariant region detectors are highly repeatable on the same imaged
scene texture with respect to significant changes of viewpoint and
illumination \cite{Mikolajczyk-PAMI04,Mishkin-ECCV18}. Their proven
robustness in the multi-view matching task makes them good candidates
for representing the local geometry of repeated textures. In
particular, we use the Maximally-Stable Extremal Region and
Hessian-Affine detectors \cite{Matas-BMVC02,Mikolajczyk-IJCV04}. The
affine-covariant regions are given by an affine basis (see
\secref{sec:region_parameterization}), equivalently three distinct points,
in the image space \cite{Obdrzalek-BMVC02}. The image patch local to
the affine frame is embedded into a descriptor vector by the RootSIFT
transform \cite{Arandjelovic-CVPR12,Lowe-IJCV04}. See
\figref{fig:local_features} for a visualization.

\begin{figure*}[!ht]
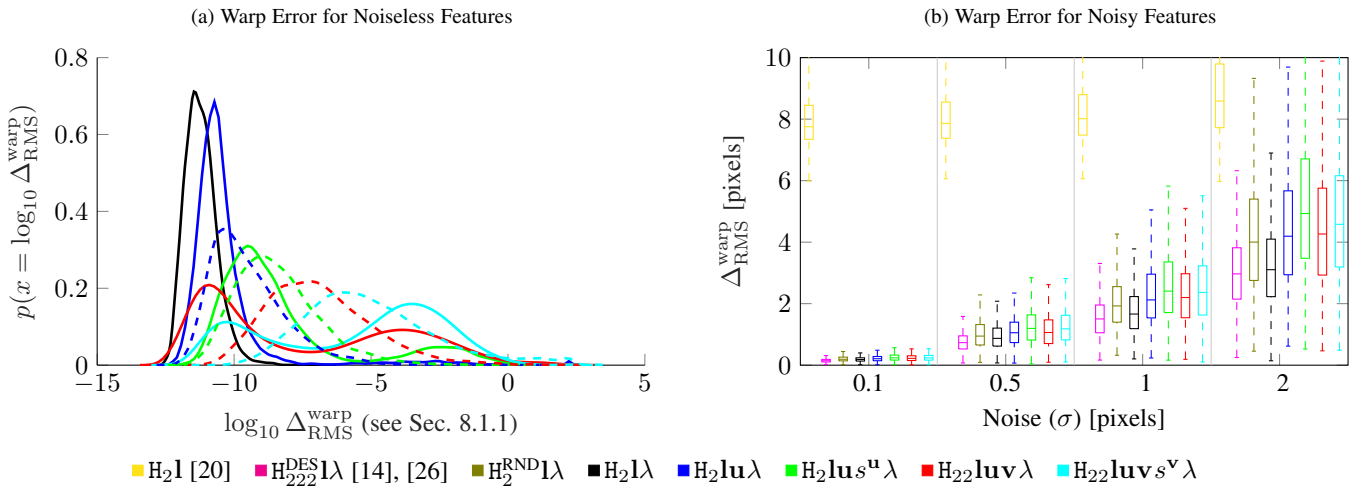

  \captionsetup[subfigure]{oneside,margin={0.75cm,0cm}}
  \subfloat[Warp Error for Noiseless Features \label{fig:warp_error_stability}] {
    \setlength\fwidth{0.4\textwidth}
    \input{fig2/rms_stability.tikz}
  }
  \hfill
  \captionsetup[subfigure]{oneside,margin={1.25cm,0cm}}
  \subfloat[Warp Error for Noisy Features \label{fig:warp_error_sensitivity}] {
    \setlength\fwidth{0.4\textwidth}
    \input{fig2/ransac_rewarp_sensitivity_ct.tikz}
  }
  \centering
  \definecolor{color2}{rgb}{0.0, 1.0, 1.0}
\definecolor{color3}{rgb}{1.0000, 0.5490,  0}
\definecolor{color4}{rgb}{0.5451,0.2706,0.0745}
\definecolor{color5}{rgb}{0,0.5451,0.5451}  
\definecolor{olive}{rgb}{0.5020,0.5020,0}  
\definecolor{betteryellow}{rgb}{1,0.8824,0.0980}
\definecolor{lavender}{rgb}{0.902,0.7451,1.0}

\begin{tikzpicture}
\begin{customlegend}
[legend columns=-1,
legend style={draw=none,/tikz/every even column/.append style={column sep=0.25cm},cells={align=center}},
legend entries={\rgntwoct \cite{Schaffalitzky-BMVC98}, \rgntwotwotwodes \cite{Pritts-ACCV18,Pritts-IJCV20}, \rgntwordctevlrand, \rgntwordctevl, \rgntwordctevp,  \rgntwordsctevp, \rgntwotwordctevp, \rgntwotwordsctevp }]
    \addlegendimage{betteryellow,fill=betteryellow,only marks,mark=square*}
    \addlegendimage{magenta,fill=magenta,only marks,mark=square*}
    \addlegendimage{olive,fill=olive,only marks,mark=square*}            
    \addlegendimage{black,fill=black,only marks,mark=square*}            
    \addlegendimage{blue,fill=blue,only marks,mark=square*}          
    \addlegendimage{green,fill=green,only marks,mark=square*}
    \addlegendimage{red,fill=red,only marks,mark=square*}            
    \addlegendimage{color2,fill=color2,only marks,mark=square*}          
    \end{customlegend}  
\end{tikzpicture}  
  \caption[pami19_stability_and_warp_error]{ \begin{enumerate*}[(a)]
    \item The $\log_{10}$ RMS warp error \rmswarp is reported for
      noiseless scenes generated as described in
      \secsref{sec:warp_error} and \ref{sec:stability}. Hidden variable
      trick solvers are solid; solvers generated without simplified
      constraints equations are dashed.  The hidden-variable trick
      increases stability. The \EVL \rgntwordctevl solver is the most
      stable since it does not require solving a complicated
      polynomial system of equations. \item Reports the RMS error
      \rmswarp after 25 iterations of a simple \RANSAC for the bench
      of solvers with increasing levels of white noise added to the
      affine-covariant region correspondences, where the normalized
      division model parameter is set to -4, which is similar to the
      distortion of a GoPro Hero 4. Results are for radial-distorted
      conjugate translations. The proposed solvers demonstrate
      excellent robustness to noise, and the \EVL solver
      \rgntwordctevl is competitive with \rgntwotwotwodes, which
      requires two more correspondences. The \rgntwordctevl solver
      uses best minimal solution selection, which improves its
      performance compared to \rgntwordctevlrand, which randomly
      selects a solution.
  \end{enumerate*}}
  \label{fig:stabilty_and_warp_error}
\end{figure*}

\subsection{Detection, Description, and Clustering}
\label{sec:clustering_and_sampling}
Affine frames are tentatively labeled as repeated texture by their
appearance. The appearance of an affine frame is given by the RootSIFT
embedding of the image patch local to the affine frame
\cite{Arandjelovic-CVPR12}. Affine-covariant regions are also
extracted and embedded in the reflected image, where the detections in
are transformed into the original image space and have a left-handed
representation.

The RootSIFT descriptors are agglomeratively clustered, which
establishes pair-wise tentative correspondences amongst connected
components. Since the proposed \rgntwordctevp, \rgntwordsctevp,
\rgntwordctevl, and \rgntwotwordctevp solvers do not admit
reflections, the appearance-clusters are partitioned based on the
handedness of the affine frames associated with the clustered embedded
regions. Reflection partitioning is not necessary for the
\rgntwotwordsctevp, which admits reflections of similarity-covariant
regions. Each appearance cluster has some proportion of its indices
corresponding to affine frames that represent the same coplanar
repeated scene content, which are the \emph{inliers} of that
appearance cluster. The remaining affine frames are the
\emph{outliers}.

\subsection{Sampling}
\label{sec:sampling}
Sample configurations for the proposed minimal solvers are illustrated
in \figsref{fig:first}, \ref{fig:input_configurations}, and
\ref{fig:evl_geometry} as well as detailed in
\secsref{sec:evp_constraints} and \ref{sec:evl_constraints}. For each
\RANSAC trial, appearance clusters are selected with the probability
given by its relative cardinality to the other appearance clusters,
and the required number of correspondences are drawn from the selected
clusters.

\subsection{Metric Upgrade and Local Optimization}
\label{sec:metric_upgrade_and_lo}
The affine-covariant regions that are members of the minimal sample
are affine rectified by each feasible model returned by the solver;
typically there is only 1. Correspondences for the selected solver are
sampled as detailed in \secref{sec:sampling}. The affine rectification
estimated by the minimal solver is used to build an affine-rectified
scale consensus set. The scale consensus set is built by using the scale
constraint of affine-rectified space: two instances of
rigidly-transformed coplanar repeats occupy identical areas in the
scene plane and in the affine rectified image of the scene plane
\cite{Criminisi-BMVC00,Chum-ACCV10,Hartley-BOOK04,Pritts-ACCV18,Pritts-IJCV20}.
Note that if clustered left and right-handed regions were partitioned
for sampling with the \rgntwordctevp, \rgntwordsctevp, \rgntwordctevl,
and \rgntwotwordctevp solvers, then they are merged so they are
jointly verified for scale consistency. Absolute scales are calculated
to account for handedness. The log-scale ratio of the each region in a
cluster is computed with respect to the median affine-rectified
scale. Note that covariant regions extracted from imaged
rigidly-transformed coplanar texture can enter the scale consensus set
since they will be equi-scalar after affine rectification, too. This
admits the possibility of a full-metric upgrade. Regions with near 0
log-scale ratio with respect to the median scale of their cluster are
considered tentatively inlying, and are used as inputs to the metric
upgrade of Pritts \etal \cite{Pritts-CVPR14}, which restores
congruence.

The congruence consensus set is measured in the metric-rectified space
by verifying the congruence of the linear basis vectors of the
corresponded affine frames.  Congruence is an invariant of metric
rectified space and is a stronger constraint than, \eg, the
equal-scale invariant of affine-rectified space that was used to
derive the solvers proposed in
\cite{Chum-ACCV10,Pritts-ACCV18,Pritts-IJCV20}.  The metric upgrade
essentially comes for free by inputting the covariant regions that are
members of the scale consensus set to the linear metric-upgrade solver
proposed in \cite{Pritts-CVPR14}. By using the metric upgrade, the
verification step of \RANSAC can enforce the congruence of
corresponding covariant region extents (equivalently, the lengths of
the linear basis vectors) to estimate an accurate consensus set. A
model with the maximal congruence consensus set at the current \RANSAC
iteration is locally optimized in a method similar to
\cite{Pritts-CVPR14}.

%

\section{Experiments}
\label{sec:experiments}
We compare the proposed solvers to the bench of state-of-the-art
solvers listed in \tabref{tab:solver_properties}. We apply the
denotations for the solvers introduced in
\secref{sec:naming_convention} to all tested solvers. Included is the
state-of-the-art joint undistorting and rectifying solver
\rgntwotwotwodes of Pritts \etal \cite{Pritts-ACCV18,Pritts-IJCV20},
which requires 3 correspondences of affine-covariant regions extracted
from the image of rigidly-transformed coplanar repeated scene
textures. While 6 variants of undistorting and rectifying solvers are
proposed in \cite{Pritts-ACCV18,Pritts-IJCV20}, we test only the
\rgntwotwotwodes solver since all variants are reported to have
similar noise sensitivities. Also included is the \rgntwoct solver of
Schaffalitzky \etal \cite{Schaffalitzky-BMVC98}, which incorporates
similar constraints from conjugate translations that are used to
derive the proposed solvers. Two full-homography and
radial-undistortion solvers are included: the \rgntwotwofitz solver of
Fitzgibbon \etal \cite{Fitzgibbon-CVPR01} and the \rgntwotwokukelova
solver of Kukelova \etal \cite{Kukelova-CVPR15}, which are used to
assess the benefits of jointly solving for radially-distorted
conjugate translations (and lens undistortion) from the minimal
problem, as done with the proposed solvers, versus the
over-parameterized problem as in
\cite{Fitzgibbon-CVPR01,Kukelova-CVPR15}. The solvers are evaluated on
synthetic scenes and challenging real images.

\subsection{Synthetic Data}
\label{sec:synthetic_data}
The sensitivity studies evaluate the solvers on noisy measurements
over 3 task-related performance metrics:
\begin{enumerate*}[(i)]
\item the transfer error, which measures the accuracy of
  radially-distorted conjugate translation estimation \item the warp
  error which measures rectification accuracy, and \item the relative
  error of the division-model parameter estimate, which reports the
  accuracy of the lens undistortion estimate.
\end{enumerate*} A solver stability study evaluates the proposed solvers by the warp
error on noiseless measurements. The study demonstrates the benefit of
constraint simplification by the hidden-variable trick, which is used
to derive both the \EVP solvers and \EVL solver, and shows that it
improves the stability of all solvers, and, in fact, it is sometimes
necessary to generate usable solvers \cite{Cox-BOOK05}.

These studies are evaluated on 1000 synthetic images of 3D scenes with
known ground-truth parameters. A camera with a random but realistic
focal length is randomly placed with respect to a scene plane such
that it is mostly in the camera's field-of-view. The image resolution
is set to 1000x1000 pixels.  The noise sensitivities of the solvers
are evaluated on imaged translated coplanar repeats. Affine frames are
generated on the scene plane such that their scale with respect to the
scene plane is realistic. The modeling choice reflects the use of
affine-covariant region detectors on real images. The image is
distorted according to the division model. For the sensitivity
experiments, isotropic white noise is added to the distorted affine
frames at increasing levels.

In addition, the convergence speed of \RANSAC is compared with respect
to the use of each of the proposed and state-of-the-art solvers to
hypothesize models. In particular, convergence is evaluated by
plotting rectification accuracy with respect to trial number.

\begin{figure*}[!ht]
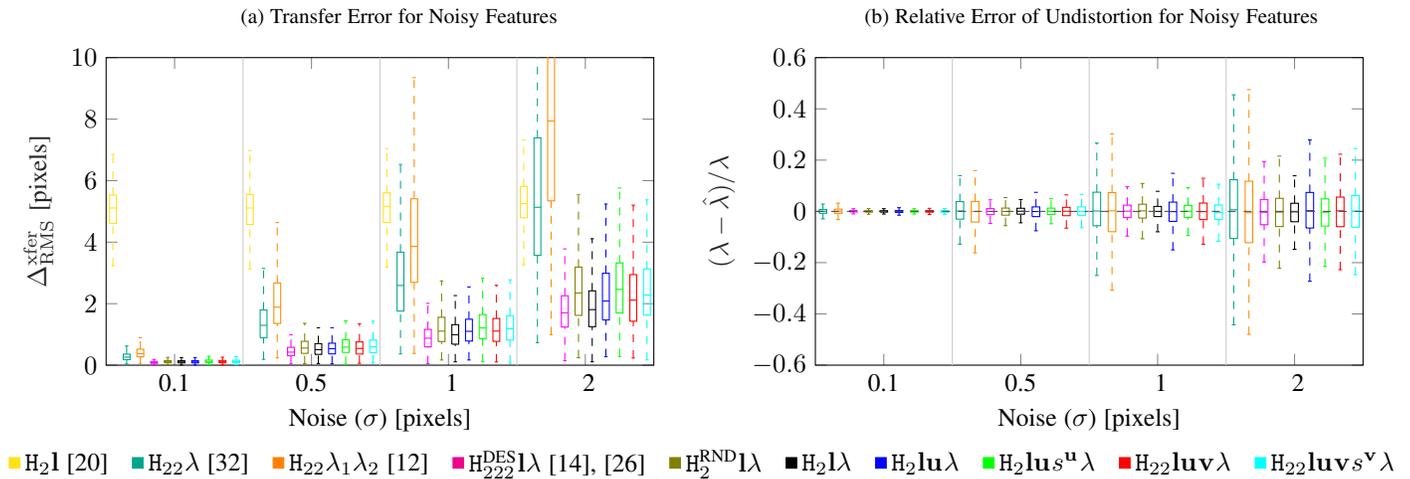

  \captionsetup[subfigure]{oneside,margin={1.8cm,0cm}}
  \subfloat[Transfer Error for Noisy Features\label{fig:ransac_xfer_error_sensitivity}] {
    \setlength\fwidth{0.4\textwidth}
    \input{fig2/ransac_xfer_sensitivity_ct.tikz}
  }
  \hfill
  \captionsetup[subfigure]{oneside,margin={1.75cm,0cm}}
  \subfloat[Relative Error of Undistortion for Noisy Features\label{fig:lambda_rel_error_sensitivity}] {
    \setlength\fwidth{0.4\textwidth}
    \input{fig2/ransac_rel_lambda_sensitivity_ct.tikz}
  }
\centering
\definecolor{color1}{rgb}{0.0, 0.6510, 0.5490}
\definecolor{color2}{rgb}{0.0, 1.0, 1.0}
\definecolor{betteryellow}{rgb}{1,0.8824,0.0980}
\definecolor{olive}{rgb}{0.5020,0.5020,0} 
\definecolor{lavender}{rgb}{0.902,0.7451,1.0}

\begin{tikzpicture}
\begin{customlegend}
[legend columns=-1,
legend style={draw=none,/tikz/every even column/.append style={column sep=0.25cm},cells={align=center}},
legend entries={\rgntwoct \cite{Schaffalitzky-BMVC98},\rgntwotwofitz \cite{Fitzgibbon-CVPR01}, \rgntwotwokukelova \cite{Kukelova-CVPR15},  \rgntwotwotwodes \cite{Pritts-ACCV18,Pritts-IJCV20}, \rgntwordctevlrand, \rgntwordctevl, \rgntwordctevp,  \rgntwordsctevp, \rgntwotwordctevp, \rgntwotwordsctevp }]
    \addlegendimage{betteryellow,fill=betteryellow,only marks,mark=square*}
    \addlegendimage{color1,fill=color1,only marks,mark=square*}
    \addlegendimage{orange,fill=orange,only marks,mark=square*}
    \addlegendimage{magenta,fill=magenta,only marks,mark=square*}
    \addlegendimage{olive,fill=olive,only marks,mark=square*}            
    \addlegendimage{black,fill=black,only marks,mark=square*}            
    \addlegendimage{blue,fill=blue,only marks,mark=square*}          
    \addlegendimage{green,fill=green,only marks,mark=square*}
    \addlegendimage{red,fill=red,only marks,mark=square*}            
    \addlegendimage{color2,fill=color2,only marks,mark=square*}          
    \end{customlegend}  
\end{tikzpicture}  
\caption[pami19_warp_and_undistortion_error]{Comparison of two error
  measures after 25 iterations of a simple \RANSAC for different
  solvers with increasing levels of white noise added to the affine
  covariant region correspondences, where the normalized division
  model parameter is set to -4 (see Sec. 3.1), which is similar to the
  distortion of a GoPro Hero 4. Results are for imaged translated
  coplanar repeats: \begin{enumerate*}[(a)] \item Reports the root
    mean square transfer error \rmsxfer. With the exception of the
    \rgntwotwotwodes solver, the proposed solvers are significantly
    more robust for both types of repeats on both error measures;
    however \rgntwotwotwodes requires the most correspondences,
    and \item reports the relative error of the estimated division
    model parameter.\end{enumerate*} The \rgntwordctevl solver uses
  best minimal solution selection, which improves its performance
  compared to \rgntwordctevlrand, which randomly selects a solution.}
\label{fig:ransac_sensitivity_study}
\end{figure*}

\subsubsection{Warp Error}
\label{sec:warp_error}
Since the accuracy of scene-plane rectification is a primary concern,
an extended version of the warp error for rectifying homographies
proposed by Pritts \etal~\cite{Pritts-BMVC16} that incorporates the
division model for radial lens distortion of
Fitzgibbon~\cite{Fitzgibbon-CVPR01} is used to evaluate a solver's
stability and robustness to noise. A scene plane is tessellated by a
10x10 square grid of points $\buildset{\vX[i]}{i=1}{100}$ and imaged
as $\buildset{\vxd[i]}{i=1}{100}$ by the lens-distorted ground-truth
camera. The tessellation ensures that error is uniformly measured over
the scene plane. A round trip between the image space and rectified
space is made by affine-rectifying $\buildset{\vxd[i]}{i=1}{100}$
using the estimated division model parameter $\hat{\lambda}$ and
rectifying homography $\mH(\vlhat)$ (see \eqref{eq:recthg}) and then
imaging the rectified plane by the ground-truth camera \mP. Ideally,
the ground-truth camera \mP images the rectified points
$\buildset{\vxr[i]}{i=1}{100}$ onto the distorted points
$\buildset{\vxd[i]}{i=1}{100}$.  There is an affine ambiguity, denoted
\mA, between $\mH(\vlhat)$ and the ground-truth camera matrix \mP. The
ambiguity is estimated during computation of the warp error,
\begin{equation} 
  \label{prg:warp_residual} \Delta^{\mathrm{warp}}=\min_{\mA} \sum_{i}
  d^2(\vxd[i],f^d(\mP\mA\mH(\vlhat) f(\vxd[i],\hat{\lambda})),\lambda),
\end{equation}
where $d(\cdot,\cdot)$ is the Euclidean distance, $f^d$ is the inverse
of the division model (the inverse of \eqref{eq:division_model}).
The root mean square warp error for
$\buildset{\vxd[i]}{i=1}{100}$ is reported and denoted as
$\Delta^{\mathrm{warp}}_{\mathrm{RMS}}$.

Note that the \rgntwotwofitz solver of \cite{Fitzgibbon-CVPR01} and
the \rgntwotwokukelova solver of \cite{Kukelova-CVPR15} are omitted
from the warp error since the vanishing line is not directly
estimated.

\subsubsection{Transfer Error}
\label{sec:xfer_error}
The geometric transfer error measures the accuracy of the estimated
radially-distorted conjugate translation.  The scene plane is
tessellated by a 10x10 grid of points spaced one unit apart.  The
tessellation ensures that the geometric transfer error is uniformly
sampled across the image. Denote the tessellation as
\buildset{\vX[i]}{i=1}{100}. Suppose that \xcspond are
conjugately-translated points that are consistent with
$\mH_{\vu}=[\ma{I}_3+\vu\vl^{\T}]$. The preimage \vU of the vanishing
translation direction \vu is recovered as
$\beta\ve[U]=\mP^{\inv}\vu=\beta\rowvec{3}{u_x}{u_y}{0}^{\T}$.  The
tessellation is translated by $\vU/\|\ve[U]\|$ on the scene plane. Let
$\ma{T}(\vU/\|\ve[U]\|)$ be a homogeneous translation matrix as
defined in \eqref{eq:conjugate_translation}.  Then by
\eqref{eq:decomposition} the image of the unit-magnitude translation
on the scene plane in the direction of correspondences \xcspond is
\begin{equation}
  \mH_{\vu/\|\ve[U]\|} = [\ma{I}_3+\frac{\vu}{\|\ve[U]\|}\vl^{\T}].
  \label{eq:scaled_transfer_error}
\end{equation}

The unit conjugate translation $\mH_{\vu/\|\ve[U]\|}$ can be written
in terms of $\mH_{\vu}$ as
\begin{equation}
 \label{eq:transfer_error_derived}
 \begin{split}
    \ma{I}_3+\frac{\vu}{\|\ve[U]\|}\vl^{\T} &=
    \ma{I}_3+\frac{1}{\|\ve[U]\|}[\ma{I}_3+\vu\vl^{\T}-\ma{I}_3] \\ &=
    \ma{I}_3+\frac{1}{\|\ve[U]\|}[\mH_{\vu}-\ma{I}_3].
 \end{split}
\end{equation}
The points are distorted with the ground-truth division-model parameter
$\lambda$ as $\vxd[i] = f^d(\vx[i],\lambda)$ and $\vxdp[i]
=f^d(\vxp[i],\lambda)$, where $f^d$ transforms from pinhole points to
radially-distorted points.  Then the geometric transfer error is
defined as
\begin{equation}
  \label{eq:transfer_error} 
  \Delta^{\mathrm{xfer}} = d(f^d([\ma{I}_3 +
    \frac{1}{\|\vU\|}(\hat{\mH}_{\vu} - \ma{I}_3)]
  f(\vxd,\hat{\lambda}_1),\hat{\lambda}_2),\vxdp),
\end{equation} where
$d(\cdot,\cdot)$ is the Euclidean distance and $\hat{\mH}_{\vu}$ and
$\hat{\lambda}_1,\hat{\lambda}_2$ are the estimated conjugate
translations and division model parameters, respectively. All solvers except \rgntwotwokukelova have the constraint that
$\hat{\lambda}_1=\hat{\lambda}_2$~\cite{Kukelova-CVPR15}.

The root mean square transfer error \rmsxfer for radially-distorted
conjugately-translated correspondences \xdcspond[i] is reported. For
two-direction solvers, the transfer error in the second direction is
included in \rmsxfer. The transfer error is used in the sensitivity
study, where the solvers are tested over varying noise levels with a
fixed division model parameter.

\subsubsection{Numerical Stability}
\label{sec:stability}
The stability study measures the RMS warp error \rmswarp of solvers
for noiseless radially-distorted conjugately-translated affine frame
correspondences across realistic scene and camera configurations
generated as described in the introduction to this section. The
normalized ground-truth division-model parameter$\lambda$ is drawn
uniformly at random from the interval $[-6,0]$. For a reference, the
division parameter of $\lambda=-4$ is typical for wide field-of-view
cameras like the GoPro Hero 4, where the image is normalized by
${1}/({\text{width}+\text{height}})$. \figref{fig:warp_error_stability}
reports the histogram of $\log_{10}$ warp errors \rmswarp.

For the proposed \EVP solvers we evaluate a solver generated from
constraints derived with (solid histogram) and without (dashed
histogram) the hidden-variable trick. The hidden-variable trick
significantly improves the stability of the proposed solvers.  The
increased stabilities of the hidden-variable solvers most likely
result from the reduced size of the Gauss-Jordan elimination problems
needed by these solvers. The hidden-variable \EVP solvers are used for
the remainder of the experiments. The proposed \EVL solver
\rgntwordctevl is derived with the hidden-variable trick as well,
which results in a quartic. The superior stability of the
\rgntwordctevl solver demonstrates the benefits of the elementary
formulation.

\subsubsection{Noise Sensitivity}
\label{sec:noise_sensitivity}
The accuracy of the proposed and state-of-the-art solvers is measured
by the warp error, transfer error, and relative error of lens
undistortion with respect to increasing levels of white noise added to
radially-distorted conjugately-translated point correspondences.  The
amount of white noise is given by the standard deviation of a
zero-mean isotropic Gaussian distribution, and the solvers are tested
at noise levels of $\sigma \in \buildset{ 0.1,0.5,1,2 }{}{}$. The
ground-truth normalized division model parameter is set to
$\lambda=-4$, which is typical for GoPro-type imagery in normalized
image coordinates.

\newcommand{\imgcolumn}[6]{
\begin{minipage}{#1\textwidth}
  \centering
  \footnotesize
  #2\\[2pt]
  $#3\%$ inliers\\[2pt]
  $\hat\lambda = #4$\\[5pt]
  \includegraphics[width=\textwidth]{#5}\\
  \includegraphics[width=\textwidth]{#6}
\end{minipage}
}

\definecolor{block-gray}{gray}{0.9}
\newtcolorbox{highlight}{arc=0pt,colback=block-gray,boxrule=0pt,boxsep=0pt,breakable,left=1.5pt,right=1.5pt,top=1.5pt,bottom=1.5pt,sidebyside gap=1pt}

\newcommand{\imgcolumnh}[6]{
\begin{minipage}{#1\textwidth}
  \begin{highlight}
  \centering
  \footnotesize
  #2\\[2pt]
  $#3\%$ inliers\\[2pt]
  $\hat\lambda = #4$\\[5pt]
  \end{highlight}
  \includegraphics[width=\textwidth]{#5}\\
  \includegraphics[width=\textwidth]{#6}
\end{minipage}
}
\newcommand{\rulesep}{\quad\unskip\ \vrule\ ~}

\begin{figure*}[t!]
\centering
\begin{minipage}{0.115\textwidth}
\centering
\footnotesize GoPro Hero 4 Wide, 17.2mm\\[2pt]
$\lambda_{\text{gt}} = -0.2823$\\[5pt]
\includegraphics[width=\textwidth]{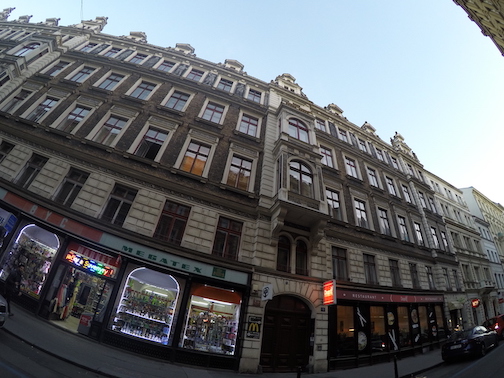}\\
\includegraphics[width=\textwidth]{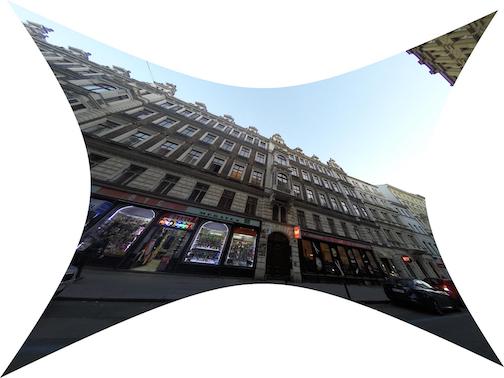}
\end{minipage}
\rulesep
\imgcolumn{0.115}{\rgntwotwoct \cite{Pritts-CVPR14}}{1.2}{-0.0326}
{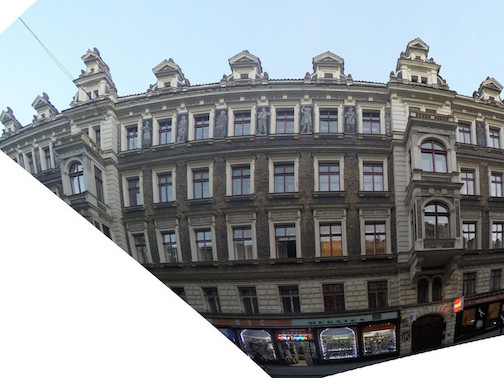}
{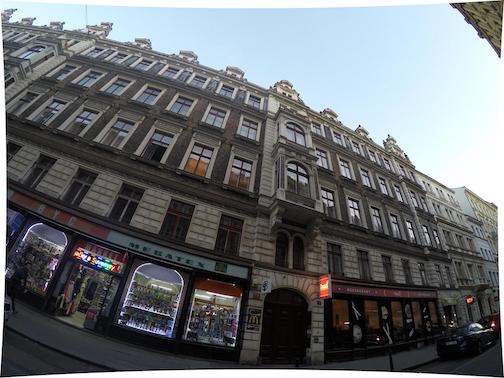}
\hspace{-8pt}
\imgcolumn{0.115}{\rgntwotwotwodes \cite{Pritts-ACCV18,Pritts-IJCV20}}{2.1}{-0.3479}
{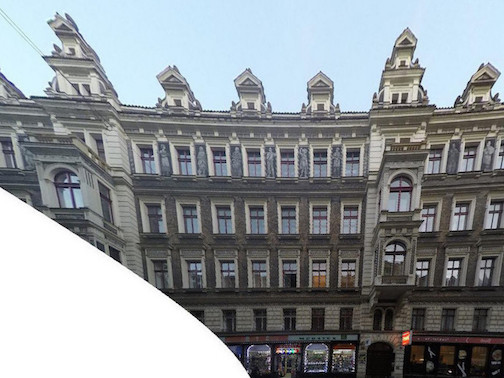}
{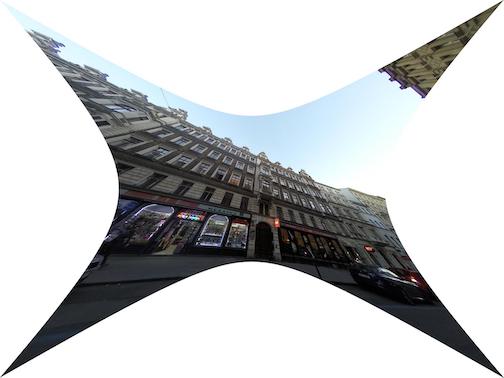}
\imgcolumnh{0.115}{\rgntwordctevl}{2.6}{-0.3086}
{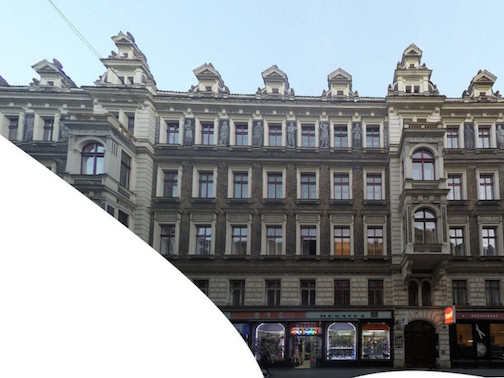}
{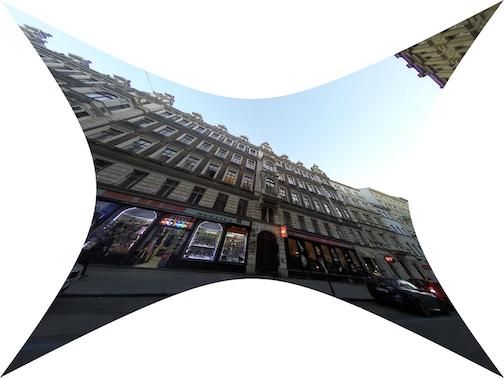}
\hspace{-8pt}
\imgcolumnh{0.115}{\rgntwordctevp}{2.4}{-0.2501}
{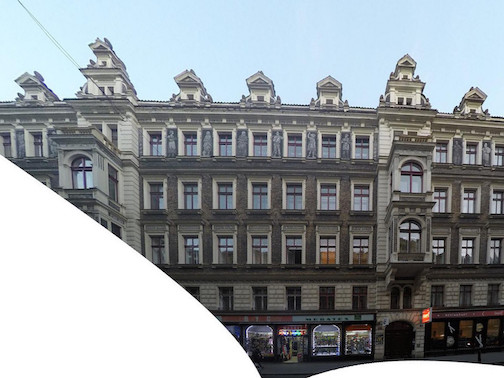}
{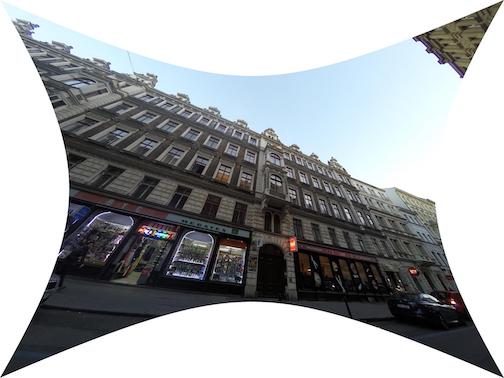}
\hspace{-8pt}
\imgcolumnh{0.115}{\rgntwordsctevp}{2.2}{-0.2651}
{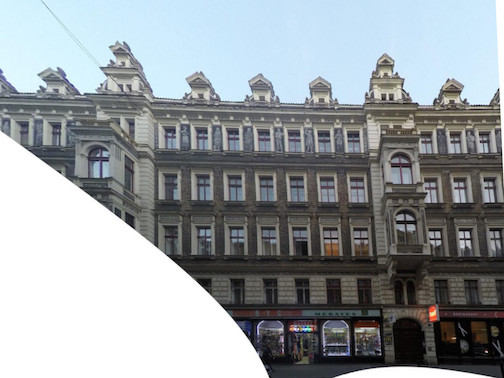}
{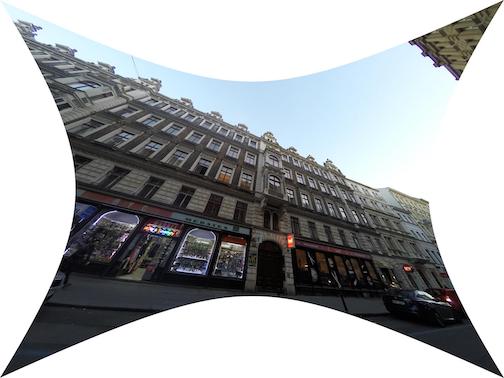}
\hspace{-8pt}
\imgcolumnh{0.115}{\rgntwotwordctevp}{1.9}{-0.2674}
{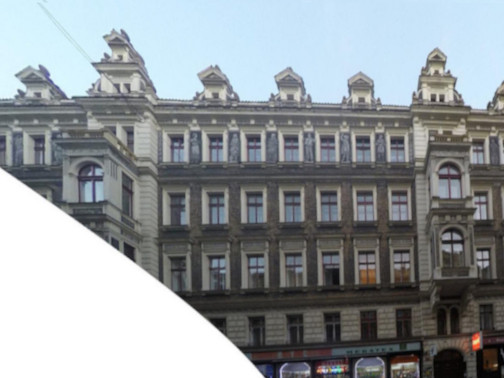}
{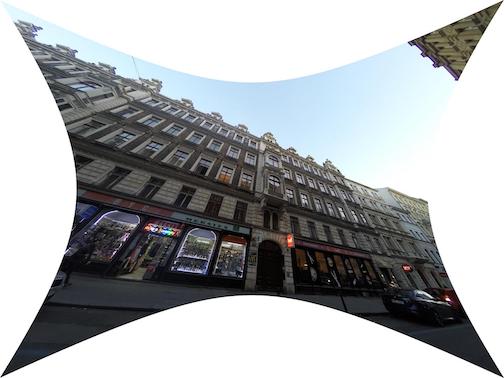}
\hspace{-8pt}
\imgcolumnh{0.115}{\rgntwotwordsctevp}{2.0}{-0.3118}
{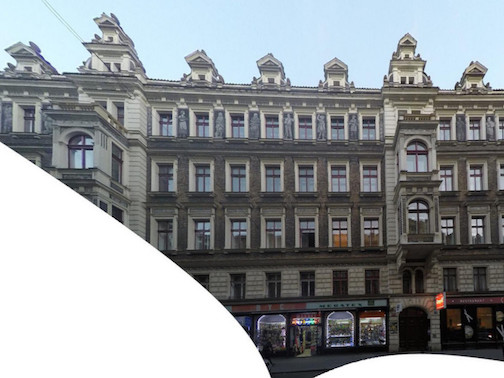}
{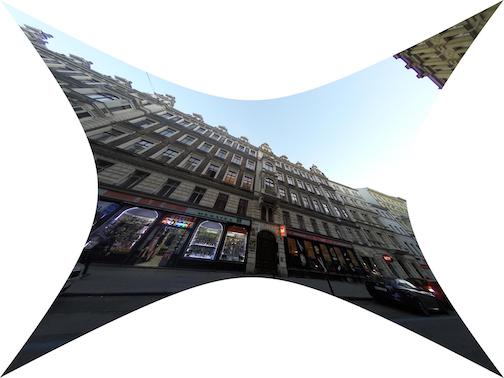}
\caption{The bench of solvers is evaluated on an image with poorly
  corresponded repeated regions. The leftmost column shows the
  original image (top) and its undistortion using ground truth
  (bottom) for comparison. The proposed solvers are highlighted in
  gray. The estimated rectifications and undistorted images are
  synthesized, and the normalized values of the division-model
  parameter and ratio of inliers are reported. The \rgntwordctevl
  solver gives the best result, and all proposed solvers perform
  better than the state-of-the-art.}
\label{fig:real_wide}
\vspace{-5pt}
\end{figure*}

The solvers are wrapped by a basic \RANSAC estimator that minimizes
either the RMS warp error \rmswarp (see
\figref{fig:warp_error_sensitivity}), the RMS transfer error (see
\figref{fig:ransac_xfer_error_sensitivity}) \rmsxfer, or the relative
error of lens distortion (see
\figref{fig:lambda_rel_error_sensitivity}) over 25 minimal samples of
affine frames. The \RANSAC estimates are summarized in boxplots for
1000 synthetic scenes. The interquartile range is contained within the
extents of a box, and the median is the horizontal line dividing the
box. 

The proposed
solvers---\rgntwordctevp,\rgntwordsctevp,\rgntwotwordctevp,\rgntwotwordsctevp,
and \rgntwordctevl---demonstrate excellent robustness to noisy
features across all three error measures. In particular, the
\rgntwordctevl solver is the least sensitive to noise of the proposed
solvers and gives the best undistortion estimates of any solver in the
bench (see \figref{fig:lambda_rel_error_sensitivity}). All proposed
solvers estimate the correct lens distortion parameter more than half
the time. \figref{fig:warp_error_sensitivity} shows that at the 2
pixel noise level, all the proposed solvers rectify with less than 5
pixel RMS warp error \rmswarp more than half the
time. \figref{fig:ransac_xfer_error_sensitivity} shows that
radially-distorted conjugate translations are estimated with less than
3 pixel RMS transfer error \rmsxfer error more than half the time.

For both the warp error and transfer error studies, the \rgntwoct
solver of Schaffalitzky \etal \cite{Schaffalitzky-BMVC98} shows
significant bias since it does not model lens distortion, making it
essentially unusable as a minimal solver at GoPro-like levels of
radial lens distortion.  As expected, the overparameterized
radial-distortion homography solvers of \rgntwotwofitz
\cite{Fitzgibbon-CVPR01} Fitzgibbon and \rgntwotwokukelova
\cite{Kukelova-CVPR15} of Kukelova \etal have significantly higher
transfer errors with respect to the proposed solvers, which suggests
that the extraneous degrees of freedom are used to explain feature
noise by incorrect geometry. In fact, at the two pixel noise level of
the transfer error study in
\figref{fig:ransac_xfer_error_sensitivity}, the performance of these
solvers is worse than the \rgntwoct solver, which does not model
radial lens distortion.

The state-of-the art solver \rgntwotwotwodes of
Pritts~\etal~\cite{Pritts-ACCV18,Pritts-IJCV20} shows slightly better
noise robustness than the proposed solvers on the warp and transfer
error sensitivity studies. However, the proposed solvers are
competitive and require fewer correspondences.  In particular, the
\rgntwordctevl reaches near parity with the \rgntwotwotwodes solver
and requires only one region correspondence versus three required by the
\rgntwotwotwodes solver. As is shown in \secref{sec:wall_clock}, the
proposed solvers are magnitudes faster in wall clock time. Given their
competitive performance in the sensitivity studies and the fact that
they require fewer correspondences and have faster times to solution,
the proposed solver should be preferred to the \rgntwotwotwodes solver
for images with radially-distorted conjugate translations.

Each of the \rgntwordctevl and \rgntwotwotwodes solvers requires the
ex-post estimation of vanishing point of the translation direction,
which is computed by the method proposed in
\secref{sec:unknown_vp_estimation}. Surprisingly, the sequential
estimation used by the proposed \rgntwordctevl and the
\rgntwotwotwodes solver of \cite{Pritts-ACCV18,Pritts-IJCV20} achieve
the best performances on the transfer error \rmsxfer. This is
explainable by the improved performance of the \rgntwordctevl \EVL
solver with respect to the \EVP solvers on all measures, and the fact
that the \rgntwotwotwodes solver uses three correspondences, the most of
any in the bench of solvers (see \tabref{tab:solver_properties}).

The benefit of best minimal solution selection as proposed in
\eqref{sec:choosing_constraint_eqs} can be seen by comparing the
\rgntwordctevlrand and \rgntwordctevl solvers in all sensitivity
studies. To quickly recap, The \rgntwordctevlrand solver randomly
selects a minimal solution from 10 possible solutions given by the
\EVL geometry shown in \figref{fig:evl_geometry}, while the
\rgntwordctevl chooses the solution that minimizes a geometric error
on the unused constraints. The sensitivity improvements using best
minimal solution selection are considerable: at the 2 pixel noise
levels, the RMS warp error \rmswarp
(\figref{fig:warp_error_sensitivity}) and RMS transfer error
(\figref{fig:ransac_xfer_error_sensitivity}) decreased by $26\%$ and
$28\%$, respectively, and the interquartile range of division model
parameter estimates decreased by $61\%$. In fact, the incorporation of
best minimal solution selection puts the performance of the
\rgntwordctevl solver on par with the \rgntwotwotwodes solver, which
requires two more region
correspondences. 

%

\subsubsection{\RANSAC convergence study}
\label{sec:ransac_convergence}
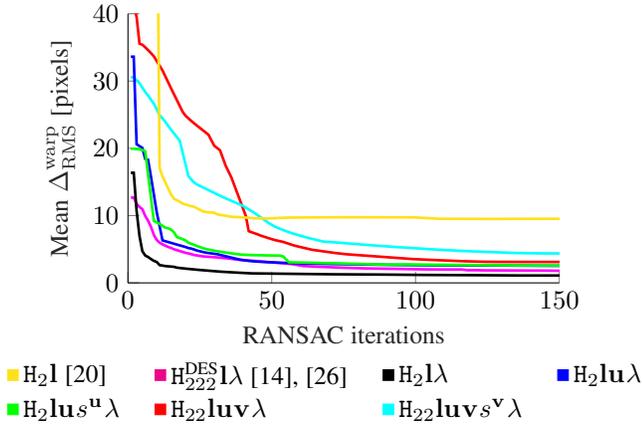
\begin{figure}[!t]
\centering
    \setlength\fwidth{0.35\textwidth}
%
%
\definecolor{mycolor1}{rgb}{1.00000,0.00000,1.00000}%
\definecolor{mycolor2}{rgb}{0.00000,1.00000,1.00000}%
\definecolor{mycolor3}{rgb}{1.00000,0.88240,0.09800}%
\begin{tikzpicture}

\begin{axis}[%
width=0.9\fwidth,
height=0.562\fwidth,
at={(0\fwidth,0\fwidth)},
scale only axis,
xmin=0,
xmax=150,
xlabel style={font=\color{white!15!black}},
xlabel={RANSAC iterations},
ymin=0,
ymax=40,
ylabel style={font=\color{white!15!black}},
ylabel={Mean $\Delta^{\mathrm{warp}}_{\mathrm{RMS}}$ [pixels]},
axis background/.style={fill=white},
axis x line*=bottom,
axis y line*=left,
enlargelimits=false
]
\addplot [color=mycolor1, forget plot, line width=1.0pt]
  table[row sep=crcr]{%
1	12.6823195644199\\
2	12.6823195644199\\
3	11.739817290862\\
4	11.2827216664956\\
5	10.9471120432453\\
6	9.94364146822355\\
7	9.23149535716917\\
8	8.32377202289402\\
9	7.12006795758651\\
10	6.3842507750675\\
11	5.97307317608431\\
12	5.73106453963612\\
13	5.52123512592098\\
14	5.33878806983193\\
15	5.19040871567447\\
16	5.00235241991641\\
20	4.4691557544206\\
21	4.38401125950344\\
23	4.15092160996539\\
26	3.93770097936775\\
31	3.78270614334892\\
35	3.62700993986576\\
38	3.45779354829781\\
45	3.23725387231582\\
52	2.98126764805983\\
57	2.79532809168737\\
61	2.50590490952308\\
66	2.40038859886937\\
75	2.24946471322642\\
82	2.15946151420422\\
103	2.00827831665356\\
111	1.98660502826152\\
116	1.99700708341965\\
118	1.88582859701256\\
122	1.88283178851407\\
123	1.86181302638485\\
144	1.83087594734204\\
149	1.8053320437823\\
151	1.8053320437823\\
};
\addplot [color=red, forget plot, line width=1.0pt]
  table[row sep=crcr]{%
1	41.2283688169217\\
2	41.2283688169217\\
3	39.5229484475796\\
4	35.5635551866313\\
5	35.3965328755414\\
6	35.0093298341937\\
7	34.5892830535046\\
8	34.1464441246574\\
9	33.6688707842827\\
12	31.5810997271889\\
15	28.9059103804189\\
19	25.4386950877073\\
20	24.8513056171936\\
21	24.4553703145764\\
23	23.7151462779835\\
28	22.0185169614776\\
30	20.3844256825977\\
32	19.717118943921\\
34	17.3915193110431\\
36	15.6285290932033\\
40	11.0872723708467\\
41	9.93363236781926\\
42	7.67668061634291\\
45	7.22846617059534\\
46	7.05291269048124\\
49	6.6332323428702\\
51	6.3743708992223\\
54	6.08579971465014\\
57	5.65781399543641\\
58	5.52618365014249\\
64	4.9217411669018\\
72	4.44466983142084\\
76	4.25456841194526\\
99	3.54212732026818\\
105	3.42200375711846\\
117	3.21529654794406\\
124	3.13573118928335\\
128	3.12204197619809\\
143	3.11111479949793\\
151	3.11111479949793\\
};
\addplot [color=mycolor2, forget plot, line width=1.0pt]
  table[row sep=crcr]{%
1	30.5592180415938\\
2	30.5592180415938\\
4	29.6716766134911\\
6	28.368113499247\\
7	27.8348492375243\\
9	26.6369477168142\\
11	24.9148318447997\\
12	24.4499082148766\\
13	23.9372187925951\\
14	23.3286421169963\\
15	22.7492309854035\\
18	21.1381507980098\\
21	15.8828522564399\\
22	15.366160222994\\
23	14.9300757407286\\
24	14.6193489324669\\
27	13.9998178408622\\
31	13.1686301017795\\
33	12.7510661604431\\
40	11.4475577803047\\
44	10.3910680282698\\
46	9.79683963695629\\
49	8.88644078655122\\
51	8.37803006921376\\
54	7.83938042049016\\
57	7.38329911155023\\
62	6.7603921935831\\
68	6.13000440294704\\
71	6.05328553112042\\
84	5.61276560731832\\
92	5.36571779473243\\
108	4.92787309545727\\
114	4.7939527632536\\
119	4.69532651532688\\
135	4.43974256484438\\
144	4.38231038116197\\
151	4.37486387530029\\
};
\addplot [color=blue, forget plot, line width=1.0pt]
  table[row sep=crcr]{%
1	33.6166162388286\\
2	33.6166162388286\\
3	20.601073796141\\
5	20.0654396519887\\
6	18.4284790042482\\
7	18.2865777018606\\
8	15.7680686581914\\
10	10.5016102772182\\
11	8.43615507723473\\
12	6.3151555134607\\
18	5.57327517427339\\
20	5.33631794008181\\
21	5.17424132660548\\
22	5.10196223709411\\
23	4.96495951190315\\
24	4.8009207799206\\
27	4.49892435122467\\
31	4.2674750602805\\
39	3.44739729952286\\
42	3.27095312114321\\
44	3.19411000731529\\
47	3.11317322478121\\
51	3.03437434128205\\
60	2.89416542383023\\
71	2.80545964027962\\
93	2.66768473096849\\
105	2.62404529297157\\
126	2.58402573769268\\
147	2.54200420040573\\
151	2.54200420040573\\
};
\addplot [color=green, forget plot, line width=1.0pt]
  table[row sep=crcr]{%
1	19.9363198225805\\
2	19.9363198225805\\
4	19.8523373628204\\
5	19.7443248576885\\
6	19.5628524828583\\
8	12.5971072364873\\
9	9.19189466598999\\
11	8.77835241746621\\
13	8.2109566461194\\
14	8.10651939530533\\
15	7.91855086393926\\
16	7.51634091608585\\
17	6.81216080173084\\
18	6.66809685796349\\
20	6.318039207033\\
21	5.9794684383572\\
24	5.47598801552954\\
26	5.17633144950216\\
27	5.04380278606916\\
29	4.86940121822983\\
32	4.64174330766824\\
34	4.50724322086774\\
39	4.21745912725765\\
43	4.12963132353676\\
53	4.05034453902192\\
54	3.91633754038207\\
56	3.06565914219541\\
74	2.89869484008844\\
88	2.72966836948336\\
92	2.72196026627091\\
97	2.74936129104231\\
107	2.69465237607639\\
142	2.5738394422672\\
151	2.57585251350176\\
};
\addplot [color=mycolor3, forget plot, line width=1.0pt]
  table[row sep=crcr]{%
10.237160198044	55\\
11	17.0817829156506\\
12	15.8199698238806\\
13	14.8946880573601\\
15	13.1309097166718\\
16	12.5126979690394\\
18	12.0606919472572\\
19	11.8065727316005\\
20	11.652892411194\\
22	11.4139877296093\\
23	11.3046099489873\\
24	11.0904491537804\\
26	10.6418842446109\\
28	10.4776160656591\\
30	10.4400524167335\\
31	10.19194304066\\
32	10.1170142327327\\
33	10.0631371530136\\
35	9.90102187494037\\
36	9.81422508613875\\
38	9.77597087379152\\
39	9.76594719619516\\
45	9.58221924085899\\
48	9.57714417677596\\
50	9.58689833702178\\
53	9.66512605099211\\
56	9.70571883833031\\
70	9.74723444538427\\
72	9.75628881536767\\
99	9.74159484405112\\
109	9.52070192994569\\
127	9.51064595412848\\
151	9.53245445714774\\
};
\addplot [color=black, forget plot, line width=1.0pt]
  table[row sep=crcr]{%
1	16.368274465895\\
2	16.368274465895\\
3	10.6807142379417\\
4	6.97975156459185\\
5	4.71109106038764\\
6	4.06543694520181\\
7	3.84497771560169\\
8	3.53770967919706\\
9	3.27629842580134\\
10	3.10826620810354\\
11	2.64471116019692\\
13	2.49819998352527\\
16	2.38467908717959\\
17	2.26432963619473\\
20	2.12003957703439\\
25	1.91854516338853\\
32	1.67270537391059\\
37	1.54931718832017\\
43	1.41375755151017\\
71	1.27803012518433\\
89	1.23405933816403\\
102	1.18276722962665\\
122	1.14051944837431\\
134	1.10693523794109\\
137	1.09960449053011\\
151	1.09816291324154\\
};
\end{axis}
\end{tikzpicture}%
\definecolor{color1}{rgb}{0.0, 0.6510, 0.5490}
\definecolor{color2}{rgb}{0.0, 1.0, 1.0}
\definecolor{betteryellow}{rgb}{1,0.8824,0.0980}
\definecolor{olive}{rgb}{0.5020,0.5020,0} 
\definecolor{lavender}{rgb}{0.902,0.7451,1.0}

\begin{tikzpicture}
\begin{customlegend}
[legend columns=4,
legend style={draw=none,/tikz/every even column/.append style={column sep=0.4cm}},
legend cell align={left},
legend entries={\rgntwoct \cite{Schaffalitzky-BMVC98},  \rgntwotwotwodes \cite{Pritts-ACCV18,Pritts-IJCV20}, \rgntwordctevl, \rgntwordctevp,  \rgntwordsctevp, \rgntwotwordctevp, \rgntwotwordsctevp }]
    \addlegendimage{betteryellow,fill=betteryellow,only marks,mark=square*}
    \addlegendimage{magenta,fill=magenta,only marks,mark=square*}
    \addlegendimage{black,fill=black,only marks,mark=square*}            
    \addlegendimage{blue,fill=blue,only marks,mark=square*}          
    \addlegendimage{green,fill=green,only marks,mark=square*}
    \addlegendimage{red,fill=red,only marks,mark=square*}            
    \addlegendimage{color2,fill=color2,only marks,mark=square*}          
    \end{customlegend}  
\end{tikzpicture}  
\caption[mean_warp_error]{Each solver is used to generate models for
  \RANSAC on a sets of synthetic noisy region correspondences with
  50\% outliers. The mean RMS warp error \rmswarp over all scenes at
  each iteration is shown. The \rgntwordctevl gives the most accurate
  rectifications the fastest.}
\label{fig:trials_to_solution}
\end{figure}

The speed of convergence of \RANSAC is evaluated using each of the
proposed and state-of-the-art rectifying solvers to hypothesize
models. Each \RANSAC variant is run on a set of thirty synthetic
scenes generated as described in \secref{sec:synthetic_data} with
one-pixel white noise and with corresponded sets of regions corrupted
by 50\% outliers. \figref{fig:trials_to_solution} reports the mean RMS
warp error \rmswarp over all scenes at each iteration for the bench of
solvers. The proposed \rgntwordctevl solver converges fastest, which
demonstrates its robustness to noise and the advantage of
one-correspondence sampling combined with best minimal-solution
selection. The \rgntwotwotwodes performs second best, despite
requiring three region correspondences; however, it not so surprising
since its robustness is known (\eg, see
\figref{fig:ransac_sensitivity_study} and
\cite{Pritts-ACCV18,Pritts-IJCV20}). As expected, the solver of
\cite{Schaffalitzky-BMVC98} performs poorly since it does not estimate
lens undistortion. The solvers of
\cite{Fitzgibbon-CVPR01,Kukelova-CVPR15} are omitted from this study
since they do not directly recover the vanishing line. In theory, it
is possible to use a rank-one decomposition to recover the estimated
vanishing line from a full homography for warp error computation, but
we found this to give poor results.

\subsection{Computational Complexity}  
\label{sec:wall_clock}
\tabref{tab:wall_clock} lists the wall-clock time to solution for the
optimized C++ implementations of the proposed solvers and the
\rgntwotwotwodes solver \cite{Pritts-ACCV18,Pritts-IJCV20}, which was
the only competitive solver from the sensitivity experiments reported
in \figsref{fig:ransac_xfer_error_sensitivity},
\ref{fig:warp_error_sensitivity}, and
\ref{fig:lambda_rel_error_sensitivity}. Also reported for easy
comparison are the relative speeds with respect to the \rgntwordctevl
solver and the elimination template sizes, where applicable. The
proposed \EVL \rgntwordctevl solver is an astounding $2153.6 \times$
faster than the \rgntwotwotwodes solver and significantly faster than
all \EVP solvers (\rgntwordctevp,\rgntwordsctevp,\rgntwotwordctevp,
and \rgntwotwordsctevp), which require the \Gb method to solve
polynomial systems of equations. All of the proposed solvers are much
faster than the \rgntwotwotwodes solver, making them more suitable for
fast sampling in \RANSAC for scenes containing translational
symmetries.

\begin{table}[H]
  \centering
  \caption{Runtime Analysis}
  \vspace{-7pt}
  \ra{1}
   \resizebox{\columnwidth}{!} {
    \begin{tabular}{@{} rC{15ex}C{15ex}C{15ex} @{} }
      \toprule
      Solver & Wall Clock & Relative Speed & Template Size \\
      \midrule
      \rgntwordctevl & $\mathbf{0.5}$ \textbf{{\textmu}s} & $\mathbf{1.0 \times}$ & N/A  \\
      \rgntwordctevp & $3.7$ {\textmu}s & $7.4 \times$ & $14 \times 18$  \\
      \rgntwordsctevp & $6.1$ {\textmu}s & $12.2 \times$ & $24 \times 26$ \\ 
      \rgntwotwordctevp & $34.6$ {\textmu}s & $69.2 \times$ & $54 \times 60$  \\
      \rgntwotwordsctevp & $66.1$ {\textmu}s  & $132.2 \times$ & $76 \times 80$  \\
      \rgntwotwotwodes \cite{Pritts-ACCV18,Pritts-IJCV20} & $1076.8$ {\textmu}s & $2153.6 \times$ & $133 \times 187$  \\
      \bottomrule
    \end{tabular}
   }
  \tabcap{Wall-clock times are reported for
    optimized C++ implementations of the proposed solvers versus
    \rgntwotwotwodes of \cite{Pritts-ACCV18,Pritts-IJCV20}, which was
    the only competitive solver from the noise sensitivity
    experiments. The \EVL solver is $2153.6 \times$ faster than
    \rgntwotwotwodes, and the other proposed variants are orders of
    magnitude faster.}
  \label{tab:wall_clock}
  \vspace{-5pt}
\end{table}

\subsection{Real Images}
\label{sec:real_images}
In the experiments on real images shown in \figsref{fig:first} and
~\ref{fig:field_of_view}, we tested the proposed solvers on GoPro4
Hero 4 images with increasing field-of-view settings---medium and
wide, where the wider field-of-view setting generates more extreme
radial distortion since the full extent of the lens is used. To span
the gamut of lens distortions in the field-of-view study of
\figref{fig:field_of_view}, we included a Samyang 7.5mm fisheye
lens. The consistency of the undistortion estimate at the same GoPro
Hero4 field-of-view setting can be seen by comparing the undistortions
between the medium GoPro Hero 4 images in
\figref{fig:GoPro_Hero4_medium_fov} and the undistortions between the
wide GoPro images in \figsref{fig:first} and
\ref{fig:GoPro_Hero4_wide_fov}. Regardless of the significantly
different image content and sensor orientation, the undistortions are
of comparable magnitude at the same setting. Rectification are
accurate for all GoPro Hero 4 images, and the image of the distorted
vanishing line is correctly positioned (rendered in green) in the
original images. Despite using the 1-parameter division model for lens
undistortion, an excellent rectification is achieved for the fisheye
distorted image taken with the Samyang 7.5mm lens in
\figref{fig:samyang75mm_fov}, and the horizon line is perfectly
estimated.

\figref{fig:evl_results} shows results obtained with 1-correspondence
sampling using the proposed \rgntwordctevl \EVL solver on very
challenging fisheye images.
Images from five distinct fisheye lenses
are evaluated with \figsref{fig:NikonD300_15mm_evl},
\ref{fig:unknown_evl}, and \ref{fig:Nikon_D810_14mm_evl} having highly
oblique viewpoints of the dominant scene plane. Accurate
rectifications and undistortions are achieved for all images, and the
distorted image of the vanishing line (rendered in green) is correctly
positioned. The limitations of the 1-parameter division model can be
seen with extreme radial distortions, as, \eg,
\figsref{fig:unknown_evl} and \ref{fig:Olympus_E_M1_15mm_evl} exhibit
some mustache distortion, which cannot be modeled with 1
parameter. However, the local optimizer of \cite{Pritts-CVPR14} could
be modified to regress a higher-order distortion model using the
results of \figref{fig:evl_results} as an initial guess. We leave this
for future work.

\begin{figure*}[th!]
  \centering
  \newcommand{\trirow}[1]{%
    \setlength{\tabcolsep}{-0.06cm}
    \begin{tabular}{c}
      \includegraphics[height=2.6cm]{img2/#1.jpg}\\
      \includegraphics[height=2.6cm]{img2/#1_ud.jpg}\\
      \includegraphics[height=2.4cm]{img2/#1_rect_cropped.jpg}\\
    \end{tabular}
  }

  \trirow{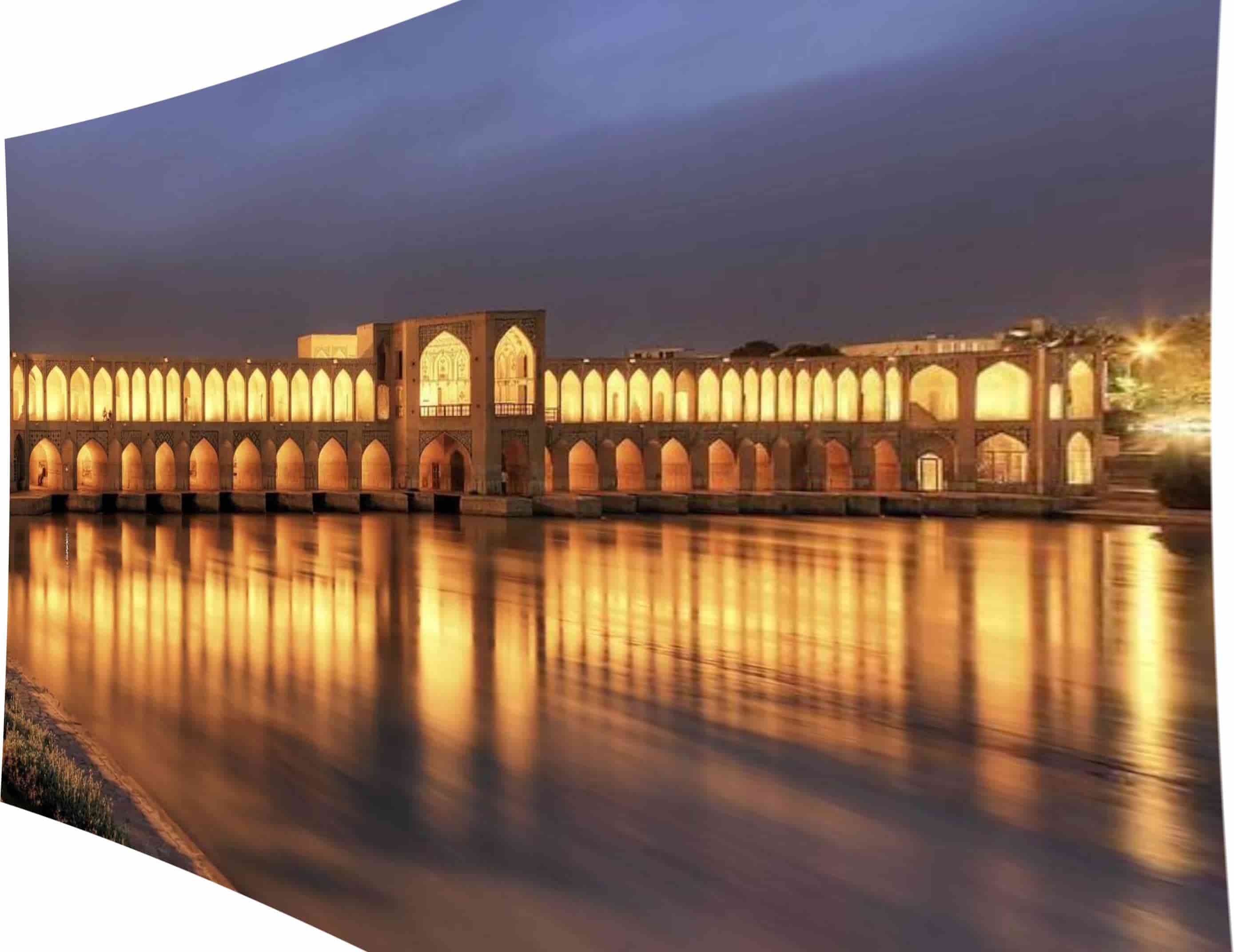}
  \trirow{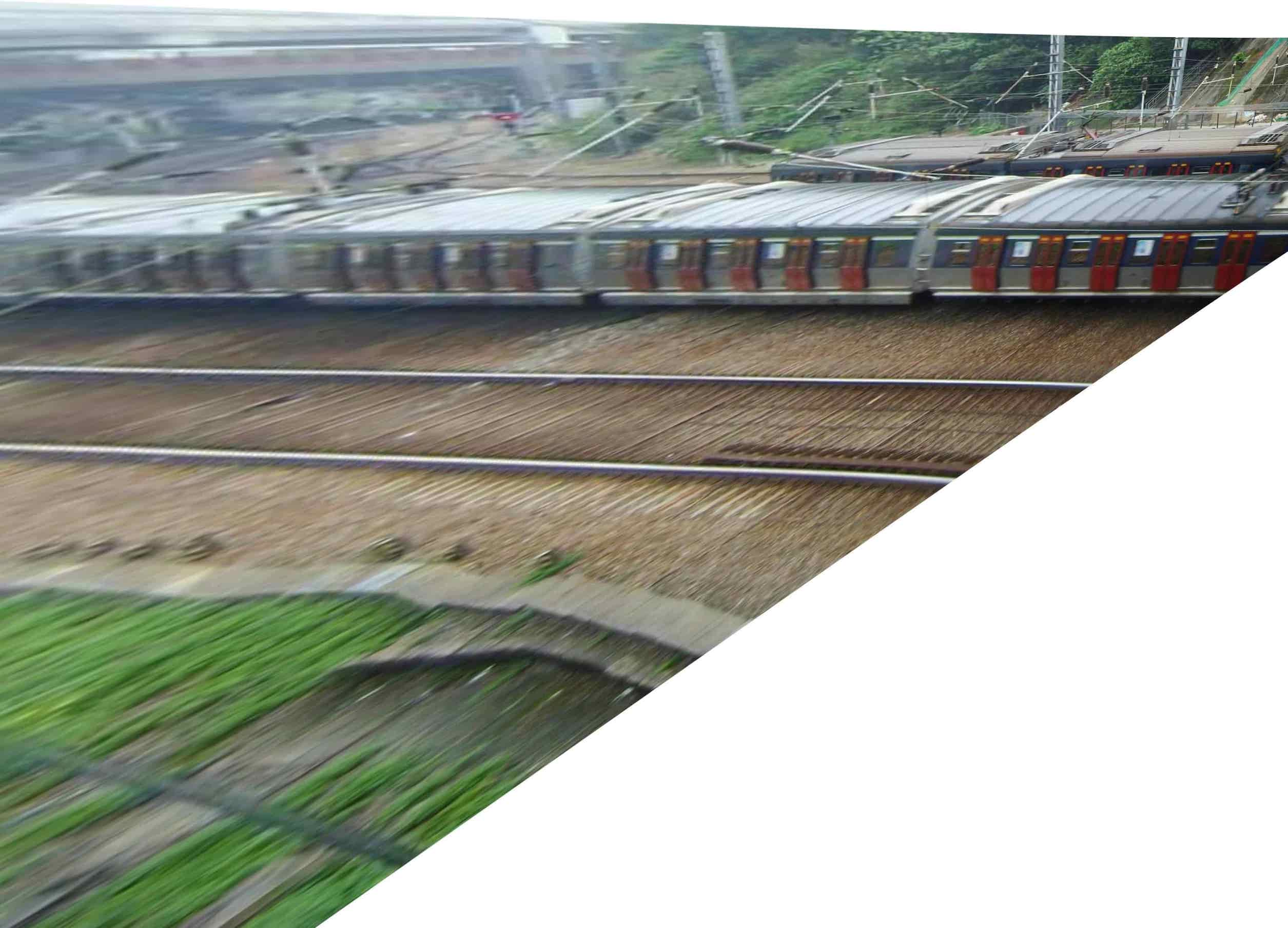}
  \trirow{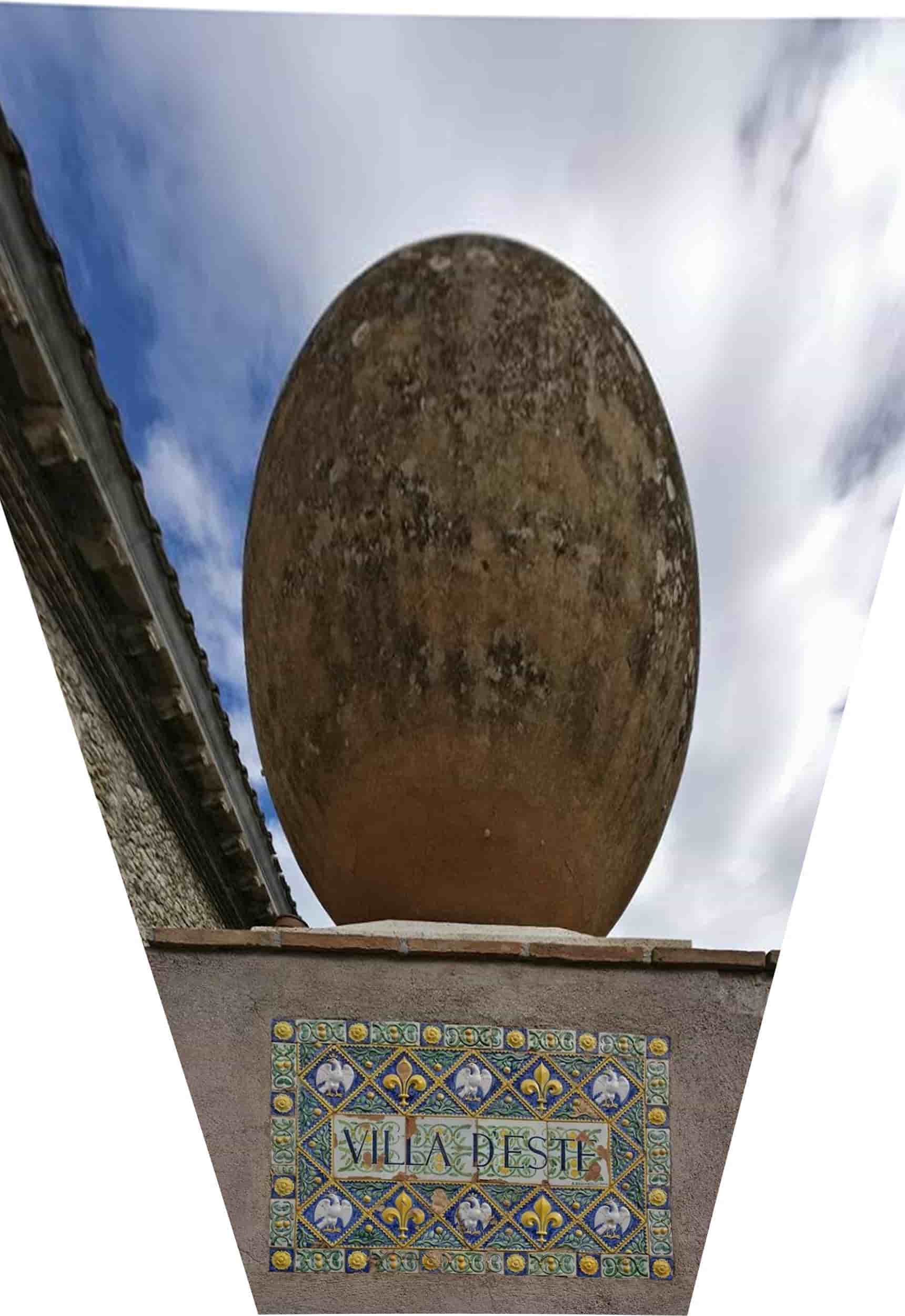}
  \trirow{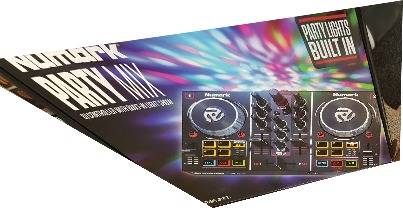}
  \trirow{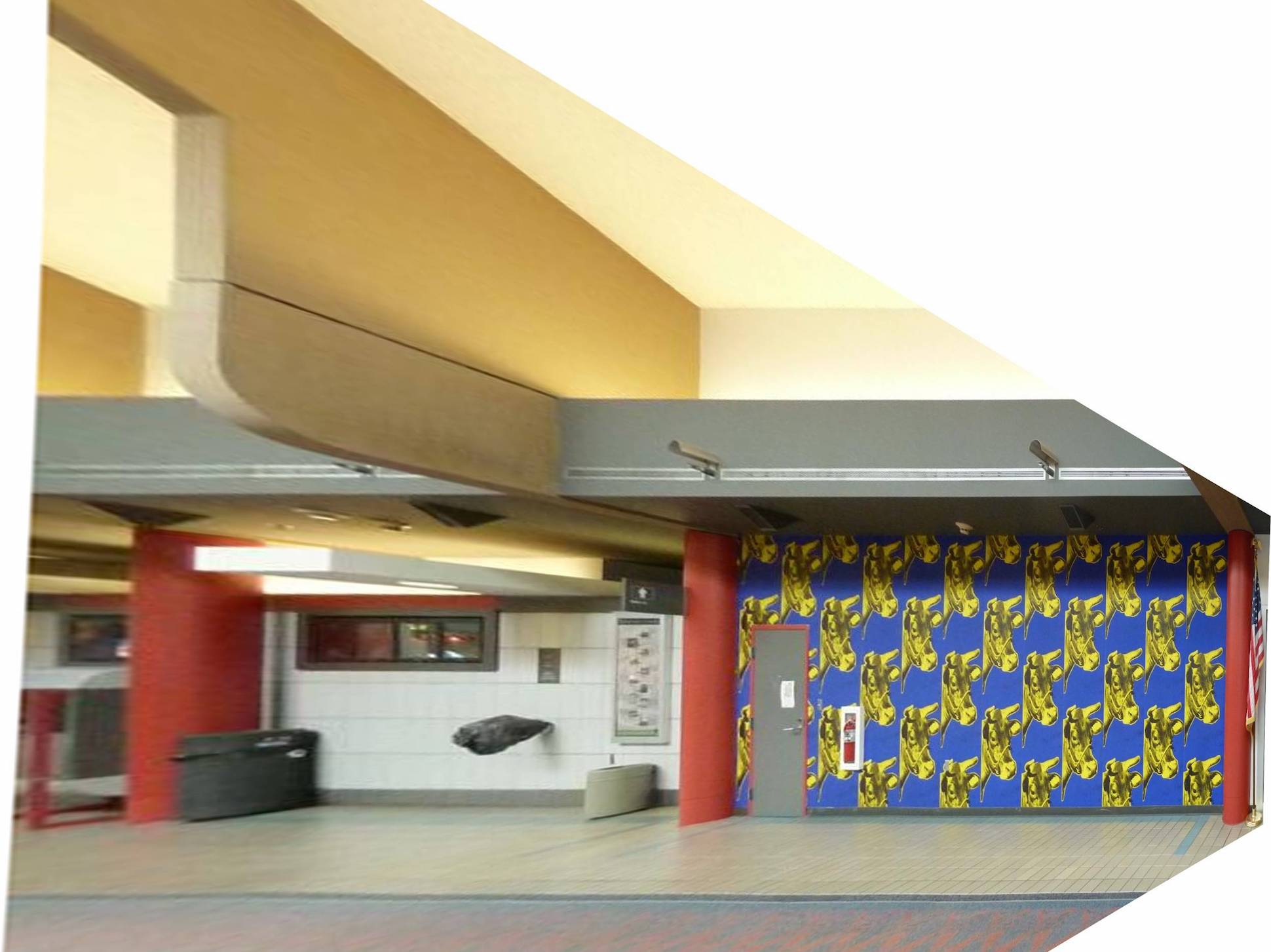}
  \caption{The proposed solvers works well on images with small
    lens distortions.  Input images are on the top row;
    undistorted images are on the middle row, and the rectified images
    are on the bottom. Results were generated with the \rgntwordctevp
    solver.}
  \label{fig:narrow_fov}
  \vspace{-5pt}
\end{figure*}

\figsref{fig:samyang75mm_fov}, \ref{fig:unknown_evl}, and
\ref{fig:Nikon_D810_14mm_evl} contain imaged scene planes with
vanishing lines that pass near the image origin (equivalently, center
of distortion), which is a degeneracy of the solver (see
\secref{sec:pami19_degeneracies}). Still excellent results are
achieved, which empirically demonstrates that even for vanishing lines
passing very close to the image center, the lens distortion is
sufficiently observable. In practice the degeneracy does not seem to
be a problem.

The distorted image evaluated in \figref{fig:real_wide} has a very low
inlier ratio of corresponded coplanar regions, which is typical for
repeated content that is clustered by appearance. The low-inlier
example is rectified with the five proposed solvers and the solvers
\rgntwotwotwodes of \cite{Pritts-ACCV18,Pritts-IJCV20} and
\rgntwotwoct of \cite{Chum-ACCV10}. All solvers were used within an
extension to the coplanar repeat detection and rectification framework
of Pritts \etal \cite{Pritts-CVPR14}. The ground truth division-model
parameter $\lambda_{\text{gt}}$ was obtained using calibration
software and chessboard images. The proposed \rgntwordctevl gives the
best undistortion and rectification both quantitatively and
qualitatively. The estimation framework using the \rgntwoct solver,
which does not solve for lens undistortion, is unable to recover a
reasonable lens undistortion. In general, the solvers requiring more
correspondences and having more degrees of freedom give less accurate
results. The experiment demonstrates the non-convexity of the problem,
and emphasizes the need for a good initial guess by the minimal solver
for the local optimizer of \cite{Pritts-CVPR14}.

The narrow field of view and diverse content experiment of
\figref{fig:narrow_fov} shows the performance of the proposed method
on imagery typical from cell phone cameras and near rectilinear
lenses. The left 3 columns of the study are challenging since the
conjugate translations and reflections are extracted a small strip of
the image. Still the rectifications are accurate. 

\section{Conclusions}
This paper proposes a suite of simple high-speed solvers for jointly
undistorting and affine-rectifying images containing
radially-distorted conjugate translations. The proposed solvers
contain variants that relax the assumptions that the preimages of
radially-distorted conjugately-translated point correspondences are
translated by the same magnitude in the scene plane, and that all
point correspondences translate in the same direction. Furthermore, a
variant is proposed that admits reflections of similarity-covariant
region correspondences, which is helpful for searching for
correspondences for semi-metric rectification. 

The \EVL \rgntwordctevl solver admits the same point configuration as
the one-direction \EVP solver \rgntwordctevp, but is much simpler
(\ie, does not require the \Gbs method), more stable, and is $7.4
\times$ faster in terms of wall-clock time to solution. The
improvement is given by the choice to eliminate the vanishing line
instead of the vanishing point. The significant difference emphasizes
the importance of care in solver design; in particular, the need to
simplify the constraint equations. While \Gbs related methods are
powerful and somewhat general, their blind application for solver
generation can result in slow and unstable solvers. \Eg, Pritts \etal
in \cite{Pritts-ACCV18,Pritts-IJCV20} were unable to reduce the
degree of their constraint equations used for the \rgntwotwotwodes
solver, which resulted in slow solver (see
\tabref{tab:wall_clock}). Furthermore, stability sampling was required
to generate useful solvers \cite{Larsson-CVPR18}.

Synthetic experiments show that the \EVP and \EVL solvers are
significantly more robust to noise in terms of the accuracy of
rectification and radially-distorted conjugate translation estimation
than the radial-distortion homography solvers of Fitzgibbon and
Kukelova \etal \cite{Fitzgibbon-CVPR01,Kukelova-CVPR15}. The
experiment verifies the importance of solving the minimal problem
since the extraneous degrees of freedom of the radial-distortion
homography solvers are free to explain the noise with incorrect
geometry. Furthermore, the proposed solvers are competitive with the
robustness of the state-of-the-art \rgntwotwotwodes solver of
\cite{Pritts-ACCV18,Pritts-IJCV20} despite the fact that the
\rgntwotwotwodes solver requires two more region correspondences as
input (compared to \rgntwordctevl,\rgntwordctevp, and
\rgntwordsctevp). The advantage of the proposed solvers is more
pronounced if the combinatorics of the robust \RANSAC estimator are
considered, where one correspondence sampling makes it possible to solve
scenes with a very-low proportion of good correspondences.

Experiments on difficult images with large radial distortions confirm
that the solvers give high-accuracy rectifications if used inside a
robust estimator. By jointly estimating rectification and radial
distortion, the proposed minimal solvers eliminate the need for
sampling lens distortion parameters in \RANSAC. 

\ifCLASSOPTIONcompsoc
\section*{Acknowledgments}
\else
\section*{Acknowledgment}
\fi

James Pritts and Yaroslava Lochman acknowledge the European Regional
Development Fund under the project Robotics for Industry 4.0
(reg. no. CZ.02.1.01/0.0/0.0/15\_003/0000470); Zuzana Kukelova the ESI
Fund, OP RDE programme under the project International Mobility of
Researchers MSCA-IF at CTU No. CZ.02.2.69/0.0/0.0/17\_050/0008025; and
Ond{\v r}ej Chum grant OP VVV funded project
CZ.02.1.01/0.0/0.0/16\_019/0000765 ``Research Center for Informatics''
and the ERC-CZ grant MSMT LL1901. Viktor Larsson was funded by the ETH
Zurich Postdoctoral Fellowship program and the Marie Sklodowska-Curie
Actions COFUND program. Yaroslava Lochman acknowledges ELEKS Ltd.

\ifCLASSOPTIONcaptionsoff
  \newpage
\fi

\bibliographystyle{IEEEtran}
\bibliography{IEEEabrv,main.bib}

%

\begin{IEEEbiography}[\vspace{-16pt}{\includegraphics[width=0.9in,height=1.25in,clip,keepaspectratio]{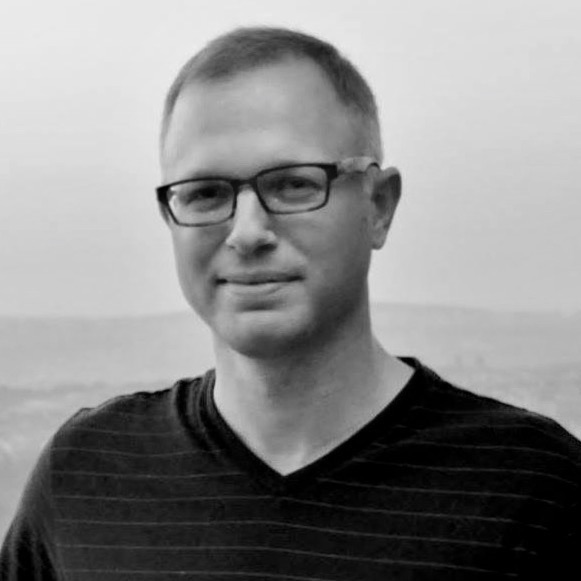}}]{James Pritts}
  is a Post-Doctoral Research Scientist at Facebook Reality
  Labs. Previously, Mr. Pritts was a researcher at the Czech Institute
  of Informatics, Robotics and Cybernetics at Czech Technical
  University in Prague. Mr. Pritts also serves as an adviser for The
  Machine Learning Lab at Ukrainian Catholic University in Lviv. He is
  the recipient of the ACCV 2018 Saburo Tsuji Best Paper Award and the
  ICVNZ 2013 Best Paper Award. He works mostly on minimal solvers and
  multi-model estimation.
\end{IEEEbiography}
\vspace{-15pt}
\begin{IEEEbiography}[{\vspace{-20pt}\includegraphics[width=0.9in,height=1.25in,clip,keepaspectratio]{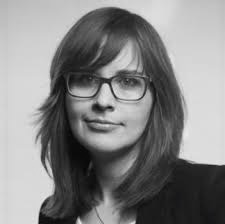}}\vspace{-25pt}]{Zuzana Kukelova}
is a research fellow at the Visual Recognition Group at the Department
of Cybernetics, Faculty of Electrical Engineering, Czech Technical
University in Prague.  She was a postdoctoral researcher at Microsoft
Research Cambridge, United Kingdom.
Her research interests include algebraic geometry and computer vision,
where she focuses on minimal problems and methods for solving systems
of polynomial equations.
She is the recipient of the 2015 Cor Baayen Award, given to a
promising young researcher in computer science and applied mathematics
from countries associated with the European Research Consortium for
Informatics and Mathematics.
\end{IEEEbiography}
\vspace{-12pt}
\begin{IEEEbiography}[\vspace{-28pt}{\includegraphics[width=0.9in,height=1.25in,clip,keepaspectratio]{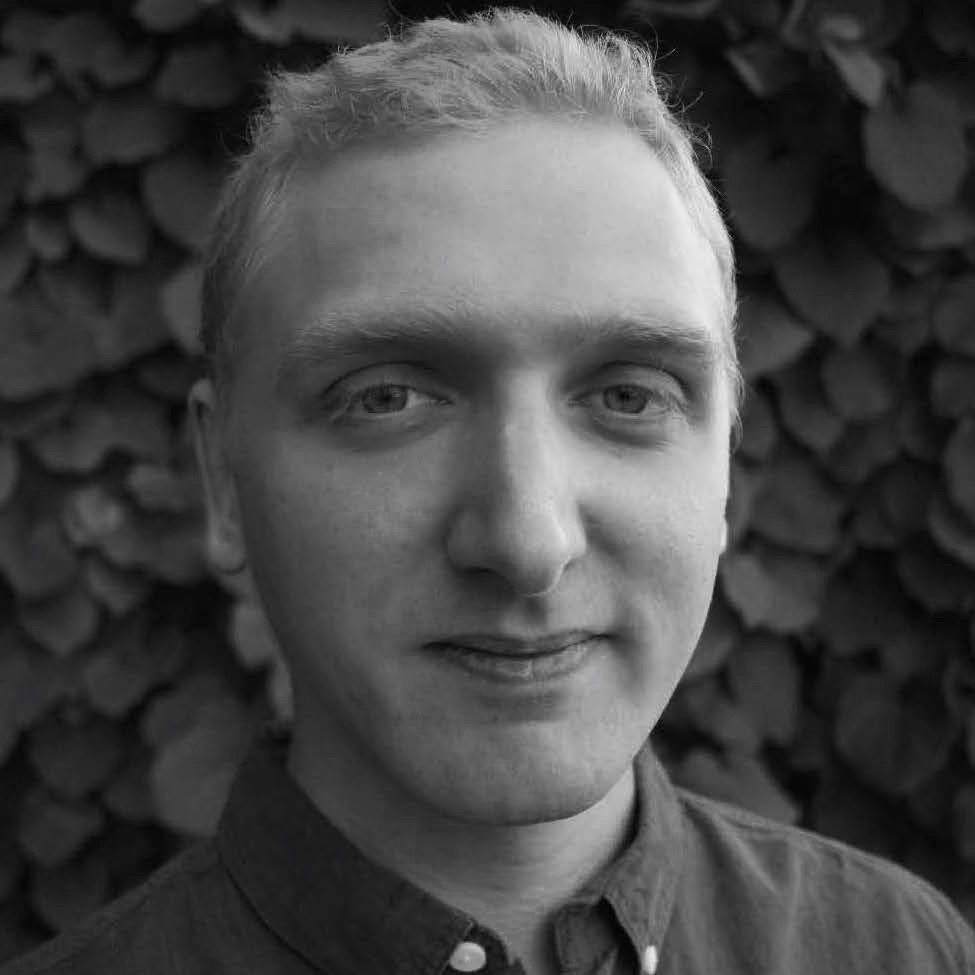}}]{Viktor Larsson} received his PhD degree from the Centre for Mathematical Sciences at Lund University in Sweden. He is currently a post-doctoral researcher at the Computer Vision and Geometry Group at ETH Zurich led by Prof. Marc Pollefeys. His work is mostly focused on minimal problems in geometric computer vision.
\end{IEEEbiography}
\vspace{-35pt}
\begin{IEEEbiography}[\vspace{-19pt}{\includegraphics[width=0.9in,height=1.25in,clip,keepaspectratio]{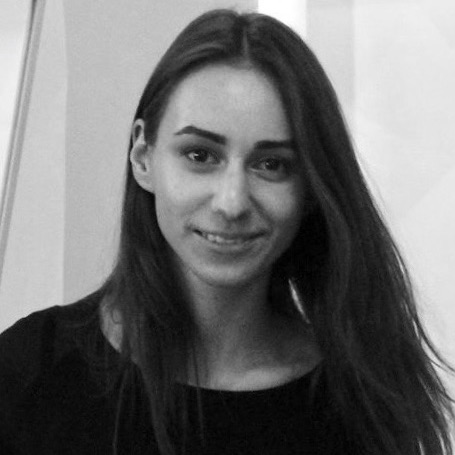}}]{Yaroslava Lochman} is a research intern at Facebook Reality Labs. She received her MSc in data science from Ukrainian Catholic University in Lviv, where she also conducted research in computer vision at The Machine Learning Lab.
She is currently focused on robust methods for camera calibration.
\end{IEEEbiography}
\vspace{-35pt}
\begin{IEEEbiography}[\vspace{-18pt}{\includegraphics[width=1in,height=1.25in,clip,keepaspectratio]{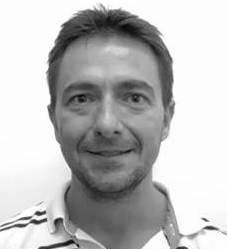}}\vspace{-18pt}]{Ond{\v r}ej~Chum}
 is an associate professor at the Czech Technical University in
 Prague, where he leads a team within the Visual Recognition Group at
 the Department of Cybernetics, Faculty of Electrical Engineering. He
 received the MSc degree in computer science from Charles University,
 Prague, in 2001 and the PhD degree from the Czech Technical
 University in Prague, in 2005. From 2006 to 2007, he was a
 postdoctoral researcher at the Visual Geometry Group, University of
 Oxford, United Kingdom. The research interests include large-scale
 image and particular object retrieval, object recognition, and robust
 estimation of geometric models. He is a member of Image and Vision
 Computing editorial board, and has served in various roles at major
 international conferences (e.g., ICCV, ECCV, CVPR, and BMVC). He was
 the recipient of the Best Paper Prize at the BMVC in 2002, the Best
 Science Paper Honorable Mention at BMVC 2017, Longuet-Higgins Prize
 at CVPR 2017, and the Saburo Tsuji Best Paper Award at ACCV
 2018. Ond{\v r}ej was awarded the 2012 Outstanding Young Researcher in
 Image and Vision Computing runner up for researchers within seven
 years of their PhD.
\end{IEEEbiography}


\vfill


\end{document}